%% file: template.tex
\documentclass[twocolumn,10pt]{article}          %
\usepackage{cvpr}

\usepackage{times}

\usepackage{graphicx}
\graphicspath{{./}{./figure}{./figures}}

\usepackage{amsmath}
\usepackage{amssymb, cite, url}

\usepackage{bm}
\usepackage{icomma}
\usepackage[ruled,lined]{algorithm2e}

\newcommand{\eg}{\emph{e.g.}}

\newcommand{\ie}{\emph{i.e.}}
\newcommand{\cf}{\emph{c.f.}}
\newcommand{\itm}{a}
\renewcommand{\i}{{\itm}}
\newcommand{\vs}{\emph{vs.}}

\newcommand{\supp}{{\mathrm{supp}}}
\newcommand{\conf}{{\mathrm{conf}}}

\def\citep{\cite}

\definecolor{blue}{rgb}{0,0,0}

\definecolor{green}{rgb}{0,0,1}

\cvprfinalcopy

\begin{document}

\title{Mining Mid-level Visual Patterns with Deep CNN Activations\thanks{The authors are with the
School of Computer Science, The University of Adelaide, Australia.
 C. Shen is the corresponding author:
(e-mail: chunhua.shen@adelaide.edu.au).
}
}

\author{Yao Li,
        Lingqiao Liu,
        Chunhua Shen,
        Anton van den Hengel  %
}

\date{}

\maketitle

\begin{abstract}

The purpose of mid-level visual element discovery is to find clusters of
image patches that are representative of, and which discriminate between, the contents of the relevant images. Here we propose a pattern-mining approach to the problem of identifying mid-level elements within images, motivated by the observation that such techniques have been very effective, and efficient, in achieving similar goals when applied to other data types. We show that  \textcolor{blue}{CNN activations} extracted from image patches typical possess two appealing properties that enable seamless integration with pattern mining techniques. The marriage between \textcolor{blue}{CNN activations and a pattern mining technique} leads to fast and effective discovery of representative and discriminative patterns from a huge number of image patches, \textcolor{blue}{from which mid-level elements are retrieved}.
 Given the patterns and
  retrieved mid-level visual elements, we propose two methods to generate image feature representations. The first encoding method uses the patterns as codewords in a dictionary in a manner similar to the Bag-of-Visual-Words model.
We thus label this a Bag-of-Patterns representation. The second relies on mid-level visual elements to construct a Bag-of-Elements representation. We evaluate the two encoding methods on object and scene classification tasks, and demonstrate that our approach outperforms or matches the performance of the state-of-the-arts on these tasks.

\end{abstract}
{\rm keywords: Mid-level visual element discovery,  pattern mining, convolutional neural networks}

\tableofcontents
\clearpage

\section{Introduction}
\label{sec:introduction}
\input{intro.tex}
\section{Related work}
\label{sec:related_work}
\input{relate_work.tex}
\section{Background on pattern mining}
\label{sec:pattern_mining_revisited}
\input{mining_background.tex}

\section{Mid-level deep pattern mining}
\label{sec:algorithm}
\input{algorithm.tex}
\section{Image representation}
\label{sec:encoding}

\input{encoding1.tex}

\section{Experiments}
\label{sec:experiment}
\input{experiments.tex}

\section{Discussion }
\label{sec:discussion}
\input{discussion.tex}
\section{Conclusion and future work}
\label{sec:conclusion}
\input{conclusion_future_work.tex}

{\small
\bibliographystyle{ieee}      %
\bibliography{egbib}
}
\end{document}

%% file: intro.tex
Image patches that capture important aspects of objects are crucial to a variety of state-of-the-art object recognition systems.
For instance, in the Deformable Parts Model (DPM) \cite{DBLP:journals/pami/FelzenszwalbGMR10}
such image patches
represent
object parts that are treated as latent variables in the training process.
In Poselets~\citep{DBLP:conf/iccv/BourdevM09}, such image patches are
used to represent
human body parts, which have been shown to be beneficial for human detection~\citep{DBLP:conf/eccv/BourdevMBM10}
and human attribute prediction~\citep{DBLP:conf/iccv/BourdevMM11} tasks.
Yet, obtaining these informative image patches in both DPM and Poselets
require extensive
human annotations (DPM needs object bounding boxes while the Poselets model needs the information of human body keypoints).
Clearly, the discovery of these representative image patches with minimal human
supervision would be desirable.
Studies on \emph{mid-level visual elements} (\emph{a.k.a, mid-level discriminative patches})
offer one possible solution to this problem.

Mid-level visual elements are clusters of image patches discovered from
a dataset where only image labels are available.
As noted in the pioneering work of~\cite{DBLP:conf/eccv/SinghGE12},
such patch clusters are suitable for interpretation as mid-level visual elements only if they
satisfy two requirements, \ie, \emph{representativeness} and \emph{discriminativeness}.
Re\-pre\-sen\-ta\-ti\-ve\-ness re\-qu\-i\-r\-e\-s that mid-level visual elements should frequently occur in the images with same label (\ie, target category), while discriminativeness implies that they should be seldom found in %
images not containing the object of interest.
For instance, image patches containing the wheel of a car may be a mid-level visual element for the car category,
as
most car images contain wheels, and car wheels are seldom found in images of other objects (this implies also that they are visually distinct from other types of wheels).
The discovery of mid-level visual elements has boosted performance in a variety of vision tasks, such as image classification~\citep{DBLP:conf/eccv/SinghGE12,DBLP:conf/cvpr/JunejaVJZ13,
DBLP:conf/nips/DoerschGE13} and action recognition~\citep{DBLP:conf/cvpr/JainGRD13,DBLP:conf/cvpr/WangQT13}.
\textcolor{blue}{As another line of research, pattern mining techniques have also enjoyed popularity amongst the computer vision community, including image classification~\citep{DBLP:conf/cvpr/YaoF10,
DBLP:conf/eccv/FernandoFT12,DBLP:journals/ijcv/FernandoFT14,
DBLP:conf/cvpr/Voravuthikunchai14}, image retrieval~\citep{DBLP:conf/iccv/FernandoT13} and action recognition
~\citep{DBLP:journals/pami/GilbertIB11,DBLP:conf/accv/GilbertB14}, largely
to due to their capability of discovering informative patterns hidden inside massive of data.}

\begin{figure}[t] \vspace{-0.0cm} \begin{center}
\scalebox{0.9}{

\begin{tabular}{@{}c@{}c@{}c}
\includegraphics[width=0.35\linewidth]{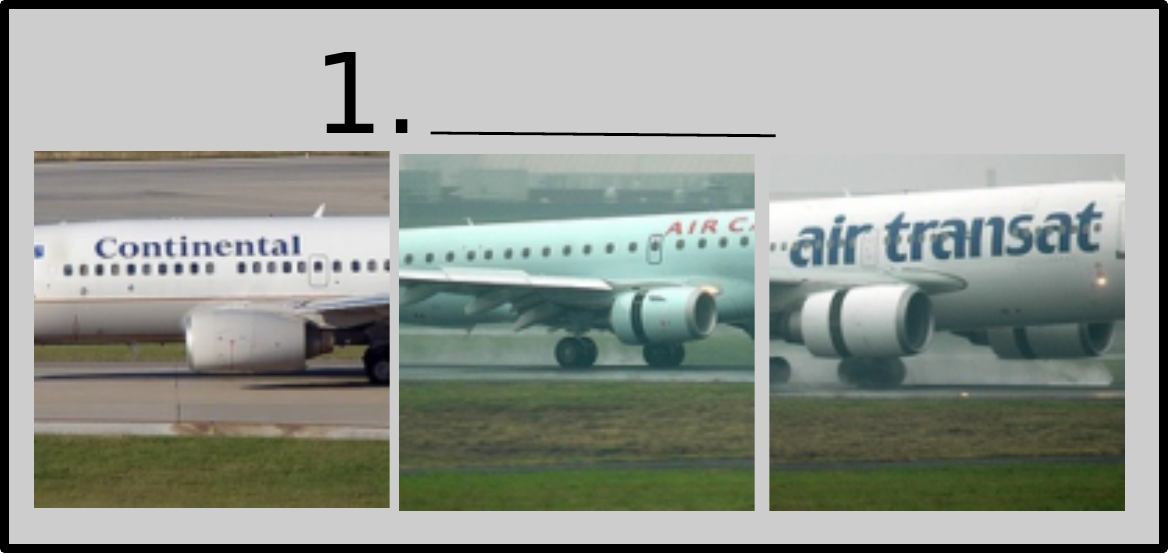} \ &
\includegraphics[width=0.35\linewidth]{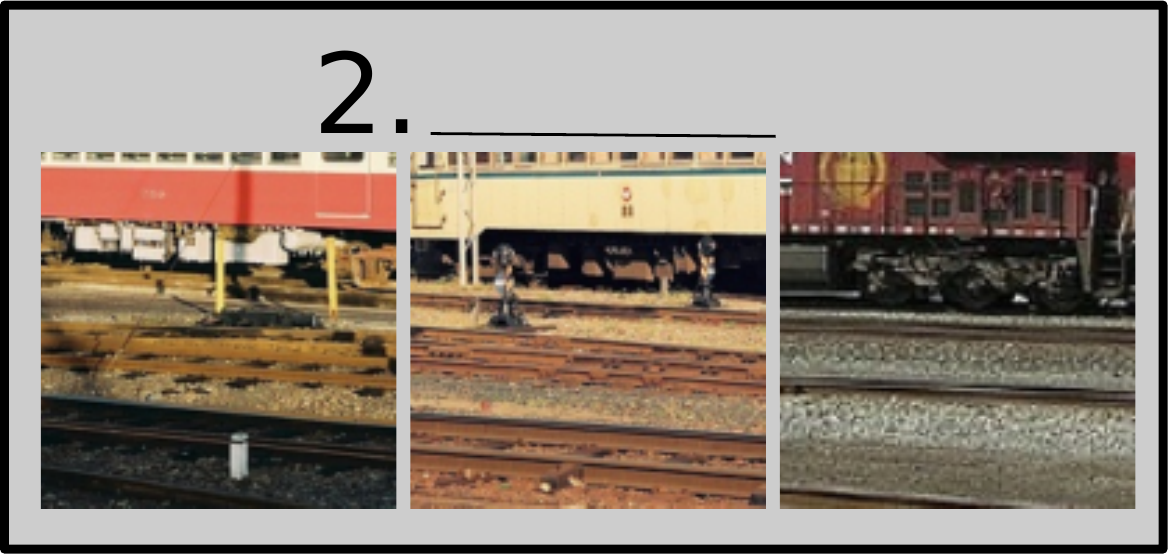} \ &
\includegraphics[width=0.35\linewidth]{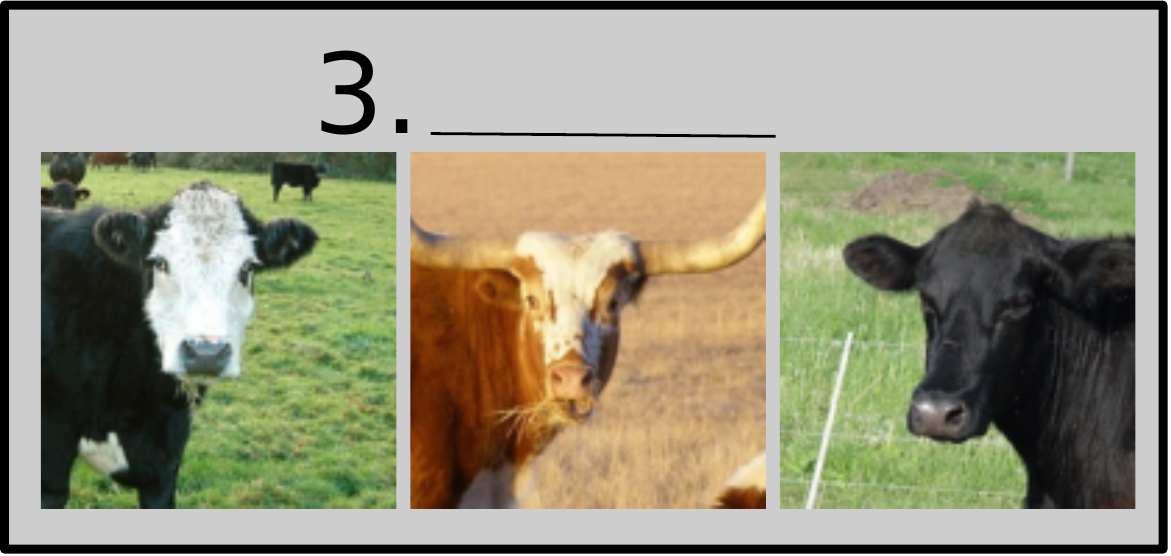} \\
\includegraphics[width=0.35\linewidth]{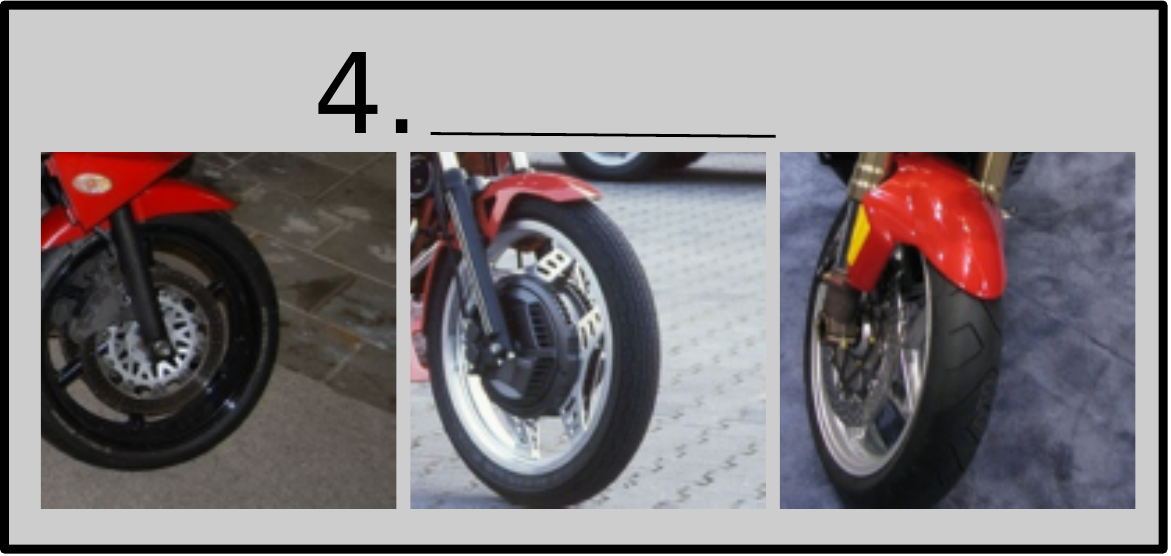} \ &
\includegraphics[width=0.35\linewidth]{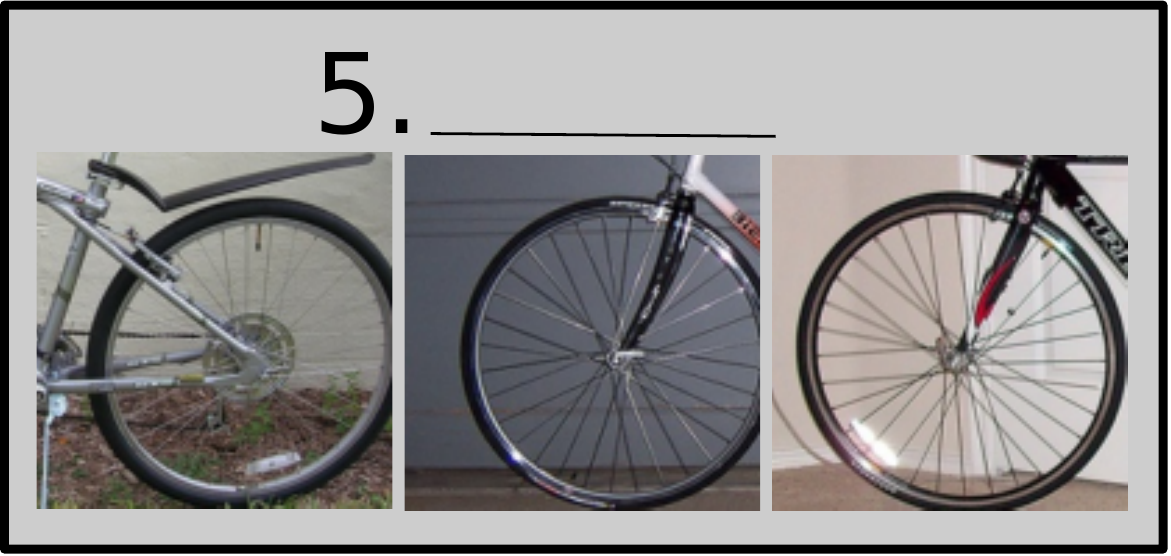} \ &
\includegraphics[width=0.35\linewidth]{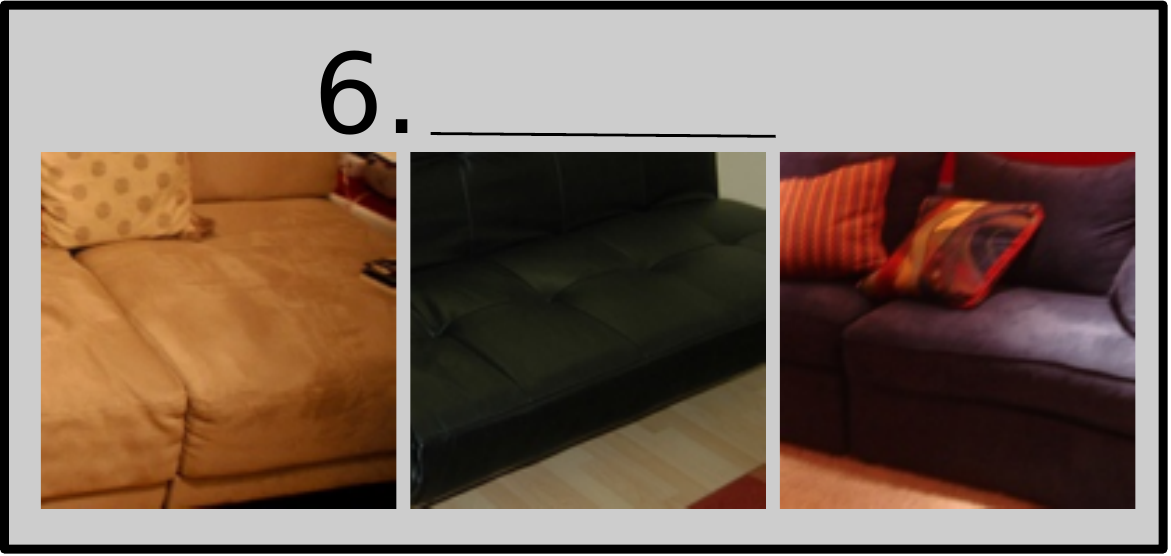} \\
\end{tabular} } \end{center} \caption{\emph{Name that Object:} Given the mid-level visual elements discovered by our algorithm from the Pascal VOC 2007 dataset, can you guess what categories are they from?}
\label{fig:Name_that_object}
\end{figure}

\textcolor{blue}{In this paper, we address mid-level visual element discovery from a pattern mining perspective.}
The novelty in our approach of  is that it systematically brings together
Convolutional Neural Networks (CNN) activations and association rule mining, a
well-known pattern mining technique.
Specifically, we observe that for an image patch, activations extracted from
fully-connected layers of a CNN possess two appealing properties which enable
their seamless integration with this pattern mining technique.
Based on this observation, we formulate mid-level visual element discovery from
the perspective of pattern mining and propose a \emph{Mid-level Deep Pattern Mining} (MDPM) algorithm that effectively and efficiently discovers representative and discriminative
patterns from a huge number of image patches.
When we retrieve and visualize image patches with the same pattern, it turns
out that they are not only visually similar, but also semantically consistent
(see by way of example the game in Fig.~\ref{fig:Name_that_object} and then check your answers
  below\footnote{Answer key: 1.~aeroplane, 2.~train, 3.~cow, 4.~motorbike, 5.~bike,
6.~sofa.}).
Relying on the discovered patterns and retrieved mid-level visual elements, we
propose two methods to generate image features for each of them
(Sec.~\ref{sec:encoding}).
For the first feature encoding method, we compute a Bag-of-Patterns
representation which
is motivated by
the well-known Bag-of-Visual-Words
representation~\citep{DBLP:conf/iccv/SivicZ03}.
For the second method, we first merge mid-level visual elements and train
detectors simultaneously, followed by the construction of a Bag-of-Elements
representation.
We evaluate the proposed feature representations on generic object and scene
classification tasks.
Our experiments  demonstrate that  the classification performance of the
proposed feature representation not only outperforms all current methods in
mid-level visual element discovery by a noticeable margin with far fewer
elements used, but also outperform or match the performance of state-of-the-arts using CNNs for the same task.

In summary, the merits of the proposed approach can be understood from different prospectives.

\begin{itemize}

\item \noindent\emph{Efficient handling of massive image patches.}
As noted by~\cite{DBLP:conf/nips/DoerschGE13}, one of the challenges
in mid-level visual element discovery is the massive amount of random sampled patches to go through.
However, pattern mining techniques are designed to handle large data sets, and are extremely capable of doing so.
In this sense, if appropriately employed, pattern mining techniques can be a powerful tool for overcoming this data deluge in mid-level visual element
discovery.

\item \noindent\emph{A straightforward interpretation of re\-presen\-tat\-iv\-ness an\-d dis\-criminativeness}.
In previous works on mid-level visual element discovery, different methods have been proposed for interpreting the dual requirements of
representativeness and discriminativeness.
Here in this work,  %
interpreting these two requirements in the
pattern mining terminology is straightforward.
To our knowledge, we are the first to formulate mid-level visual element discovery from the perspective of pattern mining.

\item \noindent\emph{Feature encoder of CNN activations of image patches.}
Recent state-of-the-art results on many image classification tasks (\eg, indoor scene, object, texture) are achieved by applying classical feature encoding methods~\citep{DBLP:conf/eccv/PerronninSM10,DBLP:conf/cvpr/JegouDSP10} on
the top of CNN activations of image patches~\citep{DBLP:conf/eccv/GongWGL14,CimpoiMV15CVPR,
DBLP:journals/ijcv/CimpoiMKV16}.
In our work, we demonstrate that mid-level visual elements, which are discovered by the proposed MDPM algorithm, can also be a good
alternative feature encoder for CNN activations of image patches.
\end{itemize}

The remainder of the paper is organized as follows.
In Sec.~\ref{sec:related_work}, we review
some of the related work on mid-level visual element discovery as well as relevant vision applications.
In Sec.~\ref{sec:pattern_mining_revisited} we explain some of the relevant pattern mining terminology and how pattern mining techniques have been successfully applied to
computer vision tasks previously.
The details of our MDPM algorithm are provided in Sec.~\ref{sec:algorithm}.
In particular, we start by introducing two desirable properties of CNN activations extracted from
image patches (Sec.~\ref{subsec:properities}), which serve as the cornerstones of the proposed MDPM algorithm.
In Sec.~\ref{sec:encoding}, we apply the discovered patterns and mid-level visual elements to generate image feature representations, followed by extensive experimental validations in Sec.~\ref{sec:experiment}.
\textcolor{blue}{Some further discussions are presented in Sec.~\ref{sec:discussion} and
we conclude the paper in Sec.~\ref{sec:conclusion}.}
Preliminary results of this work appeared in \cite{LiLSH15CVPR}.
In this paper, we extend \cite{LiLSH15CVPR} in the following aspects.
Firstly, for the theory part, we propose a new method to generate image representations using the discovered patterns
(\ie, the Bag-of-Patterns representation).
Furthermore, more extensive experiment are presented in
this manuscript, such as more detailed analysis of different components of the proposed framework.
Last but not least, we present a new application of mid-level visual elements, which is the analysis of the role of context information using mid-level visual elements (Sec.~\ref{sec:context}).
\textcolor{blue}{At the time of preparing of this manuscript, we are aware of at least two
works~\citep{DBLP:conf/cvpr/DibaPPG,oramas2016modeling} which are built on our previous work~\cite{LiLSH15CVPR} in different vision applications,
including human action and attribute recognition~\citep{DBLP:conf/cvpr/DibaPPG} and modeling visual compatibility~\citep{oramas2016modeling},
which reflects that our work is valuable to the computer vision community.
Our code is available at \url{https://github.com/yaoliUoA/MDPM}.}

%% file: relate_work.tex
\subsection{\textcolor{blue}{Mid-level visual elements}}
\label{subsec:mid_level_related_work}
\textcolor{blue}{
Mid-level visual features have been widely used in computer vision,
which can be constructed by different methods, such as supervised dictionary learning \citep{DBLP:conf/cvpr/BoureauBLP10},
hierarchically encoding of low-level descriptors \citep{DBLP:journals/ijcv/AgarwalT08,DBLP:conf/nips/SimonyanVZ13,
DBLP:journals/ijcv/FernandoFT14} and the family of mid-level visual elements~\citep{DBLP:conf/eccv/SinghGE12,DBLP:conf/cvpr/JunejaVJZ13,
DBLP:conf/nips/DoerschGE13}.
As the discovery of mid-level visual elements is the very topic of this paper, we mainly discuss previous works on this topic.  }

Mid-level visual element discovery has been shown to be beneficial
to image classification tasks, including scene categorization~\citep{DBLP:conf/eccv/SinghGE12,
DBLP:conf/nips/DoerschGE13,DBLP:conf/cvpr/JunejaVJZ13,DBLP:conf/iccv/SunP13,
DBLP:conf/cvpr/LiWT13,DBLP:conf/icml/WangWBLT13,DBLP:conf/eccv/BossardGG14,
Parizi15ICLR,LiLSH15CVPR,DBLP:conf/iccv/MatzenS15} and fine-grained categorization~\citep{DBLP:conf/cvpr/WangCMD16}.
\textcolor{blue}{For this task, there are three key steps, (1) discovering candidates of mid-level visual elements, (2) selecting a subset of the candidates, and finally (3) generating image feature representations. }

\textcolor{blue}{
In the first step, various methods have been proposed in previous work to discover candidates of mid-level visual elements in previous works.
Usually starting from random sampled patches which are weakly-labeled (\eg, image-level labels are known),
candidates are discovered from the target category by different methods, such as cross-validation training patch detectors~\citep{DBLP:conf/eccv/SinghGE12},
training Exemplar LDA detectors~\citep{DBLP:conf/cvpr/JunejaVJZ13},
discriminative mode seeking~\citep{DBLP:conf/nips/DoerschGE13},
minimizing a latent SVM object function with a group sparsity regularizer~\citep{DBLP:conf/iccv/SunP13,DBLP:journals/ijcv/SunP16},
and the usage of Random Forest~\citep{DBLP:conf/eccv/BossardGG14}.
In this work, we propose a new algorithm for discovering the candidates from a pattern mining perspective (Sec.~\ref{sec:algorithm}). }

\textcolor{blue}{
The goal of the second step is to select mid-level visual elements from a large pool of candidates, which can best interpret the requirements of representative and
discriminative.
Some notable criteria in previous includes a combination of purity and discriminativeness scores~\citep{DBLP:conf/eccv/SinghGE12}, entropy ranking
~\citep{DBLP:conf/cvpr/JunejaVJZ13,DBLP:conf/iccv/LeeEM13}. the Purity-Coverage plot~\citep{DBLP:conf/nips/DoerschGE13} and the squared whitened norm response
~\citep{DBLP:conf/cvpr/AubryMERSJ14,DBLP:journals/tog/AubryRS14}.
In our work, we select mid-level visual elements from the perspective of pattern selection
(Sec.~\ref{subsubsec:pattern_Sselection}) and merging (Sec.~\ref{subsubsec:merging_detectors}). }

\textcolor{blue}{
As for the final step of generating image feature representation for classification, most previous
works~\citep{DBLP:conf/eccv/SinghGE12,DBLP:conf/cvpr/JunejaVJZ13,
DBLP:conf/nips/DoerschGE13} follow the same principle, that is,  the combination of maximum detection scores of all mid-level elements from different categories in a spatial pyramid~\citep{DBLP:conf/cvpr/LazebnikSP06}.
This encoding method is also adopted in our work (Sec.~\ref{subsubsec:Bag-of-Elements}).}

\textcolor{blue}{
In addition to image classification, some works apply mid-level visual elements to other vision tasks as well,
including visual data mining~\citep{DBLP:journals/tog/DoerschSGSE12,Rematas15CVPR},
action recognition~\citep{DBLP:conf/cvpr/JainGRD13,DBLP:conf/cvpr/WangQT13},
discovering stylistic elements~\citep{DBLP:conf/iccv/LeeEM13}, scene understanding~\citep{DBLP:conf/iccv/FouheyGH13,DBLP:conf/iccv/OwensXTF13,
DBLP:conf/iccv/FouheyHGH15},
person re-identification~\citep{DBLP:conf/cvpr/ZhaoOW14},
image re-r\-a\-n\-k\-i\-n\-g~\citep{DBLP:conf/bmvc/CrowleyZ14},
weakly-supervised object detection~\citep{DBLP:conf/nips/SongLJD14}.
In object detection, before the popularity of R-CNN~\citep{DBLP:journals/pami/GirshickDDM16},
approaches on object detection by learning a collection of mid-level detectors are illustrated by \citep{DBLP:conf/cvpr/EndresSJH13,
DBLP:journals/pami/ShihEH15,bansal2015mid}.}
\subsection{\textcolor{blue}{Pattern mining in computer vision}}
\label{subsec:mining_related_work}
\textcolor{blue}{
Pattern mining techniques, such as frequent itemset mining and its variants, have been studied primarily amongst the data mining community, but a growing number of applications can be found in the computer vision community.}

\textcolor{blue}{Early works have used pattern mining techniques in
object recognition tasks, such as finding frequent co-occurrent
visual words~\citep{DBLP:conf/cvpr/YuanWY07} and discovering distinctive feature configurations~\citep{DBLP:conf/iccv/QuackFLG07}.
Later on, for recognizing human-object interactions, \cite{DBLP:conf/cvpr/YaoF10} introduce `gouplets' discovered in a pattern
mining algorithm, which encodes appearance, shape and spatial relations of multiple image patches.
For 3D human action recognition, discriminative actionlets are discovered in a pattern mining fashion \citep{DBLP:journals/pami/WangLWY14}.
By finding closed patterns from local visual word histograms,~\cite{DBLP:conf/eccv/FernandoFT12,DBLP:journals/ijcv/FernandoFT14} introduce Frequent Local Histograms (FLHs) which can be utilized to generate
new image representation for classification.
Another interesting work is~\cite{DBLP:conf/cvpr/Voravuthikunchai14} in which images are represented by histograms of pattern sets.
Relying on a pattern mining technique, ~\cite{DBLP:conf/iccv/FernandoT13}
illustrate how to address the image retrieval problem using mid-level patterns.
More recently, \cite{Rematas15CVPR} design a method for summarizing image collections using closed patterns.
Pattern mining techniques have been also successfully applied to some other vision problems, such as action recognition in videos
~\citep{DBLP:journals/pami/GilbertIB11,DBLP:conf/accv/GilbertB14}.}

\textcolor{blue}{For the image classification task, most of the aforementioned works are relying on hand-crafted features, especially Bag-of-visual-words~\citep{DBLP:conf/iccv/SivicZ03}, for pattern mining.
In contrast, to our knowledge, we are first to describe how pattern mining
techniques can be combine with the state-of-the-art CNN features, which have been widely applied in computer vision nowadays.
Besides, our work can be viewed as a new application of pattern mining techniques in vision, that is, the discovery of mid-level visual elements.
}

%% file: mining_background.tex
\subsection{\textcolor{blue}{Terminology}} 
\label{subsec:terminologies}
Originally developed for market basket analysis, frequent itemset and association rule are well-known terminologies within data
mining.
Both might be used in processing large numbers of customer transactions to reveal information about their shopping behaviour, for example.

More formally, let $A=\{\itm_1,\itm_2,\ldots,\itm_M\}$ denote a set of $M$ items.
A transaction $T$ is a subset of $A$ (\ie, $T \subseteq A$) which contains
only a subset of items ($|T| \ll M$).
We also define a transaction database $ \mathcal{D}=\{T_1,T_2,\ldots,T_N\}$  containing $N$ (typically millions, or more)
transactions.
Given these definitions, the frequent itemset and association rule are
defined as follows.\\

\noindent
\textbf{Frequent itemset.}
A pattern $P$ is also a subset of $A$ (\ie, itemset). 
We are interested in the fraction of transactions $T \in \mathcal{D}$ which contain $P$.
The \emph{support} of $P$ reflects this quantity:
\begin{equation}
\supp(P) = \frac{|\{T|T\in \mathcal{D},P\subseteq T\}|}{N} \in [0,1],
\end{equation}
where $|\cdot|$ measures the cardinality.
$P$ is called a \emph{frequent itemset} when $\supp(P)$ is larger than a predefined threshold.
\vspace{1mm}
\newline
\textbf{Association rule.} An \emph{association rule} $I\rightarrow \i$ implies a relationship between pattern $P$ (antecedents) and an item $\i$ (consequence).
We are interested in how likely it is that $\i$ is present in the transactions which contain $P$ within $\mathcal{D}$ .
In a typical application this might be taken to imply that customers who bought items in $P$ are also likely to buy item $\i$, for instance.
The \emph{confidence} of an association rule $\conf(P\rightarrow \i)$ can be taken to reflect this probability:
\begin{equation}
\label{eq:confidence_definition}
\begin{split}
\conf(P\rightarrow \i) &= \frac{\supp(P\cup\{\i\})}{\supp(P)}\\
    & = \frac{|\{T|T\in \mathcal{D},(P\cup\{\i\})\subseteq T\}|}{|\{T|T\in \mathcal{D},P\subseteq T\}|} \in [0,1].
\end{split}
\end{equation}
In practice, we are interested in ``good'' rules, meaning that the confidence of these rules should be reasonably high.\\

\noindent\textcolor{blue}{\emph{A running example}. Consider the case when there are $4$ items in the set (\ie, $A=\{\itm_1,\itm_2,\itm_3,\itm_4\}$) and $5$ transactions in $\mathcal{D}$, 
\begin{itemize}
\item $T_1 = \{\itm_3,\itm_4\}$,
\item $T_2 = \{\itm_1,\itm_2,\itm_4\}$,
\item $T_3 = \{\itm_1,\itm_4\}$,
\item $T_4 = \{\itm_1,\itm_3,\itm_4\}$,
\item $T_5 = \{\itm_1,\itm_2,\itm_3,\itm_4\}$,
\end{itemize}
The value of $supp(\{\itm_1,\itm_4\})$ is $0.8$ as the itemset (pattern)
$\{\itm_1,\itm_4\}$ appears in $4$ out of $5$ transactions (\ie, $\{T_2,T_3,T_4,T_5\}$). 
The confidence value of the rule $\{\itm_1,\itm_4\}\rightarrow \itm_3$ is 
$0.5$ (\ie, $\conf(\{\itm_1,\itm_4\}\rightarrow \itm_3)=0.5$) as $50$\% of the transactions containing $\{\itm_1,\itm_4\}$ also contains 
the item $\itm_3$ (\ie, $\{T_4,T_5\}$).} 

\subsection{\textcolor{blue}{Algorithms}}
\label{subsec:pattern_algorithms}
The Apriori algorithm~\citep{DBLP:conf/vldb/AgrawalS94} is the most renowned pattern mining technique for discovering frequent itemsets and
association rules from a huge number of transactions.
It employs a breadth-first, bottom-up strategy to explore item sets.
Staring from an item, at each iteration the algorithm checks the frequency of a subset of items in the transactions with the same item set size,  and then only the ones whose support values exceed 
a predefined threshold are retained, followed by increasing the item set size by one.
The Apriori algorithm relies on the heuristic that if an item set does not
meet the threshold, none of its supersets can do so. Thus the search space 
can be  dramatically reduced. 
\textcolor{blue}{For computer vision applications, the Apriori algorithm has been used by~\cite{DBLP:conf/iccv/QuackFLG07,DBLP:conf/cvpr/YaoF10} and~\cite{DBLP:journals/pami/GilbertIB11}.}

\textcolor{blue}{There are also some other well-known pattern mining techniques, such as the FP-growth~\citep{DBLP:journals/tkde/GrahneZ05}, 
LCM~\citep{DBLP:conf/fimi/UnoAUA03}, DDPMine~\citep{DBLP:conf/icde/ChengYHY08} and KRIMP~\citep{DBLP:journals/datamine/VreekenLS11} algorithms. 
These pattern mining techniques have also been adopted in computer vision research~\citep{DBLP:conf/cvpr/YuanWY07,DBLP:conf/eccv/FernandoFT12,
DBLP:journals/ijcv/FernandoFT14,Rematas15CVPR,
DBLP:conf/iccv/FernandoT13}. %
In this work, we opt for the Apriori algorithm for pattern mining.}
\subsection{\textcolor{blue}{Challenges}}
\label{subsec:challenges}
\noindent\textbf{Transaction creation.} The process of transforming data into a set of transactions is the most crucial step in applying such pattern mining techniques for vision applications. 
Ideally, the 
representation of the data in this format should allow all of the relevant information to be represented, with no information loss.
However, as noted in~\cite{DBLP:conf/cvpr/Voravuthikunchai14}, there are two strict requirements 
of pattern mining techniques that make creating 
transactions with no information loss very challenging. 
\begin{enumerate}
\item
Each transaction can only have a small number of items, as the potential search space grows exponentially with the number of items in each transaction.
\item
What is recorded in a transaction must be a set of integers (which are typically the indices of items).
\end{enumerate}

As we will show in the next section,
thanks to two appealing properties of CNN activations (Sec.~\ref{subsec:properities}), 
these two requirements can be fulfilled effortlessly if one uses CNN activations to create transactions.\\

\noindent\textcolor{blue}{\textbf{Pattern explosion.}} Known as pattern explosion in the pattern
mining literature, the number of patterns discovered with a pattern mining
technique can be enormous,
with some of the patterns being highly correlated.
Therefore, before using patterns for applications, the first step is pattern selection, that is, to select a subset of patterns which are both discriminative and not redundant.

For the task of pattern selection, some heuristic rules are proposed in previous works. 
For instance, \textcolor{blue}{\cite{DBLP:conf/cvpr/YuanWY07} compute a likelihood ratio 
to select patterns.}
\cite{DBLP:conf/eccv/FernandoFT12,DBLP:journals/ijcv/FernandoFT14} use 
a combination of discriminativity scores and representativity scores to select patterns. 
\cite{Rematas15CVPR}, instead, propose a pattern interestingness criterion and a greedy algorithm for selecting patterns. 
\textcolor{blue}{
Instead of a two-step framework which includes pattern mining and selection, some previous works in pattern mining~\citep{DBLP:conf/icde/ChengYHY08,DBLP:journals/datamine/VreekenLS11} propose to find discriminative patterns within the pattern mining algorithm itself, thus avoid the problem of pattern explosion and relieve the need of pattern selection. 
In this work, to address the problem of pattern explosion, we advocate  merging patterns describing the same visual concept rather than selecting a subset of patterns. }

%% file: algorithm.tex
\begin{figure*}[t]
\vspace{-0.0cm}
\centering
\begin{tabular}{@{}c}
  \includegraphics[width=.85\linewidth]{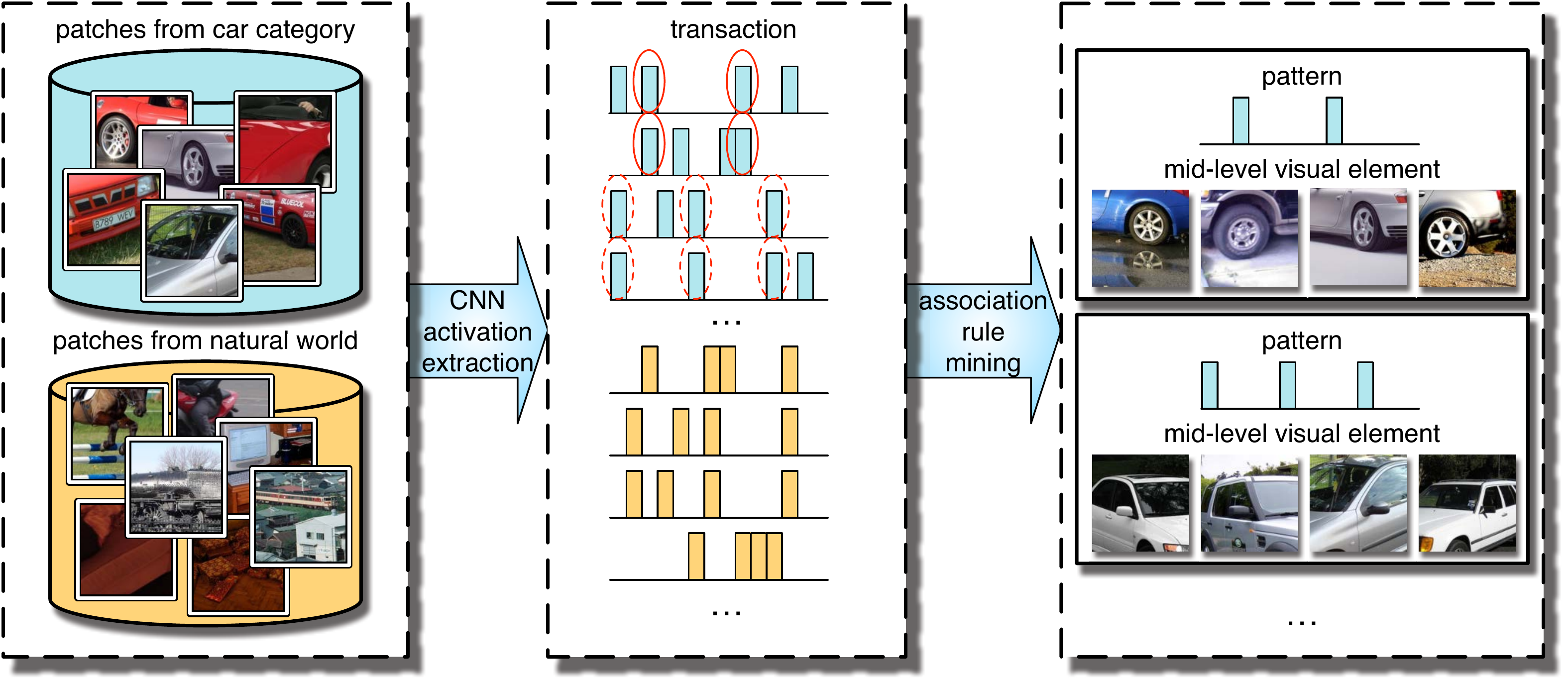} \\
\end{tabular}
\caption{An illustration of the mid-level deep pattern mining process. %
Given image patches sampled from both the target category (\eg,
car) and the 
background class
we represent each as a transaction after extracting their CNN activation.
Patterns are then discovered by the well-known association rule mining. 
Mid-level visual elements are retrieved from image patches with the same patterns.}
\label{fig:pipleline} 
\end{figure*}

An overview of the proposed the MDPM algorithm is illustrated in
Fig.~\ref{fig:pipleline}. 
Assuming that image labels are known, we start by sampling a huge number of
random patches both from images of the target category (\eg, car) and images
that do not contain the target category (\ie, the background class). %
With the two appealing properties of CNN activations of image patches
(Sec.~\ref{subsec:properities}), we then create a transaction database in which each transaction corresponds to a particular image patch
(Sec.~\ref{subsec:transaction}).
Patterns %
are then
discovered from the transaction database using association rule mining (Sec.~\ref{subsec:mining}), from which mid-level visual elements can be
retrieved efficiently (Sec.~\ref{subsec:retrieval}).

\subsection{Properties of CNN activation of patches}
\label{subsec:properities}
In this section we provide a detailed analysis of the performance of CNN activations on the MIT Indoor dataset~\cite{DBLP:conf/cvpr/QuattoniT09},
from which we are able to deduce two important properties thereof.
These two properties are critical to the suitability of such activations %
to form the basis of a transaction-based approach.

We first sample $128\times128$ patches with a stride of $32$ pixels from each image. Then, for each image patch, we extract the $4096$-dimensional non-negative output of the first fully-connected layer of \emph{BVLC Reference CaffeNet}~\cite{Jia13caffe}.
To generate image features, we consider the following three strategies.
The first strategy is our baseline, which is simply the outcome of max pooling on CNN activations of all patches in an image.
The next two strategies are variants of the baseline which are detailed as follows.

\begin{enumerate}
\item \textbf{CNN-Sparsified.} For each $4096$-dimensional CNN activation of an image patch, we 
retain the magnitudes of only the $K$ largest elements in the vector, setting 
 the remaining elements 
to zero.
The feature representation for an image is the outcome of applying max pooling to the thus revised CNN activations.

\item \textbf{CNN-Binarized.} For each $4096$-dimensional CNN activation of an image patch, we set the $K$ largest 
elements in the vector to one and the remaining elements to zero.
The feature representation for an image is the outcome of performing max pooling on these binarized CNN activations.
\end{enumerate}
For each strategy we train a multi-class linear SVM classifier in a one-vs-all fashion.
The classification accuracy achieved by each of the two above strategies for a range of $K$ values is summarized in Table~\ref{tab:top_k}.
In comparison, our baseline method gives an accuracy of $65.15$\%.
Analyzing the results in Table~\ref{tab:top_k} leads to two observations of CNN activations of fully-connected layers (expect the last classification layer):
\begin{table}[t]
\begin{center}
\begin{tabular}{l|c|c|c|c}
\hline
 $K$ & $10$ & $20$ & $50$ & $100$\\
\hline\hline
CNN-Sparsified & $50.10$ & $56.33$ & $60.34$ & $61.68$ \\
\hline
CNN-Binarized & $54.34$ & $59.15$ & $61.35$ & $61.29$ \\
\hline
\end{tabular}
\end{center}
\caption{Classification accuracies achieved by the two strategies for keeping the $K$ largest magnitudes of CNN activations of image patches on the MIT Indoor dataset. Note that our baseline, the outcome of max pooling on CNN activations of all patches in an image, gives an accuracy of $65.15$\%.
}
\vspace{-5mm}
\label{tab:top_k}
\end{table}
\begin{enumerate}
\item \textbf{Sparse.} Comparing the performance of ``CNN-Sparsified'' with that of the baseline feature ($65.15$\%), 
  it is clear that accuracy is reasonably high when using sparsified CNN activations with a small fraction of 
  non-zero magnitudes out of $4096$.
\item \textbf{Binary.} Comparing ``CNN-Binarized''  with the ``CNN-Sparsified'' counterpart, 
  it can be seen that CNN activations do not suffer from binarization when $K$ is small. Accuracy even increases slightly in some cases.
\end{enumerate}

\textcolor{blue}{Note that the above properties are also observed in recent works on analyzing CNNs~\citep{DBLP:conf/eccv/Agrawal14,dosovitskiy2015inverting}.}\\

\noindent\textbf{Conclusion.} The above two properties imply that for an image patch, the discriminative information within its CNN activation is mostly embedded in
\emph{the dimension indices of the $K$ largest magnitudes}.
\subsection{Transaction creation}
\label{subsec:transaction}
Transactions must be created before any pattern mining algorithm can proceed.
In our work, as we aim to discover patterns from image patches, 
\emph{a transaction is created for each image patch}.

The most critical issue
now is how to transform an image patch into a transaction while
retaining as much information as possible.
Fortunately the analysis above (Sec.~\ref{subsec:properities}) illustrates that CNN activations are particularly well suited to the task.
Specifically, we treat {each dimension index of a CNN activation} as an item ($4096$ items in total).
Given the performance of the binarized features shown above,
each transaction is then represented by \emph{the dimension indices of the $K$ largest elements of the corresponding image patch}.

This strategy satisfies both requirements for applying pattern mining
techniques (Sec.~\ref{sec:pattern_mining_revisited}). 
Specifically, 
given little performance is lost when using a sparse representation of CNN activations (`sparse property' in Sec.~\ref{subsec:properities}),
each transaction calculated as described contains only a small number items ($K$ is small).
And because binarization of CNN activations has little deleterious effect on classification performance (`binary property' in Sec.~\ref{subsec:properities}), most of the discriminative information within a CNN activation is retained by treating dimension indices as items.

Following the work of~\cite{DBLP:conf/iccv/QuackFLG07}, at the end of each
transaction, we add a $pos$ (or $neg$) item if the corresponding image patch comes
from the target category 
(or the background class).
Therefore, each complete transaction has $K+1$ items, consisting of the indices of the
$K$ largest elements in the CNN activation plus one class label.
For example, if we set $K=3$, given a CNN activation of an image
patch from the target category which has $3$ largest magnitudes in its 3rd,
$100$-th and $4096$-th dimensions, the corresponding transaction will be
$\{3, 100, 4096, pos\}$.

In practice,  we first sample a large number of patches from images in both the target category and the background class.
After extracting their CNN activations, a transaction database $\mathcal{D}$ is created,
containing a large number of transactions created using the proposed technique.
Note that the class labels, $pos$ and $neg$, are represented by $4097$
and $4098$ respectively in the transactions.

\begin{figure*}[t]
\vspace{-0.0cm}
\begin{center}
\scalebox{0.95}{
\begin{tabular}{@{}c@{}c@{}c@{}c@{}c}
\includegraphics[width=0.2\linewidth]{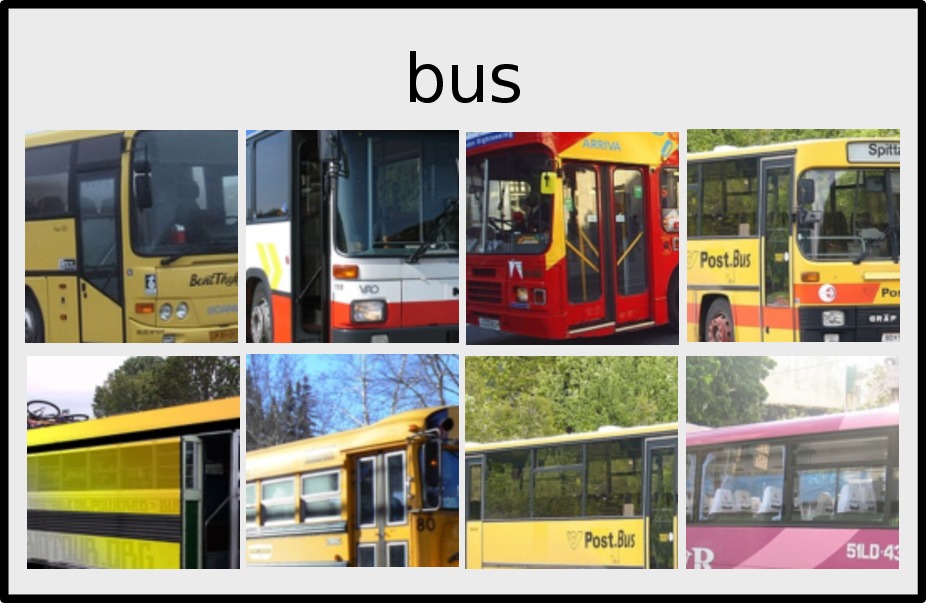} \ &
\includegraphics[width=0.2\linewidth]{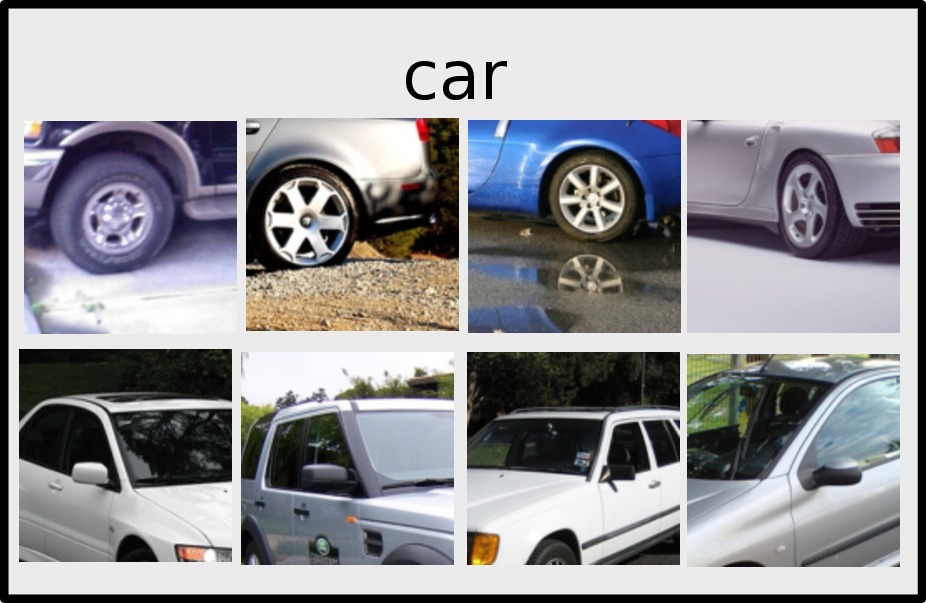} \ &
\includegraphics[width=0.2\linewidth]{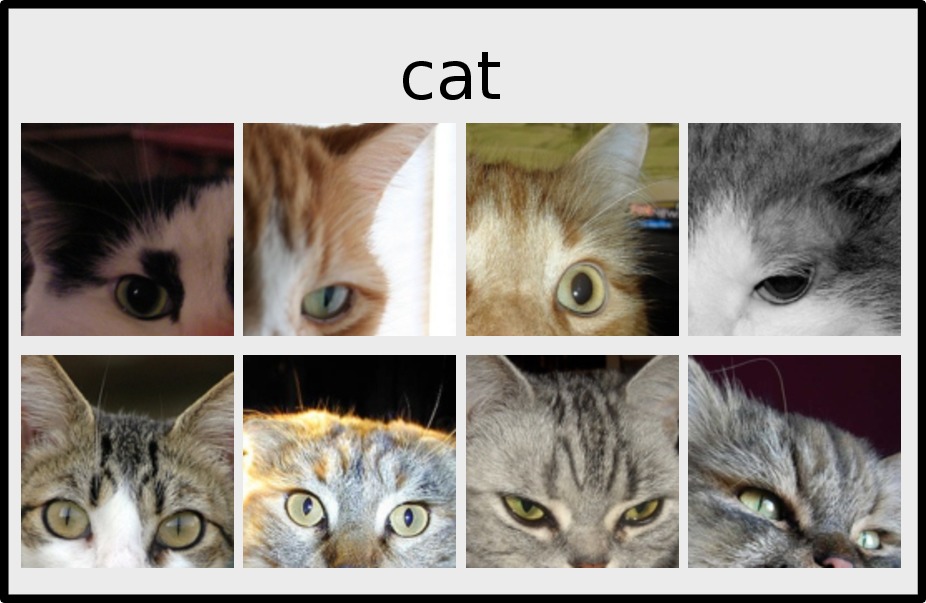} \ &
\includegraphics[width=0.2\linewidth]{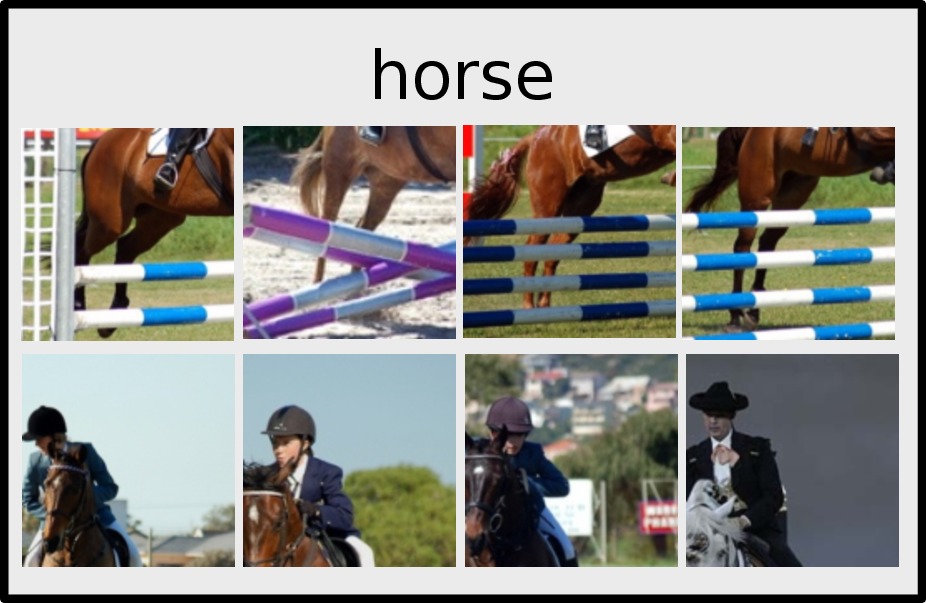} \ &
\includegraphics[width=0.2\linewidth]{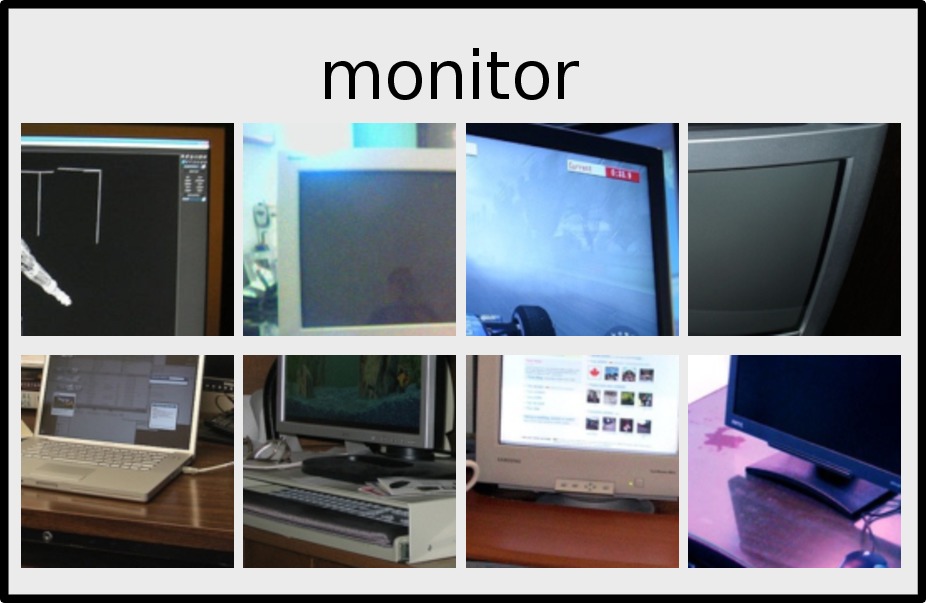} \ \\
\end{tabular}
}
\end{center}
\caption{Mid-level visual elements discovered by our algorithm on the Pascal VOC 2007 dataset (for each category, each row is one exemplar).}
\label{fig:fire_VOC}
\end{figure*}

\subsection{Mining representative and discriminative patterns}
\label{subsec:mining}
Given the transaction database $\mathcal{D}$ constructed in Sec.~\ref{subsec:transaction}, we use the Aprior algorithm~\citep{DBLP:conf/vldb/AgrawalS94} to discover a set of patterns $\mathcal{P}$ through association rule mining. More specifically, Each pattern $P \in \mathcal{P}$ 
must satisfy the following two criteria:
\begin{align}
\label{eq:support}
\supp(P)&>\supp_{\min},\\
\label{eq:confidence}
\conf(P \rightarrow pos)&>\conf_{\min},
\end{align}
where $\supp_{\min}$ and $\conf_{\min}$ are thresholds for the support value and confidence. \vspace{0.2cm}
\newline
\textbf{Representativeness and discriminativeness.} We now demonstrate how association rule mining implicitly satisfies the two requirements of mid-level visual element discovery, \ie, representativeness and discriminativeness.
Specifically, based on Eq.~\eqref{eq:support} and Eq.~\eqref{eq:confidence},
we are able to rewrite Eq.~\eqref{eq:confidence_definition} thus
\begin{equation}
\begin{split}
\supp(P\cup\{pos\})& = \supp(P)\times \conf(P \rightarrow pos)\\
                       & > \supp_{\min}\times \conf_{\min},
\end{split}
\end{equation}
where $\supp(P\cup\{pos\})$ measures the fraction of pattern $P$ found in transactions of the target category among all the transactions.
Therefore, having values of $\supp(P)$ and $\conf(P\rightarrow pos)$ larger than their thresholds ensure that
pattern $P$ is found frequently in the target category, akin to the representativeness requirement.
A high value of $\conf_{\min}$ (Eq.~\eqref{eq:confidence})  also ensures that pattern $P$ is more likely to be found in the target category rather than the background class, reflecting
the discriminativeness requirement.

\subsection{Retrieving mid-level visual elements}
\label{subsec:retrieval}

Given the set of patterns $\mathcal{P}$ discovered in Sec.~\ref{subsec:mining},
finding mid-level visual elements is straightforward. 
A mid-level visual element $V$ contains the image patches sharing the same pattern $P$, 
which can be retrieved efficiently through an inverted index. 
This process outputs a set of mid-level visual elements $\mathcal{V}$ (\ie, $V \in \mathcal{V}$).

We provide a visualization of some of the discovered mid-level visual elements in Fig.~\ref{fig:fire_VOC}. 
It is clear that image patches in each visual element are visually similar and depicting the same semantic concept while being discriminative from other categories.
For instance, some mid-level visual elements catch discriminative parts 
of objects (\eg, cat faces found in the \emph{cat} category), and some depict
typical objects or people in a category (\eg, horse-rider found in the \emph{horse} category).
An interesting observation is that mid-level elements discovered by the proposed MDPM 
algorithm are invariant to horizontal flipping.
This is due to the fact that original images 
and their horizontal flipping counterparts are fed into the CNN during the pre-training 
process. 

%% file: encoding1.tex
To discover patterns from a dataset containing $Y$ categories, each category is
treated as the target category while all remaining $Y-1$ categories in the
dataset are treated as the background class.
Thus $Y$ sets of patterns will be discovered by the MDPM algorithm, one for
each of the $Y$ categories.
Given the $Y$ sets of patterns and retrieved mid-level visual elements, we
propose two methods to generate image feature representations.
The first method is to use a subset of patterns (Sec.~\ref{subsec:pattern_encoding}),
whereas the second one relies on the
retrieved mid-level visual elements (Sec.~\ref{subsec:mid_encoding}).
The details of both methods are as follows.

\subsection{Encoding an image using patterns}
\label{subsec:pattern_encoding}

\begin{figure*}[t]
\vspace{-0.0cm}
\centering
\begin{tabular}{@{}c}
\includegraphics[width=.61\linewidth]{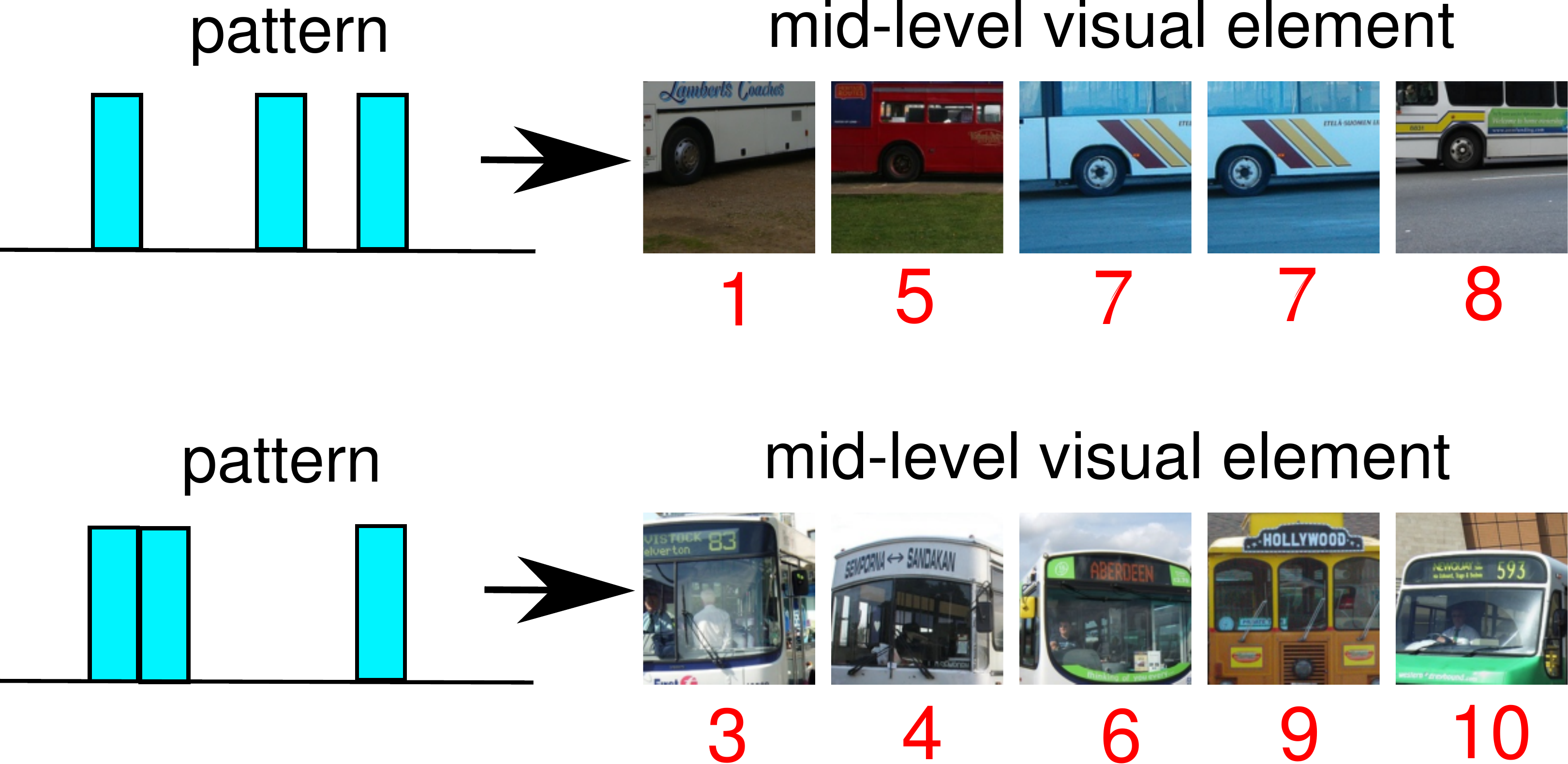} \\
\end{tabular}
\caption{ \textcolor{blue}{An illustration of the pattern selection process (Sec.~\ref{subsubsec:pattern_Sselection}). For each pattern on the left, the image patches which form the
corresponding mid-level visual elements are shown on the right.
The red number underneath each patch is the image index. Since the top and bottom pattern cover $4$ and $5$ unique images, the coverage values of them are $4$ and $5$ respectively.}}
\label{fig:pattern_sel}
\end{figure*}

\subsubsection{Pattern selection}
\label{subsubsec:pattern_Sselection}

Due the problem of pattern explosion (Sec.~\ref{subsec:challenges}),
we first select a subset of the discovered patterns based on a simple
criterion.
We define the \emph{coverage} of a pattern and its retrieved mid-level visual element as the number of unique images that image patches in this element comes from (\textcolor{blue}{see Fig.~\ref{fig:pattern_sel} for an intuitive example}).
Then, we  rank the patterns using the proposed \emph{coverage} criterion. %
The intuition here is that
we aim to find the patterns whose corresponding mid-level elements cover as many different images as possible, resembling the ``Purity-Coverage Plot'' in~\cite{DBLP:conf/nips/DoerschGE13}.
Thus, from each category, we select $X$ patterns whose corresponding mid-level elements have top-$X$ coverage values.
Then, the selected patterns from all $Y$ categories are combined into a new set of patterns $\hat{\mathcal{P}}$ which contains $X \times Y$ elements in total. %

\subsubsection{Bag-of-Patterns representation}
\label{subsubsec:BoP}
To encode
a new image using a set of patterns $\hat{\mathcal{P}}$, we first
sample image patches at multiple scales and locations,
and extract their CNN activations.
For each 4096-dimensional CNN activation vector of an image patch, after finding $C_i$, the set of indices of dimensions that have non-zero values,
we check for each selected pattern $P_k \in \hat{\mathcal{P}}$ whether $P \subseteq C_i$.
Thus, our Bag-of-Patterns representation (BoP for short) $f_{BoP} \in \mathbf{R}^{X \times Y}$ is a histogram encoding of the set of local CNN activations, satisfying $[f_{BoP}]_k = |\{i | P_k \in C_i \}|$.

Our Bag-of-Patterns representation is similar to the well-known
Bag-of-Visual-Words (BoW) representation~\citep{DBLP:conf/iccv/SivicZ03} if one
thinks of a pattern $P \in \hat{\mathcal{P}}$ as one visual word.
The difference is that in the BoW model
one local descriptor is typically  assigned to one visual word,
whereas in our BoP representation, multiple patterns can fire given on the basis of a CNN activation (and thus image patch).
\textcolor{blue}{Note that BoP representation has also been utilized by
~\cite{DBLP:conf/iccv/FernandoT13} for image retrieval.}
In practice, we also add a $2$-level ($1\times1$ and $2\times2$) spatial pyramid~\citep{DBLP:conf/cvpr/LazebnikSP06} when computing the BoP representation.
More specifically, to generate the final feature representation, we concatenate the normalized BoP representations extracted from different spatial cells.

\subsection{Encoding an image using mid-level elements}
\label{subsec:mid_encoding}
Due to the redundant nature of the discovered patterns,
mid-level visual elements
retrieved from those patterns are also likely to be redundant.

For the purpose of removing this redundancy, we merge mid-level elements that are both visually
similar and which depict the same visual concept (Sec. \ref{subsubsec:merging_detectors}).
Patch detectors trained from the merged mid-level elements can then be used to construct
a Bag-of-Elements representation (Sec.~\ref{subsubsec:Bag-of-Elements}).

\begin{algorithm}[t]
\label{alg:merging}

\SetKwFunction{Grow}{Grow}
\SetKwFunction{MergingTrain}{MergingTrain}
\SetKwFunction{Coverage}{Coverage}
\SetKwFunction{Score}{Score}
\caption{Ensemble Merging Pseudocode}
\KwIn{A set of partially redundant visual elements $\mathcal{V}$}
\KwOut{A set of clean mid-level visual elements $\mathcal{V}'$ and corresponding patch detectors $D$}
Initialize $\mathcal{V}' \leftarrow \emptyset$, $D_c \leftarrow \emptyset$\;
\While{$\mathcal{V} \neq \emptyset$}{
$[\mathcal{V}_t,d] \leftarrow \MergingTrain(\mathcal{V})$\;
$\mathcal{V} \leftarrow \mathcal{V} \setminus \mathcal{V}_t$\;
$\mathcal{V}' \leftarrow \mathcal{V}' \cup \{ \mathop{\cup}\limits_{V \in \mathcal{V}_t} V\}$\;
$D \leftarrow D \cup \{d\}$\;
}
\KwRet $\mathcal{V}'$, $D$\;
\BlankLine
\SetKwProg{myFunc}{Function}{}{}
\myFunc{\MergingTrain{$\mathcal{V}$}}{
  Select $V^{\ast} \in \mathcal{V}$ which covers the maximum number of training images\;
  Initialize $\mathcal{V}_t \leftarrow \{V^{\ast}\}$, $\mathcal{S} \leftarrow \emptyset$ \;

	\Repeat{$\mathcal{S} = \emptyset$}{
  $\mathcal{V}_t \leftarrow \mathcal{V}_t \cup \mathcal{S}$\;
  Train LDA detector $d$ using $\mathcal{V}_t$\;
  $\mathcal{S} \leftarrow \{V \in \mathcal{V}\setminus\mathcal{V}_t | \Score(V,d)>Th$\} where
  $\Score(V,d) = \frac{1}{|V|}\sum_{x \in V} d^Tx$ ($Th$ is a pre-defined threshold)\;
	}
  \KwRet $\mathcal{V}_t$, $d$\;
}
\end{algorithm}

\subsubsection{Merging mid-level elements}
\label{subsubsec:merging_detectors}
We propose to merge mid-level elements while simultaneously training corresponding detectors using an iterative approach.

Algorithm~\ref{alg:merging} summarizes the proposed ensemble merging procedure.
At each iteration, we greedily merge overlapping mid-level elements and train the corresponding detector through the \texttt{MergingTrain} function in Algorithm~\ref{alg:merging}.
In the \texttt{MergingTrain} function, we begin by selecting the element covering the maximum number of training images, and then train a Linear Discriminant Analysis (LDA) detector~\citep{DBLP:conf/eccv/HariharanMR12}. The LDA detector has the advantage that it can be computed efficiently using a closed-form solution $\Sigma^{-1}(\bar{x}_p-\bar{x})$
 where $\bar{x}_p$ is the mean of CNN activations of positive samples, $\bar{x}$ and $\Sigma$ are the mean and covariance matrix respectively which are estimated from a large set of random CNN activations.
\textcolor{blue}{Inspired by previous works~\citep{DBLP:conf/cvpr/JunejaVJZ13,DBLP:conf/iccv/LeeEM13,
DBLP:conf/eccv/SinghGE12}}, We then incrementally revise this detector.
At each step, we run the current detector on the
activations
of all the remaining mid-level elements, and retrain it by augmenting the positive training set with positive detections. We repeat this iterative procedure until no more elements can be added into the positive training set.
The idea behind this process is using the
detection score as a similarity metric,
inspired by Exemplar SVM~\citep{DBLP:conf/iccv/MalisiewiczGE11, DBLP:journals/tog/ShrivastavaMGE11}.
The output of the ensemble merging step is a merged set of mid-level elements and their corresponding
detectors.
\textcolor{blue}{The limitation of the proposed merging method is that the merging threshold $Th$ (see Algorithm~\ref{alg:merging}) needs to be tuned, which will be analyzed in the experiment (Sec.~\ref{subsubsec:ablation_study}).}

After merging mid-level elements, we again use the coverage criterion (Sec.~\ref{subsubsec:pattern_Sselection}) to select $X$
detectors of merged mid-level elements for each of the $Y$ categories and stack them together.

\subsubsection{Bag-of-Elements representation}
\label{subsubsec:Bag-of-Elements}
\begin{figure}[t]
\vspace{-0.0cm}
\centering
\begin{tabular}{@{}c}
\includegraphics[width=0.8\linewidth]{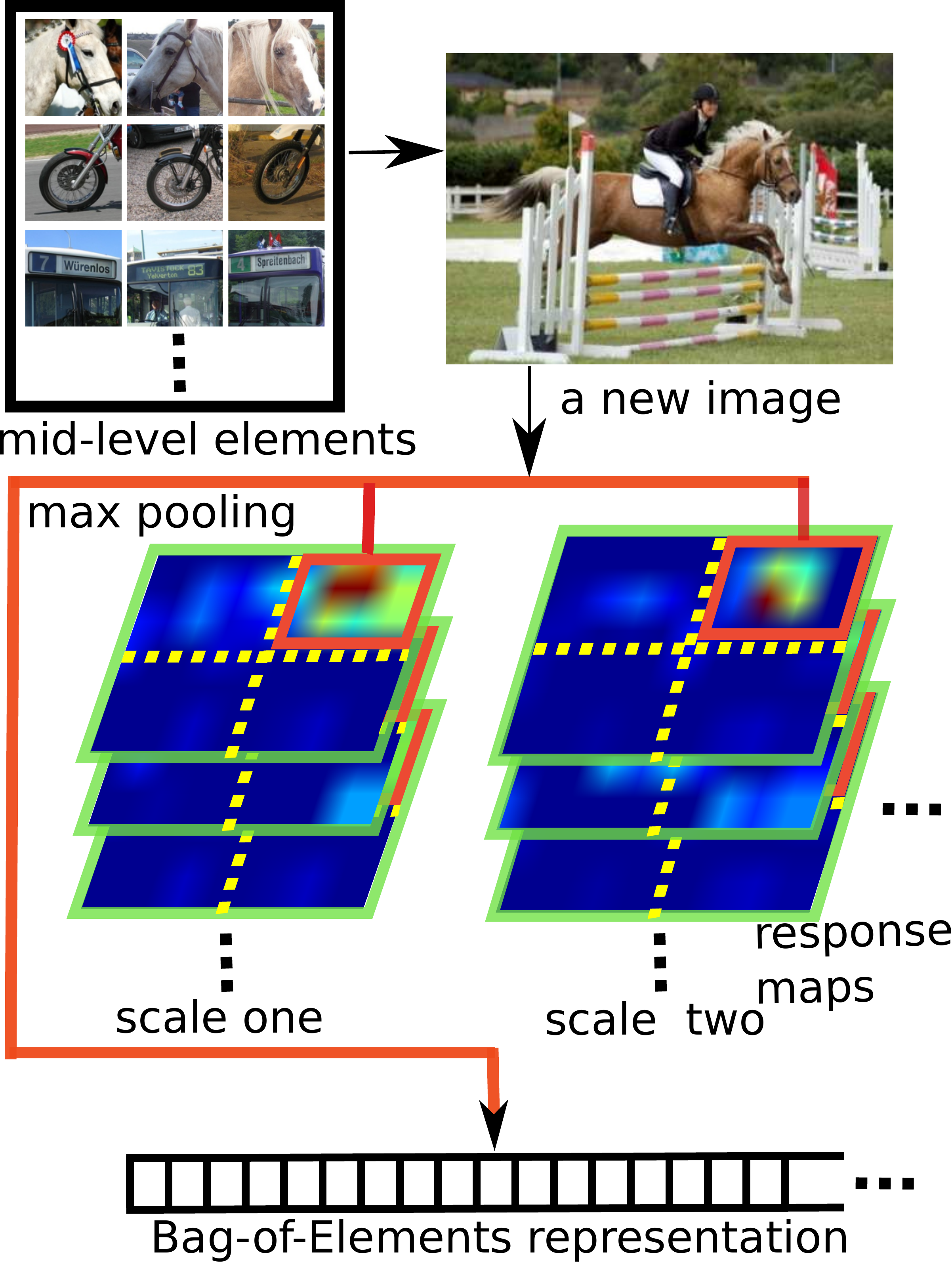} \\
\end{tabular}
\caption{Pipeline to construct a Bag-of-Elements representation,  \textcolor{blue}{which has been used in previous works as well~\citep{DBLP:conf/eccv/SinghGE12,DBLP:conf/nips/DoerschGE13,
DBLP:conf/cvpr/JunejaVJZ13,bansal2015mid}.}}
\label{fig:encoding} \vspace{-0.39cm}
\end{figure}

As shown in previous works on mid-level visual element discovery~\citep{DBLP:conf/eccv/SinghGE12,DBLP:conf/cvpr/JunejaVJZ13,
DBLP:conf/nips/DoerschGE13,bansal2015mid}, detectors of mid-level elements can be utilized to generate a Bag-of-Elements representation.
An illustration of this process is shown in Fig.~\ref{fig:encoding}.
Concretely, given an image, we evaluate each of the detectors at multiple scales, which results in a stack of response maps of
detection scores.
For each scale, we take the max score per detector per region encoded in a $2$-level ($1\times1$ and $2\times2$) spatial pyramid.
The final feature representation of an image has $X \times Y \times 5$ dimensions, which is the outcome of max pooling on the responses from all scales in each spatial cell.

%% file: experiments.tex
\textcolor{blue}{
This section contains an extensive set of experimental
result and summarizes the main findings.
Firstly, some general experimental setups (\eg, datasets, implementation details) are discussed
in Sec.~\ref{subsec:setup}, followed by detailed analysis of the
proposed approach on object (Sec.~\ref{subsec:object}) and indoor scene (Sec.~\ref{subsec:scene})  classification tasks respectively.
Rely on the discovered mid-level visual elements. Sec.~\ref{sec:context} provides further analysis of the importance of context information for recognition, which seldom appears in previous works on mid-level elements.   }

\subsection{Experimental setup}
\label{subsec:setup}
\subsubsection{CNN models}
For extracting CNN activations from image patches, we consider two state-of-the-art CNN models which are both pre-trained on the  ImageNet dataset~\citep{DBLP:conf/cvpr/DengDSLL009}.
The first CNN model is the \emph{BVLC Reference CaffeNet}~\citep{Jia13caffe} (\emph{CaffeRef} for short), whose architecture is \textcolor{blue}{similar to} that of AlexNet~\citep{DBLP:conf/nips/KrizhevskySH12}, that is, five convolution layers followed by two 4096-dimensional and one 1000-dimensional fully-connected layers.
The second CNN model is the 19-layer \emph{VGG-VD} model~\citep{Simonyan15ICLR} which has shown good performance in the ILSVRC-2014 competition~\citep{DBLP:journals/corr/RussakovskyDSKSMHKKBBF14}.
For both models, we extract the non-negative 4096-dimensional activation from the first fully-connected layer after the rectified linear unit (ReLU) transformation as image patch representations.

\subsubsection{Datasets}
We evaluate our approach on three publicly
available image classification datasets, two for generic object classification and the other for scene classification. The details of the datasets are as follows.

\vspace{1mm}
\noindent\textbf{Pascal VOC 2007 dataset}.
The Pascal VOC 2007 dataset~\citep{DBLP:journals/ijcv/EveringhamGWWZ10,DBLP:journals/ijcv/EveringhamEGWWZ15} contains a total of $9,963$ images from $20$ object classes, including
$5,011$ images for training and validation, and $4,952$ for testing.
For evaluating different algorithms, mean average precision (mAP) is adopted as the standard quantitative measurement.

\vspace{1mm}
\noindent\textbf{Pascal VOC 2012 dataset}.
The Pascal VOC 2012 dataset~\citep{DBLP:journals/ijcv/EveringhamGWWZ10,DBLP:journals/ijcv/EveringhamEGWWZ15} is
an extension of the VOC 2007 dataset, which contains a total of $22,531$ images from $20$ object classes, including $11,540$ images for training and validation, and $10,991$ for testing.
We use the online evaluation server of this dataset to evaluate
the proposed approach.

\vspace{1mm}
\noindent\textbf{MIT Indoor dataset}.
The MIT Indoor dataset~\citep{DBLP:conf/cvpr/QuattoniT09} contains 67 classes of indoors scenes.
A characteristic of indoor scenes is that unique configurations or objects are often found in a particular scene, \eg, computers are more likely to be found in a computer room rather than a laundry.
For this reason, many  mid-level element discovery algorithms~\citep{DBLP:conf/eccv/SinghGE12,DBLP:conf/cvpr/JunejaVJZ13,
DBLP:conf/nips/DoerschGE13,DBLP:conf/iccv/SunP13,DBLP:conf/eccv/BossardGG14} are evaluated on this dataset and have achieved state-of-the-art performance.
We follow the standard partition of~\citep{DBLP:conf/cvpr/QuattoniT09}, \ie, approximately 80 training and 20 test images per class.
The evaluation metric for MIT Indoor dataset is the mean classification
accuracy.

\subsubsection{Implementation details}
Given an image, we resize its smaller dimension to $256$ while maintaining its aspect ratio, then we sample $128\times128$ patches with a stride of $32$ pixels, and calculate the CNN activations from Caffe (using either the \emph{CaffeRef} or \emph{VGG-VD} models).
When mining mid-level visual elements, only training images are used to create transactions
(\texttt{trainval} set for Pascal VOC datasets).
The length of each is transaction is set as $20$, which corresponds to $20$ largest dimension indices of CNN activations of an image patch.
\textcolor{blue}{We use the implementation of association rule mining from
\cite{DBLP:journals/widm/Borgelt12}}\footnote{\url{http://www.borgelt.net/apriori.html}}.
The merging threshold $Th$ in Algorithm~\ref{alg:merging} (Sec.~\ref{subsubsec:merging_detectors}) is set as $150$.
For generating image features for classification,
CNN activations are extracted from five scales for the Pascal VOC datasets as compared to three scales for the MIT Indoor dataset (we experimentally found
using more than three scales for MIT Indoor does not improve the overall classification performance.
)  .
For training image classifiers, we use the Liblinear toolbox~\citep{DBLP:journals/jmlr/FanCHWL08}
with $5$-fold cross validation.
For association rule mining,
the value of $supo_{min}$ (Eq.~\ref{eq:support}) is always set as $0.01$\% whereas the value of
$conf_{min}$ (Eq.~\ref{eq:confidence}) is tuned for different datasets.

\subsection{Object classification}
\label{subsec:object}

In this section, we provide a detailed analysis of the proposed system for object classification on the Pascal VOC
2007 and 2012 datasets.
We begin with an ablation study which illustrates the importance of the
different components of our system (Sec.~\ref{subsubsec:ablation_study}).
In Sec.~\ref{subsubsec:state-of-the-art}, we compare our system with state-of-the-art algorithms which also rely on CNNs, followed by computational complexity analysis in Sec.~\ref{subsubsec:complexity}.
Some visualizations of mid-level visual elements are provided in Sec.~\ref{subsubsec:visualization}.
On VOC 2007 dataset, the $\conf_{\min}$ (Eq.~\ref{eq:confidence}) is set as $60\%$ for \emph{CaffRef} and $80\%$ for \emph{VGG-VD} model respectively.
On VOC 2012 dataset, we use $40\%$ for
$\conf_{\min}$ when \emph{VGG-VD} model is adopted.
\begin{figure}[t]
\vspace{-0.0cm}
\centering
\begin{tabular}{@{}c}
\includegraphics[width=0.7\linewidth]{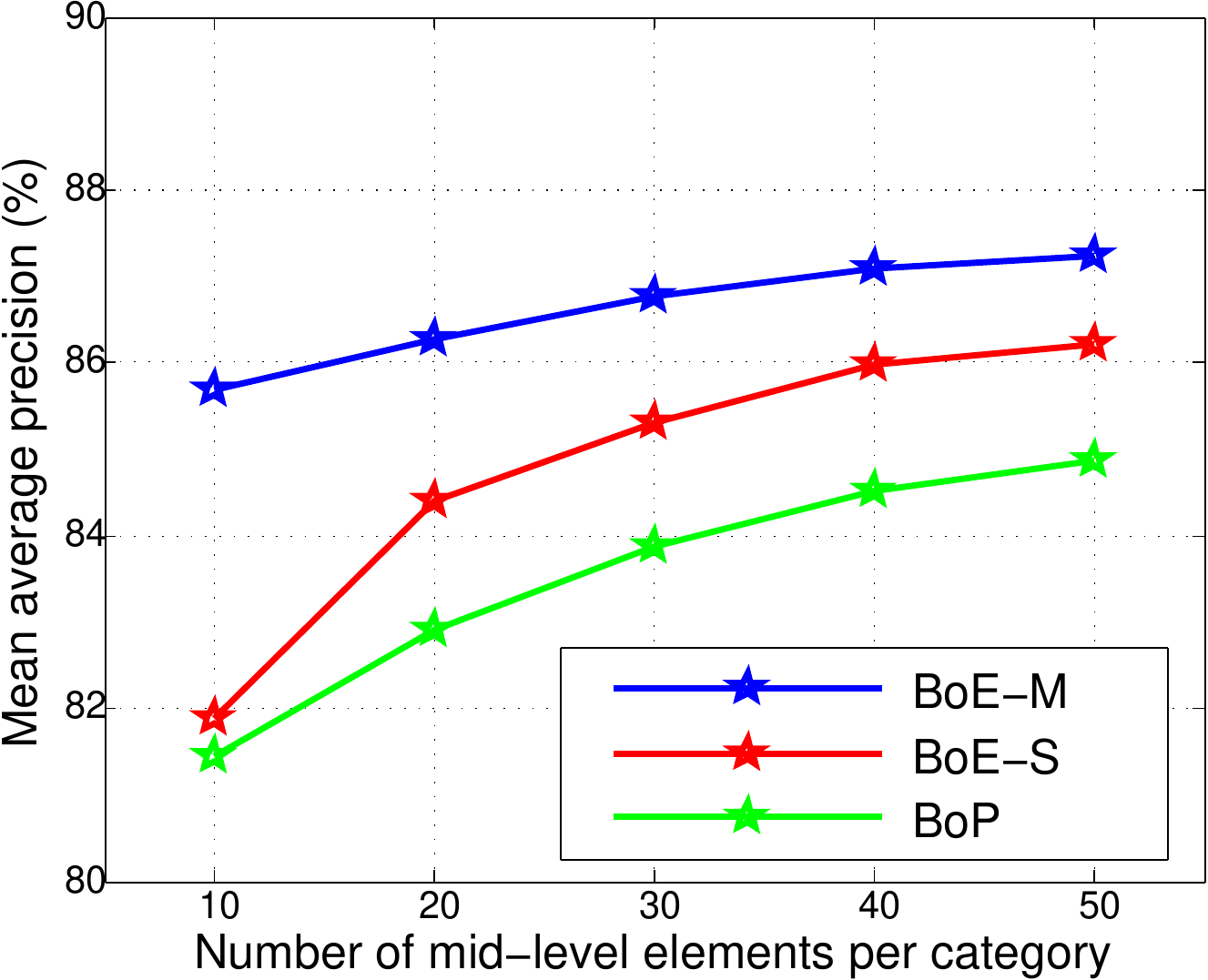} \\
\ (a)\\
\ (b)\\
\end{tabular}
\caption{Performance of proposed feature encoding methods on the Pascal VOC 2007 dataset. Note that \emph{VGG-VD} model is used for evaluation.}
\label{fig:ablation_VOC}
\end{figure}

\subsubsection{Ablation study}
\label{subsubsec:ablation_study}
\noindent\textcolor{blue}{\textbf{Bag-of-Elements vs. Bag-of-Patterns}}. We analyze the performance achieved by different encoding methods proposed in Sec.~\ref{sec:encoding}.
We denote the the Bag-of-Patterns representation as BoP, and the Bag-of-Elements representation constructed after the merging process as BoE-M.
We also implement another encoding method, BoE-S which does not merge mid-level elements but rather
select mid-level elements from a large pool of candidates using the coverage criterion. %
The performance of the above encoding methods are illustrated in
Fig.~\ref{fig:ablation_VOC}.

As is illustrated in Fig.~\ref{fig:ablation_VOC}, when using the same number of mid-level elements and the same CNN model, the Bag-of-Elements representation significantly outperforms the Bag-of-Patterns representation.
This could be interpreted as resulting from the ``hard-assignment'' process at the heart of the Bag-of-Patterns method.
In contrast, Bag-of-Elements does  not suffer from this problem because it
relies on the detection responses of the patch detectors.
Compared with direct selection of mid-level elements,
performance is consistently boosted when mid-level elements
are first merged (BoE-M \vs ~BoE-S), which shows
the importance of the proposed merging algorithm (\cf ~Algorithm~\ref{alg:merging}).
Therefore, we use our best encoding method, BoE-M, to compare with the
state-of-the-art below (note that the suffix is dropped).\\

\noindent\textcolor{blue}{\textbf{Number of mid-level elements.}} Irrespective of the CNN architecture or encoding method, adding
more mid-level elements or patterns to construct image features consistently improves classification accuracy (see Fig.~\ref{fig:ablation_VOC}).
Note also that the performance gain is large when a small
number of mid-level elements (patterns) are used (\eg, from $10$ to $20$), and seems to saturate when the number of mid-level elements reaches $50$.
This is particularly interesting given the differences between the datasets and the CNN networks used.\\

\noindent\textcolor{blue}{
\textbf{Transaction length.} We evaluate the performance of our approach under three settings of the transaction length, which are $10$, $20$ and $30$ respectively. Table~\ref{tab:transaction_length} depicts the results.
It is clear from Table~\ref{tab:transaction_length} that more information will be lost when using a smaller transaction length. However, as the search space of the association rule mining algorithm grows exponentially with the transaction length, this value cannot be set very large or otherwise it becomes both time and memory consuming. Therefore, we
opt for $20$ as the default setting for transaction length as a tradeoff between performance and time efficiency}.\\

\begin{table}[t]
\begin{center}
\begin{tabular}{l|c|c|c}
\hline
\textcolor{blue}{transaction length} & \textcolor{blue}{$10$} & \textcolor{blue}{$20$} & \textcolor{blue}{$30$}\\
\hline\hline
\textcolor{blue}{mAP ($\%$)} & \textcolor{blue}{$85.4$} & \textcolor{blue}{$87.3$} & \textcolor{blue}{$87.6$}\\
\hline
\end{tabular}
\end{center}
\caption{\textcolor{blue}{Analysis of the transaction length on the VOC 2007 dataset using the \emph{VGG-VD} model. Other parameters are frozen.}}
\label{tab:transaction_length}
\end{table}

\begin{table}[t]
\begin{center}
\begin{tabular}{l|c|c|c|c}
\hline
\textcolor{blue}{$Th$} & \textcolor{blue}{$50$} & \textcolor{blue}{$100$} & \textcolor{blue}{$150$} & \textcolor{blue}{$200$}\\
\hline\hline
\textcolor{blue}{mAP ($\%$)} & \textcolor{blue}{$86.4$} & \textcolor{blue}{$87.2$} & \textcolor{blue}{$87.3$} & \textcolor{blue}{$87.0$} \\
\hline
\end{tabular}
\end{center}
\caption{\textcolor{blue}{Analysis of the merging threshold $Th$ in Algorithm~\ref{alg:merging} on the VOC 2007 dataset using the \emph{VGG-VD} model. Other parameters are frozen.}}
\label{tab:th}
\end{table}

\begin{table*}[t]
\scriptsize{
\setlength{\tabcolsep}{3pt}
\def\arraystretch{1.2}
\center
\begin{tabular}{l@{\hspace{0.6em}}c@{\hspace{0.6em}}c@{\hspace{0.6em}}c@{\hspace{0.6em}}c
@{\hspace{0.6em}}c@{\hspace{0.6em}}c@{\hspace{0.6em}}c@{\hspace{0.6em}}c
@{\hspace{0.6em}}c@{\hspace{0.6em}}c@{\hspace{0.6em}}c@{\hspace{0.6em}}c
@{\hspace{0.6em}}c@{\hspace{0.6em}}c@{\hspace{0.6em}}c@{\hspace{0.6em}}c
@{\hspace{0.6em}}c@{\hspace{0.6em}}c@{\hspace{0.6em}}c@{\hspace{0.6em}}c | c}
\hline
\textbf{VOC 2007 test} & aero  &   bike &  bird & boat &  bottle  &  bus  &  car  &  cat  &  chair & cow & table &  dog  & horse & mbike & person  & plant & sheep & sofa & train & tv & \textbf{mAP}\\
\hline
FC (\emph{CaffeRef}) & $87.2$ & $78.1$ & $79.4$ & $79.3$ & $25.4$ & $63.3$ &
$81.2$ & $75.0$ & $46.6$ & $53.7$ & $61.8$ & $65.1$ & $82.8$ & $70.7$ &
$88.5$ & $39.4$ & $69.2$ & $51.1$ & $87.5$ & $60.0$ & $67.3$\\
FC (\emph{VGG-VD}) & $95.2$ & $85.5$ & $89.9$ & $88.1$ & $45.9$ & $81.7$ & $89.3$ & $88.6$ &
$55.2$ & $73.2$ & $75.1$ & $84.3$ & $91.3$ & $87.0$ & $92.2$ & $48.4$ & $80.1$ & $70.9$ &
$95.2$ & $74.1$ & $79.6$\\
\hline
\cite{6910029} & $88.5$ & $81.0$ & $83.5$ & $82.0$ & $42.0$ & $72.5$ & $85.3$ & $81.6$ & $59.9$ & $58.5$ & $66.5$ & $77.8$ & $81.8$ & $78.8$ & $90.2$ & $54.8$ & $71.1$ & $62.6$ & $87.4$ & $71.8$ & $73.9$\\
\cite{DBLP:conf/nips/LiuSWHW14} & $89.5$ & $84.1$ & $83.7$ & $83.7$ & $43.9$ & $76.7$ & $87.8$ & $82.5$ & $60.6$ & $69.6$ & $72.0$ & $77.1$ & $88.7$ & $82.1$ & $94.4$ & $56.8$ & $71.4$ & $67.7$ & $90.9$ & $75.0$ & $76.9$\\

\cite{DBLP:conf/bmvc/ChatfieldSVZ14} & $95.3$ & $90.4$ & $92.5$ & $89.6$ & $54.4$ & $81.9$ & $91.5$ & $91.9$ & $64.1$ & $76.3$ & $74.9$ & $89.7$ & $92.2$ & $86.9$ & $95.2$ & $60.7$ & $82.9$ & $68.0$ & $95.5$ & $74.4$ & $82.4$\\
\cite{DBLP:conf/cvpr/OquabBL14} & $88.5$ & $81.5$ & $87.9$ & $82.0$
& $47.5$ & $75.5$ & $90.1$ & $87.2$ & $61.6$ & $75.7$ & $67.3$ & $85.5$ & $83.5$ & $80.0$ &
\boldsymbol{$95.6$} & $60.8$ & $76.8$ & $58.0$ & $90.4$ & $77.9$ & $77.7$\\
\cite{DBLP:journals/pami/HeZRS15} & $91.9$ & $88.6$ & $91.2$ & $89.5$ & $63.0$ & $81.8$ & $88.7$ & $90.1$ & $62.7$ & $79.6$ & $72.8$ & $88.7$ & $90.0$ & $85.8$ & $93.5$ & $63.8$ & $88.4$ & $68.1$ & $92.1$ & $78.7$ & $82.4$\\
\cite{DBLP:journals/corr/WeiXHNDZY14} & $95.1$ & $90.1$ & $92.8$ &
$89.9$ & $51.5$ & $80.0$ & $91.7$ & $91.6$ & $57.7$ & $77.8$ & $70.9$ & $89.3$ & $89.3$ &
$85.2$ & $93.0$ & $64.0$ & $85.7$ & $62.7$ & $94.4$ & $78.3$ & $81.5$\\
\cite{DBLP:conf/cvpr/WangYMHHX16} & $96.7$ & $83.1$ & $94.2$ & \boldsymbol{$92.8$} &
$61.2$ & $82.1$ & $89.1$ & \boldsymbol{$94.2$} & $64.2$ & \boldsymbol{$83.6$} & $70.0$ &
$92.4$ & $91.7$ & $84.2$ & $93.7$ & $59.8$ & \boldsymbol{$93.2$} & $75.3$ & \boldsymbol{$99.7$} & $78.6$ & $84.0$\\
\cite{DBLP:journals/ijcv/CimpoiMKV16} & $91.3$ & $90.5$ & $91.3$ & $88.9$ &
\boldsymbol{$66.4$} & $85.6$ & $91.1$ & $90.7$ & \boldsymbol{$71.3$} & $79.8$ & $82.8$ & $90.1$ &
$90.8$ & $88.6$ &
$94.7$ & $67.7$ & $83.5$ & \boldsymbol{$78.6$} & $92.9$ & $82.2$ & $84.9$\\
\cite{Simonyan15ICLR} & - & - & - & - & - & - & - & - & - & - & - & - & - & - & - & - & - & - & - & - & \boldsymbol{$89.3$} \\
\hline
BoE (\emph{CaffeRef},50) & $90.3$ & $85.4$ & $82.9$ & $79.8$ & $45.9$ & $75.5$ &
$89.6$ & $85.1$ & $61.6$ & $60.0$ & $71.2$ & $79.9$ & $88.9$ & $83.4$ & $94.2$ & $53.3$ & $65.9$ & $67.2$ & $91.4$ & $76.0$ & $76.4$\\
BoE (\emph{VGG-VD},50) & \boldsymbol{$97.2$} & \boldsymbol{$93.3$} & \boldsymbol{$95.0$} & $91.3$ & $63.3$ & \boldsymbol{$88.2$} & \boldsymbol{$93.0$} & $94.1$ & $70.5$ & $79.9$ & \boldsymbol{$85.6$} & \boldsymbol{$93.2$} & \boldsymbol{$94.4$} & \boldsymbol{$90.4$} &
$95.4$ & \boldsymbol{$70.1$} & $87.7$ & $78.3$ & $97.2$ & \boldsymbol{$87.0$} & $87.3$\\
\hline
\end{tabular}
\vspace{3pt}
\caption{Comparison of classification results on the Pascal VOC 2007 dataset. For the sake of fair comparison, CNN models of all above methods are trained using the dataset used in the ILSVRC competition~\citep{DBLP:journals/corr/RussakovskyDSKSMHKKBBF14}, \ie, 1000 classes from the ImageNet~\citep{DBLP:conf/cvpr/DengDSLL009}.}
\label{tab:voc_2007}
}
\vspace{-0.1cm}
\end{table*}

\begin{table*}[t]
\scriptsize{
\setlength{\tabcolsep}{3pt}
\def\arraystretch{1.2}
\center
\begin{tabular}{l@{\hspace{0.6em}}c@{\hspace{0.6em}}c@{\hspace{0.6em}}c@{\hspace{0.6em}}c
@{\hspace{0.6em}}c@{\hspace{0.6em}}c@{\hspace{0.6em}}c@{\hspace{0.6em}}c
@{\hspace{0.6em}}c@{\hspace{0.6em}}c@{\hspace{0.6em}}c@{\hspace{0.6em}}c
@{\hspace{0.6em}}c@{\hspace{0.6em}}c@{\hspace{0.6em}}c@{\hspace{0.6em}}c
@{\hspace{0.6em}}c@{\hspace{0.6em}}c@{\hspace{0.6em}}c@{\hspace{0.6em}}c | c}
\hline
\textbf{VOC 2012 test} & aero  &   bike &  bird & boat &  bottle  &  bus  &  car  &  cat  &  chair & cow & table &  dog  & horse & mbike & person  & plant & sheep & sofa & train & tv & \textbf{mAP}\\
\hline
FC (\emph{VGG-VD}) & $97.0$ & $76.3$ & $86.8$ & $85.8$ & $47.9$ & $87.9$ & $72.8$ & $90.0$ & $57.1$ & $70.2$ & $67.5$ & $87.1$ & $86.3$ & $85.1$ & $89.7$ & $40.0$ & $77.7$ & $54.0$ & $94.1$ & $75.3$ & $76.4$\\
\hline
\cite{DBLP:conf/cvpr/OquabBL14} & $93.5$ & $78.4$ & $87.7$ & $80.9$ & $57.3$ & $85.0$ & $81.6$ & $89.4$ & $66.9$ & $73.8$ & $62.0$ & $89.5$ & $83.2$ & $87.6$ & $95.8$ & $61.4$ & $79.0$ & $54.3$ & $88.0$ & $78.3$ & $78.7$\\

\cite{DBLP:conf/eccv/ZeilerF14} & $96.0$ & $77.1$ & $88.4$ & $85.5$ &
$55.8$ & $85.8$ & $78.6$ & $91.2$ & $65.0$ & $74.4$ & $67.7$ & $87.8$ & $86.0$ & $85.1$ & $90.9$ & $52.2$ & $83.6$ & $61.1$ & $91.8$ & $76.1$ & $79.0$\\
\cite{DBLP:journals/corr/WeiXHNDZY14} & $97.7$ & $83.0$ & \boldsymbol{$93.2$} & $87.2$ & $59.6$ & $88.2$ & $81.9$ & $94.7$ & $66.9$ & $81.6$ & $68.0$ & $93.0$ & $88.2$ & $87.7$ & $92.7$ & \boldsymbol{$59.0$} & $85.1$ & $55.4$ & $93.0$ & $77.2$ & $81.7$\\
\cite{DBLP:conf/bmvc/ChatfieldSVZ14} & $96.8$ & $82.5$ & $91.5$ & $88.1$ & $62.1$ & $88.3$ & $81.9$ & $94.8$ & $70.3$ & $80.2$ &
$76.2$ & $92.9$ & $90.3$ & $89.3$ & $95.2$ & $57.4$ & $83.6$ & $66.4$ &
$93.5$ & $81.9$ & $83.2$\\
\cite{Simonyan15ICLR} & - & - & - & - & - & - & - & - & - & - & - & - & - & - & - & - & - & - & - & - & \boldsymbol{$89.0$} \\
\hline

BoE (\emph{VGG-VD},50) & \boldsymbol{$97.9$} & \boldsymbol{$86.3$} & $92.2$ & \boldsymbol{$88.6$} & \boldsymbol{$66.2$} & \boldsymbol{$90.4$} &
\boldsymbol{$83.6$} & \boldsymbol{$95.2$} & \boldsymbol{$76.6$} & \boldsymbol{$84.1$} & \boldsymbol{$77.3$} & \boldsymbol{$94.9$} & \boldsymbol{$94.6$} & \boldsymbol{$92.5$} & \boldsymbol{$95.5$} & $57.5$ & \boldsymbol{$87.3$} & \boldsymbol{$67.9$} & \boldsymbol{$94.6$} & \boldsymbol{$86.3$} & $85.5$\\
\hline
\end{tabular}
\vspace{3pt}
\caption{Comparison of classification results on the Pascal VOC 2012 dataset. For the sake of fair comparison, CNN models of all above methods are trained using the dataset used in the ILSVRC competition~\citep{DBLP:journals/corr/RussakovskyDSKSMHKKBBF14}, \ie, 1000 classes from the ImageNet~\citep{DBLP:conf/cvpr/DengDSLL009}.}
\label{tab:voc_2012}
}
\vspace{-0.1cm}
\end{table*}

\noindent\textcolor{blue}{
\textbf{The merging threshold.} The merging threshold $Th$ in Algorithm~\ref{alg:merging} controls how many mid-level elements should be merged together.
While keeping other parameters fixed, we evaluate this parameter under different settings.
As shown in Table~\ref{tab:th}, the best performance is reached when using value of $150$ for $Th$.}\\

\noindent\textcolor{blue}{\textbf{Pattern selection method in~\cite{Rematas15CVPR}.} To show the effectiveness of the proposed pattern selection (Sec.~\ref{subsubsec:pattern_Sselection}) and merging (Sec.~\ref{subsubsec:merging_detectors}) methods, we re-implemented the pattern selection method proposed by~\cite{Rematas15CVPR} and combine it with our framework.
In~\cite{Rematas15CVPR}, patterns are first ranked according to an interesting score and then non-overlapping patterns are selected in a
greedy fashion (please refer to Algorithm 1 in~\cite{Rematas15CVPR}).
In our case, after selecting patterns following ~\cite{Rematas15CVPR}, we
train detectors for the  mid-level elements retrieved from those patterns and construct a Bag-of-Elements representation (Sec.~\ref{subsubsec:Bag-of-Elements}).
On the VOC 2007 dataset, when using the \emph{VGG-VD} model and $50$ elements per category, this framework gives $85.0\%$ mAP, which is lower than that of our pattern selection method ($86.2\%$) and pattern merging method ($87.3\%$).}\\

\begin{figure*}[t]
\vspace{-0.0cm}
\begin{center}
\begin{tabular}{@{}c@{}c@{}c@{}c@{}c}
\includegraphics[width=0.2\linewidth]{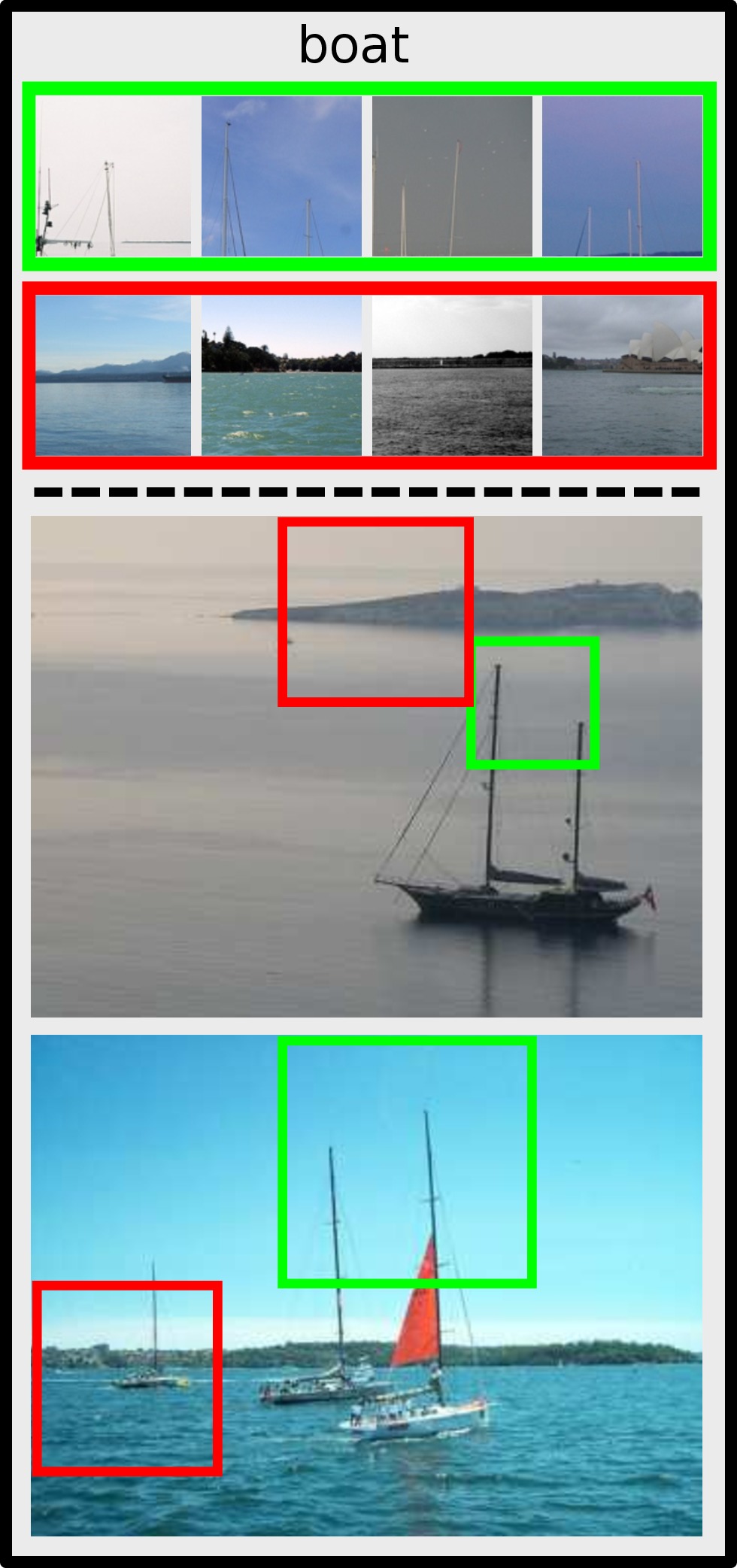} \ &
\includegraphics[width=0.2\linewidth]{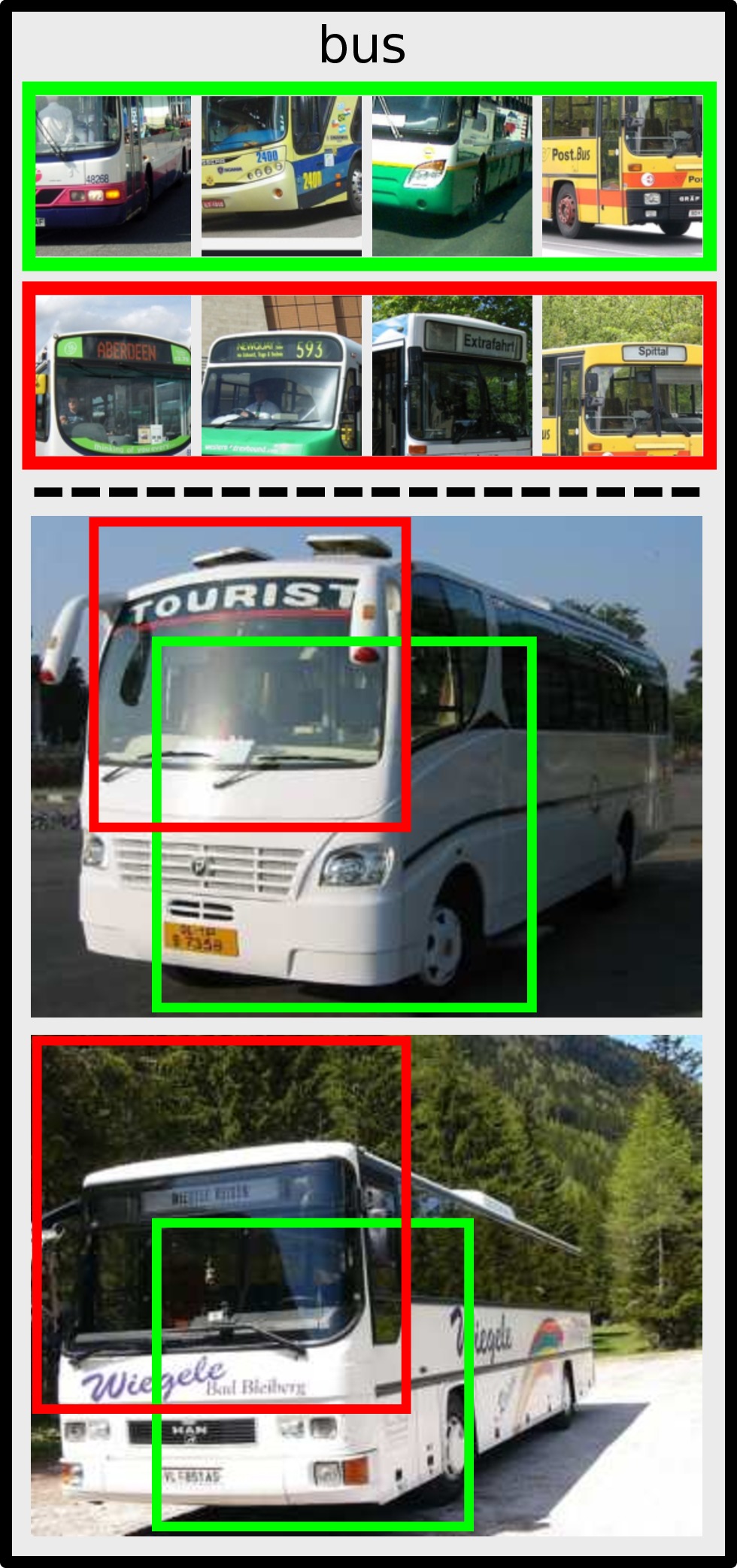} \ &
\includegraphics[width=0.2\linewidth]{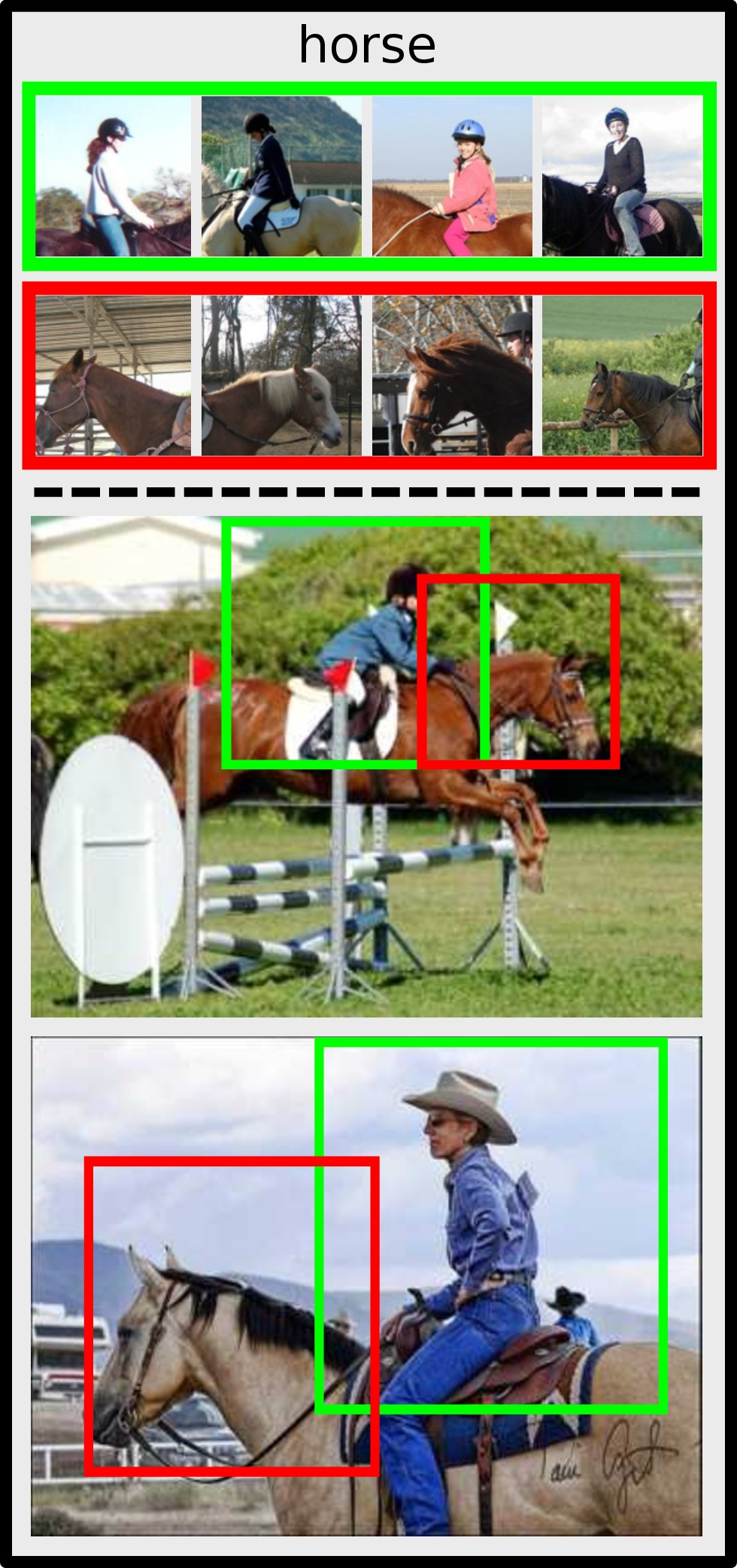} \ &
\includegraphics[width=0.2\linewidth]{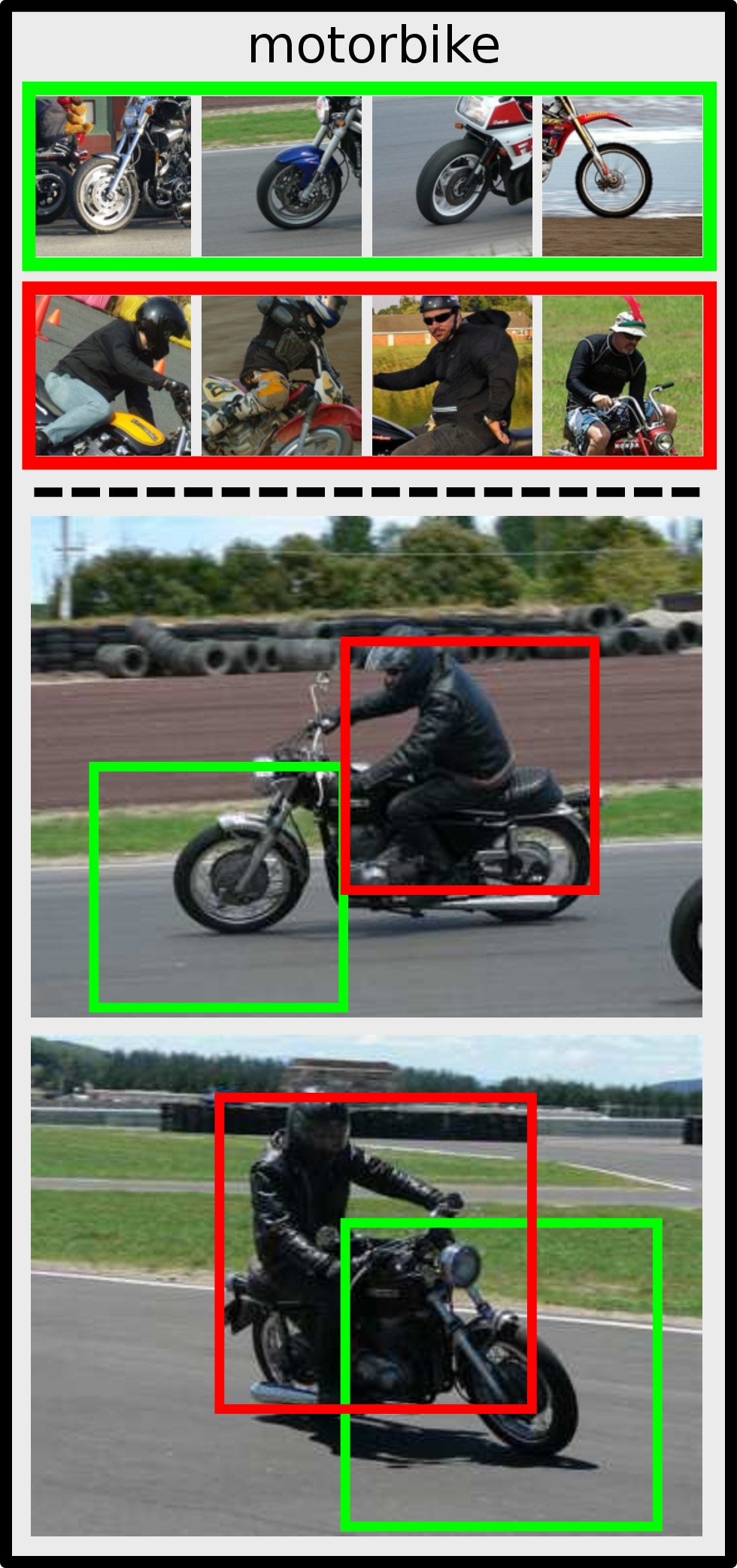} \ &
\includegraphics[width=0.2\linewidth]{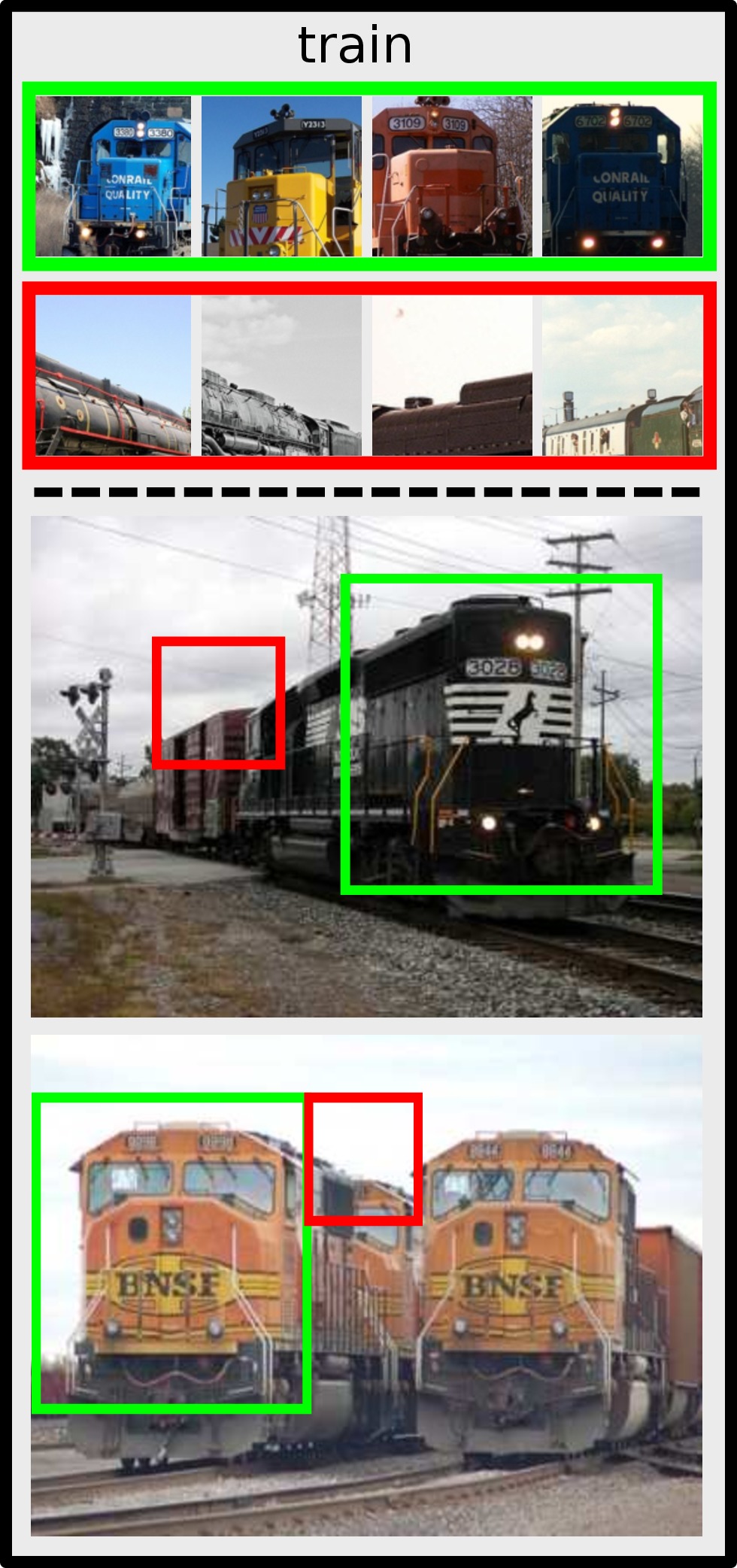} \ \\
\end{tabular}
\end{center}
\caption{Discovered mid-level visual elements and their corresponding detections on test images on the Pascal VOC 2007 dataset.}
\label{fig:det_VOC}
\end{figure*}

\subsubsection{Comparison with state-of-the-arts}
\label{subsubsec:state-of-the-art}

To compare with the state-of-the-art we use the BoE representation with $50$
mid-level elements per category, which demonstrated the best performance in the ablation
study (Fig.~\ref{fig:ablation_VOC}).
We also consider one baseline method (denoted as `FC') in which a
4096-dimensional fully-connected activation extracted from a global image is
used as the feature representation.
Table~\ref{tab:voc_2007} summarizes the performance of our approach as well as
state-of-the-art approaches on Pascal VOC 2007.

For encoding high-dimensional local descriptors,~\cite{DBLP:conf/nips/LiuSWHW14} propose a new variant of Fisher vector encoding~\citep{DBLP:conf/cvpr/PerronninLSP10}.
When the same \emph{CaffeRef} model is used in both methods, our performance is on par with that of~\cite{DBLP:conf/nips/LiuSWHW14}  ($76.4$\% \vs~$76.9$\%) whereas
{\em the feature dimension is $40$ times lower} ($5$k \vs~$200$k).
~\cite{DBLP:conf/cvpr/OquabBL14} adds two more layers on the top of fully-connected layers of the AlexNet and fine-tunes the pre-trained network on the PASCAL VOC. Although the method performs well ($77.7$\%), it relies on bounding box annotations which makes the task easier.
The FV-CNN method of~\cite{CimpoiMV15CVPR} extracts dense CNN activations
from the last convolutional layer and encodes them using the classic Fisher vector encoding.
Using the same \emph{VGG-VD} model, our BoE representation performs better than this method by a noticeable
margin ($87.3$\% \vs~$84.9\%$, despite the fact that we only use half of the image scales of FV-CNN ($5$ \vs~$10$) and feature dimension is significantly lower ($5$k \vs~$65$k).
As for the VOC 2012 dataset, as shown in Table~\ref{tab:voc_2012}, when using the \emph{VGG-VD} CNN model and $50$ elements
per category, the proposed BoE representation reaches a mAP of $85.5$\%, outperforming most state-of-the-art methods.

\subsubsection{Visualizing mid-level visual elements}
\label{subsubsec:visualization}
We visualize some mid-level elements discovered by the proposed MDPM algorithm and their firings on test images of the VOC 2007 dataset in Fig.~\ref{fig:det_VOC}.

Clearly, some mid-level visual elements capture discriminative parts of an
object (\eg, horse faces for the \emph{horse} class, the front of locomotives for the
\emph{train} class and wheels for the \emph{motorbike} class).
It is worth noting here these discriminative parts have been shown to be
extremely important for state-of-the-art object recognition systems, such as
Deformable Part Models~\citep{DBLP:journals/pami/FelzenszwalbGMR10} and
Poselets~\citep{DBLP:conf/iccv/BourdevM09}.
Moreover, rather than firing on the underlying object, some mid-level elements
focus on valuable contextual information.
For instance, as shown in Fig.~\ref{fig:det_VOC}, `people' is an important cue
both for the \emph{horse} and \emph{motorbike} classes, and `coastline' is
crucial for classifying \emph{boat}.
This fact indicates that mid-level elements may be a good tool for analysing
the importance of context for image classification (Sec.~\ref{sec:context}).

\subsubsection{Computational complexity}
\label{subsubsec:complexity}
The effectiveness of any mid-level visual element discovery process depends on being able to process very large numbers of image patches.
The recent work of \cite{Parizi15ICLR}, for example, takes $5$ days to find mid-level elements on the MIT Indoor dataset.
The proposed MDPM algorithm has been designed from the beginning with speed in mind, as it is based on a very efficient pattern mining algorithm.
Thus, for approximately $0.2$ million transactions created from CNN activations of image patches
on the Pascal VOC 2007 dataset,
association rule mining  takes only $23$ seconds to discover representative and discriminative patterns.
The bottleneck of our approach thus lies in the process of extracting
CNN activations from image patches, which is slower than the calculation of  hand-crafted HOG features.
All CNN-based approaches will suffer this time penalty, of course.
However, the process can be sped up using the technique proposed in~\cite{yoo2015multiscale} which avoids duplicated convolution operations between overlapping image patches.
GPUs can also be used to accelerate CNN feature extraction.

\subsection{Scene classification}
\label{subsec:scene}
We now provide detailed analysis of the proposed system for the task of scene classification on the MIT Indoor dataset.
As many mid-level element discovery algorithms have reported performance
on this dataset, we first provide a comprehensive comparison between
these algorithms and our method in Sec.~\ref{subsubsec:comparison_mid}.
The comparison between the performance of state-of-the-art methods with CNN involved and ours are presented in Sec.~\ref{subsubsec:MIT_state-of-the-arts}.
Finally, we visualize some mid-level elements discovered by the proposed
MDPM algorithm and their firings in Sec.~\ref{subsubsec:MIT_visualization}.
For this dataset, the value of $conf_{min}$ (Eq.~\ref{eq:confidence}) is always set as $30\%$.

\subsubsection{Comparison with methods using mid-level elements}
\label{subsubsec:comparison_mid}
\begin{table}
\begin{center}
\begin{tabular}{  l|c|c }
\hline
Method & $\#$ of elements & Acc (\%)\\
\hline\hline
\cite{DBLP:conf/eccv/SinghGE12} & $210$ & $38.10$\\
\cite{DBLP:conf/cvpr/JunejaVJZ13} & $50$ & $46.10$\\
\cite{DBLP:conf/cvpr/LiWT13} & $20$ & $46.40$\\
\cite{DBLP:conf/icml/WangWBLT13} & $11$ & $50.15$\\
\cite{DBLP:conf/iccv/SunP13} & $73$ & $51.40$\\
\cite{DBLP:conf/eccv/BossardGG14} & $50$ & $54.40$\\
\cite{DBLP:conf/nips/DoerschGE13} & $200$ & $64.03$\\
\cite{Parizi15ICLR} & $5$ & $73.30$\\
\hline
LDA-Retrained (\emph{CaffeRef}) & $20$ & $58.78$\\
LDA-Retrained (\emph{CaffeRef})& $50$ & $62.30$\\
LDA-KNN (\emph{CaffeRef}) &$20$ & $59.14$\\
LDA-KNN (\emph{CaffeRef}) & $20$ & $63.93$\\
\hline
BoE (\emph{CaffeRef}) & $20$ & $68.24$\\
BoE (\emph{CaffeRef}) & $50$ & $69.69$\\
BoE (\emph{VGG-VD}) & $20$ & $76.93$\\
BoE (\emph{VGG-VD}) & $50$ & \boldsymbol{$77.63$}\\
\hline
\end{tabular}
\end{center}
\caption{Classification results of mid-level visual element discovery algorithms on MIT Indoor dataset.}\vspace{-3mm}
\label{tab:discrminativePatch}
\end{table}

\begin{figure*}[t]
\vspace{-0.0cm}
\begin{center}
\begin{tabular}{@{}c@{}c@{}c@{}c@{}c}
\includegraphics[width=0.2\linewidth]{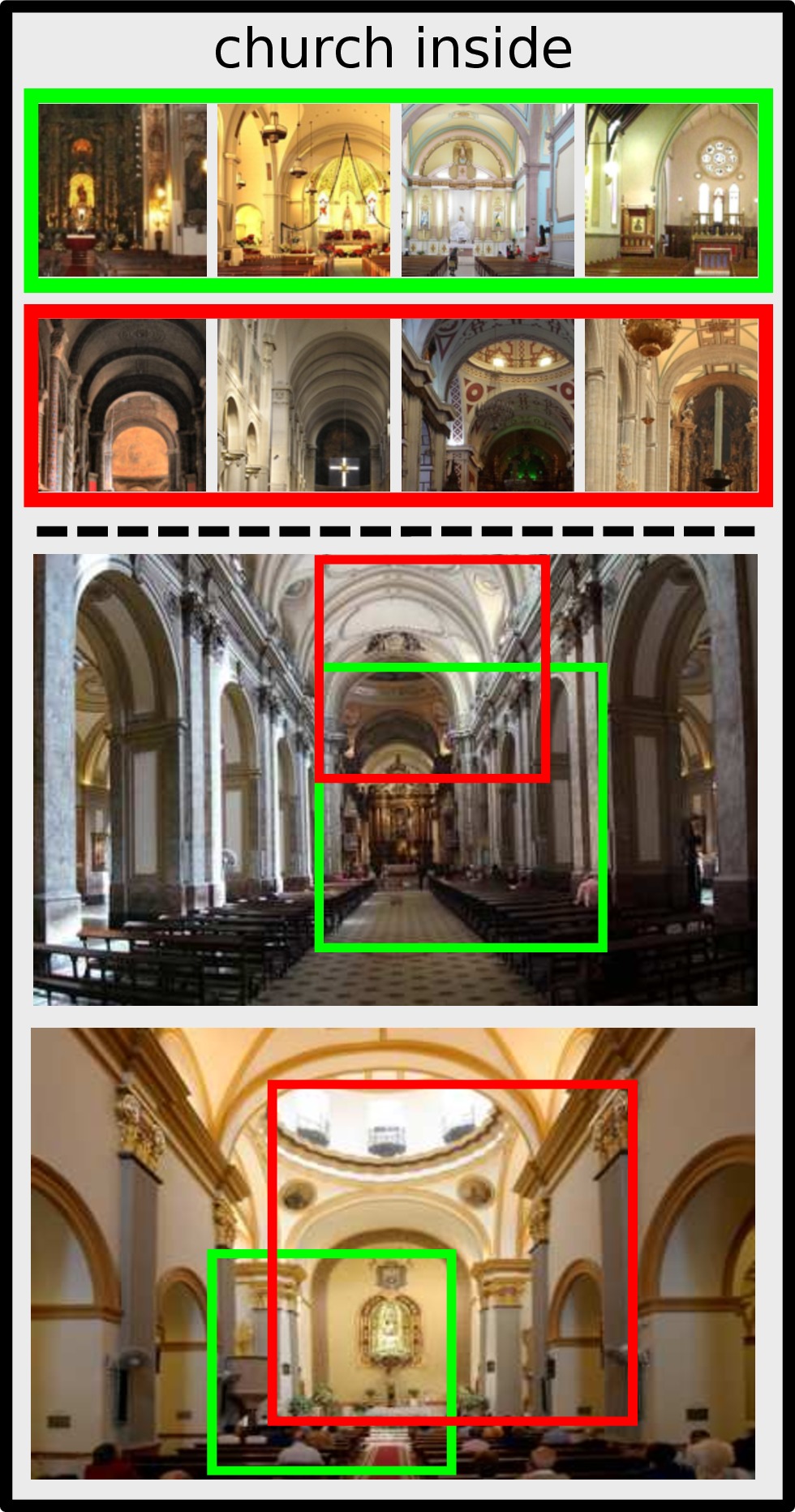} \ &
\includegraphics[width=0.2\linewidth]{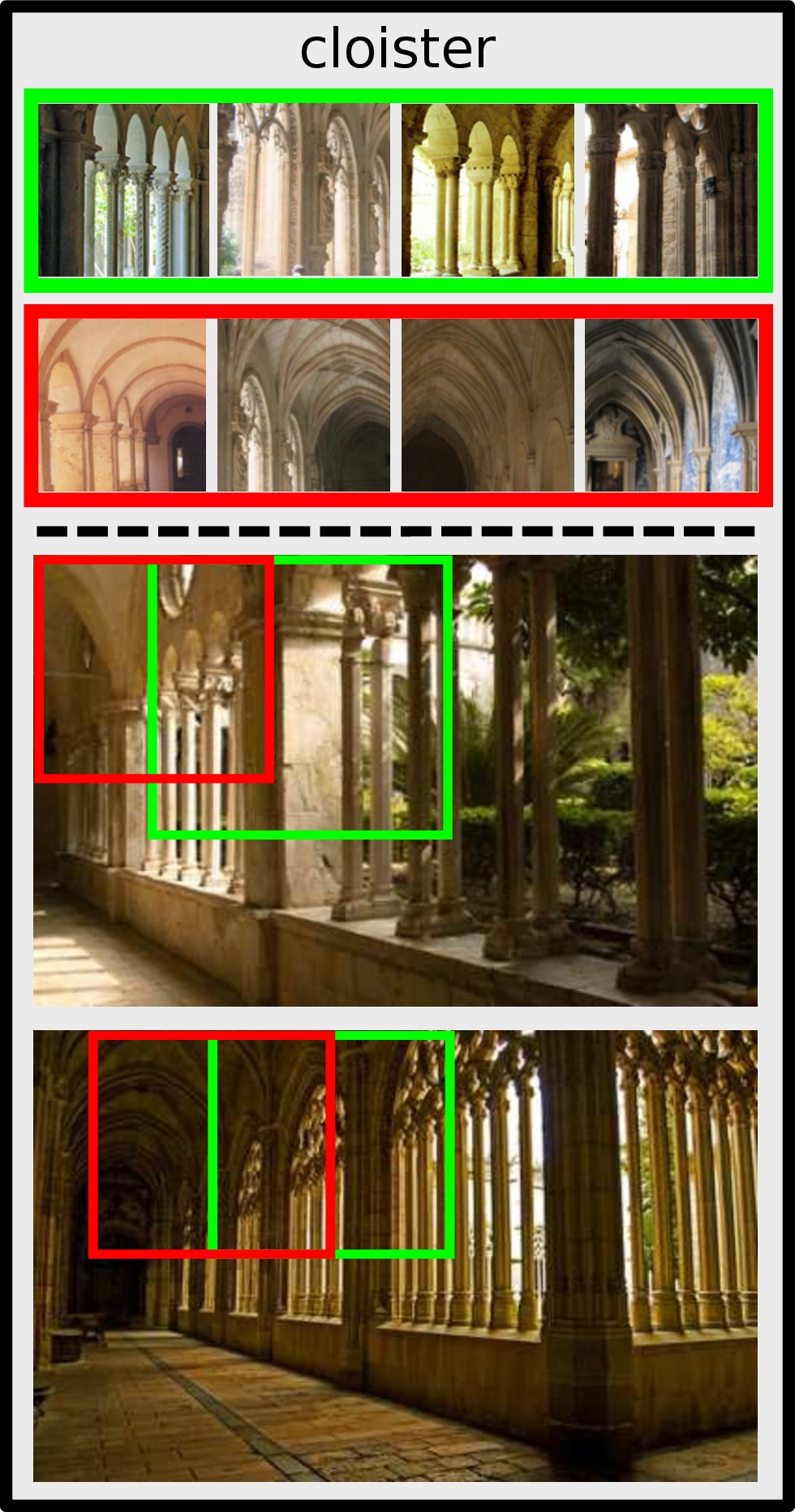} \ &
\includegraphics[width=0.2\linewidth]{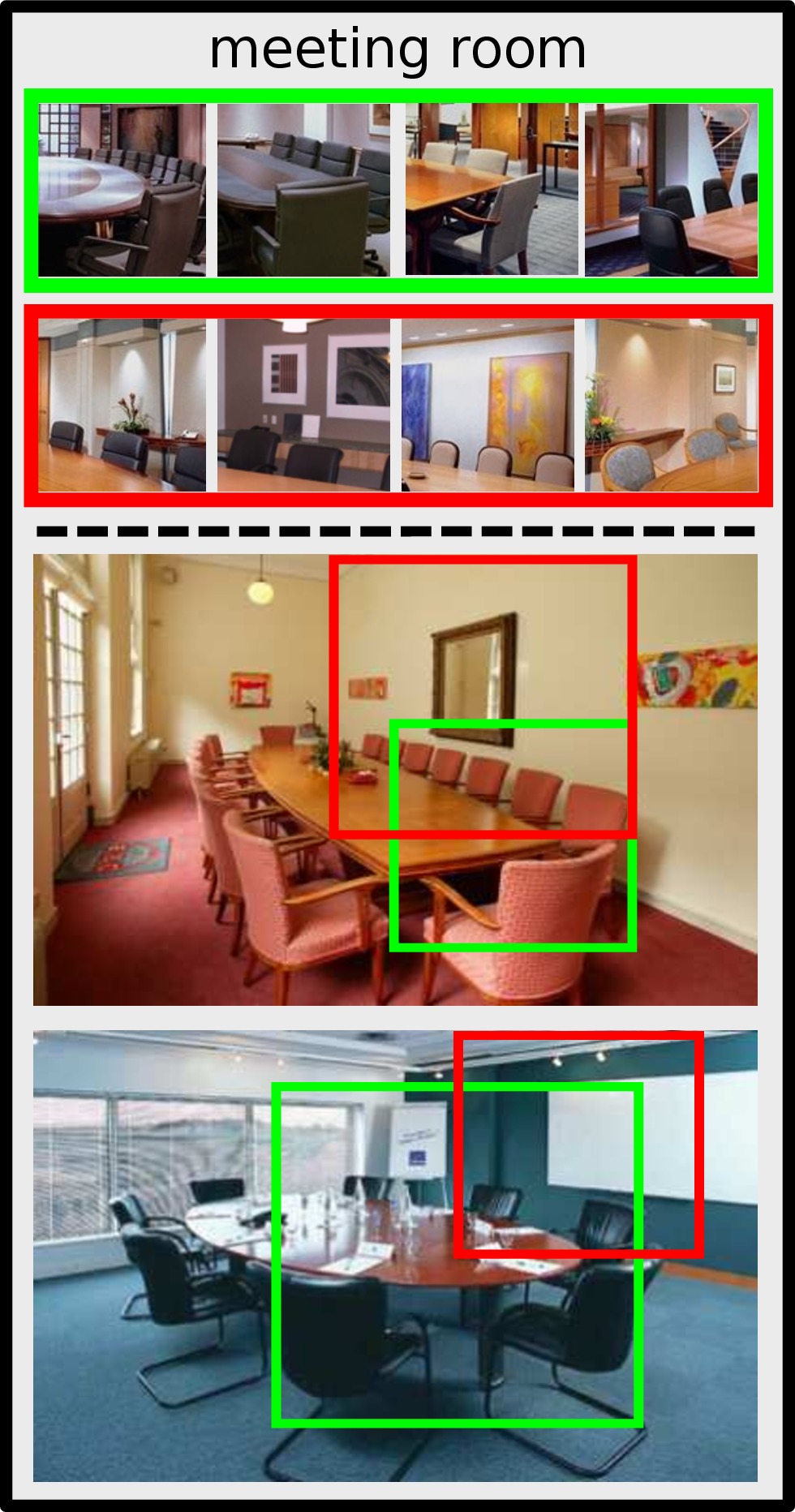} \ &
\includegraphics[width=0.2\linewidth]{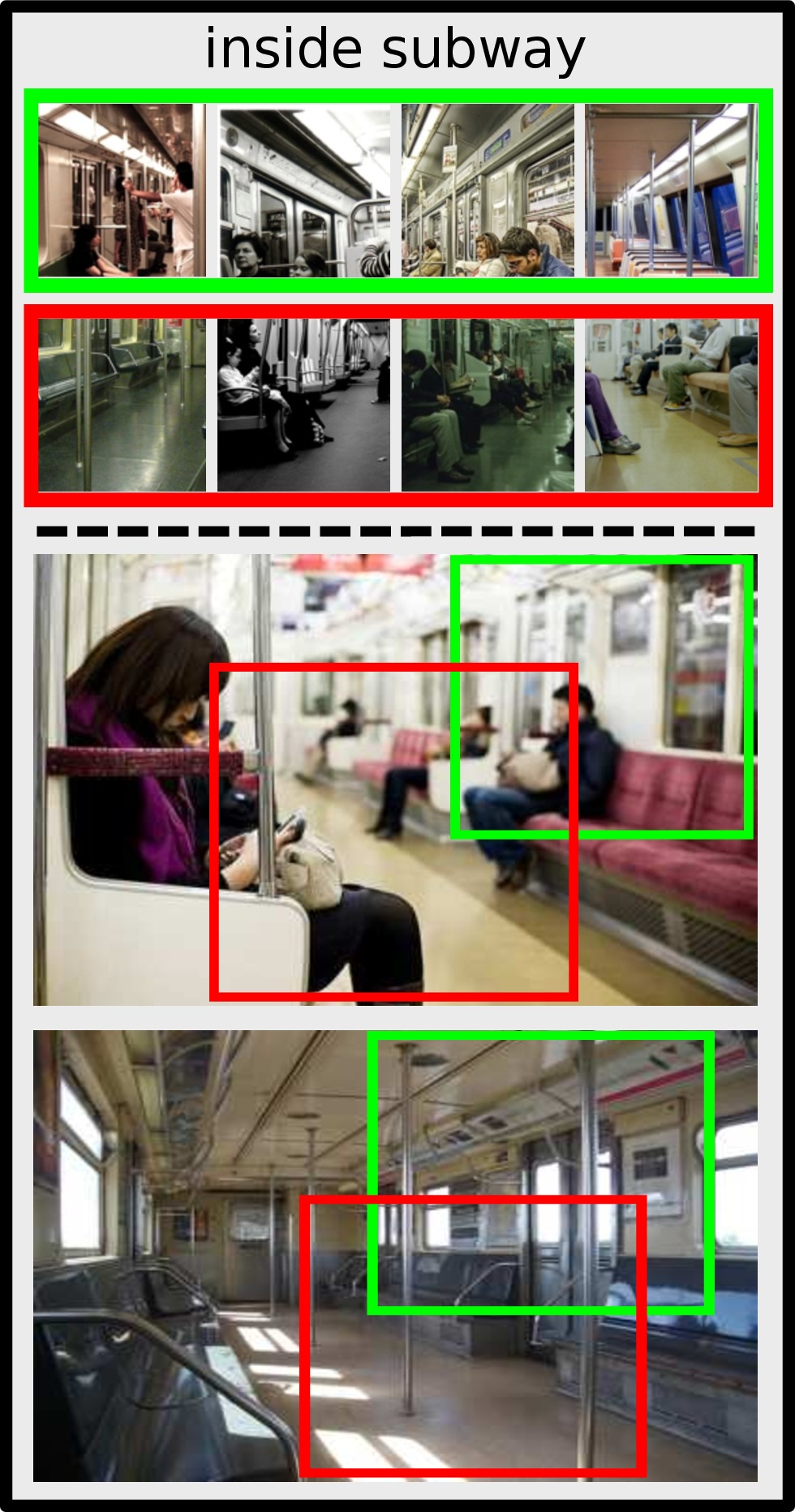} \ &
\includegraphics[width=0.2\linewidth]{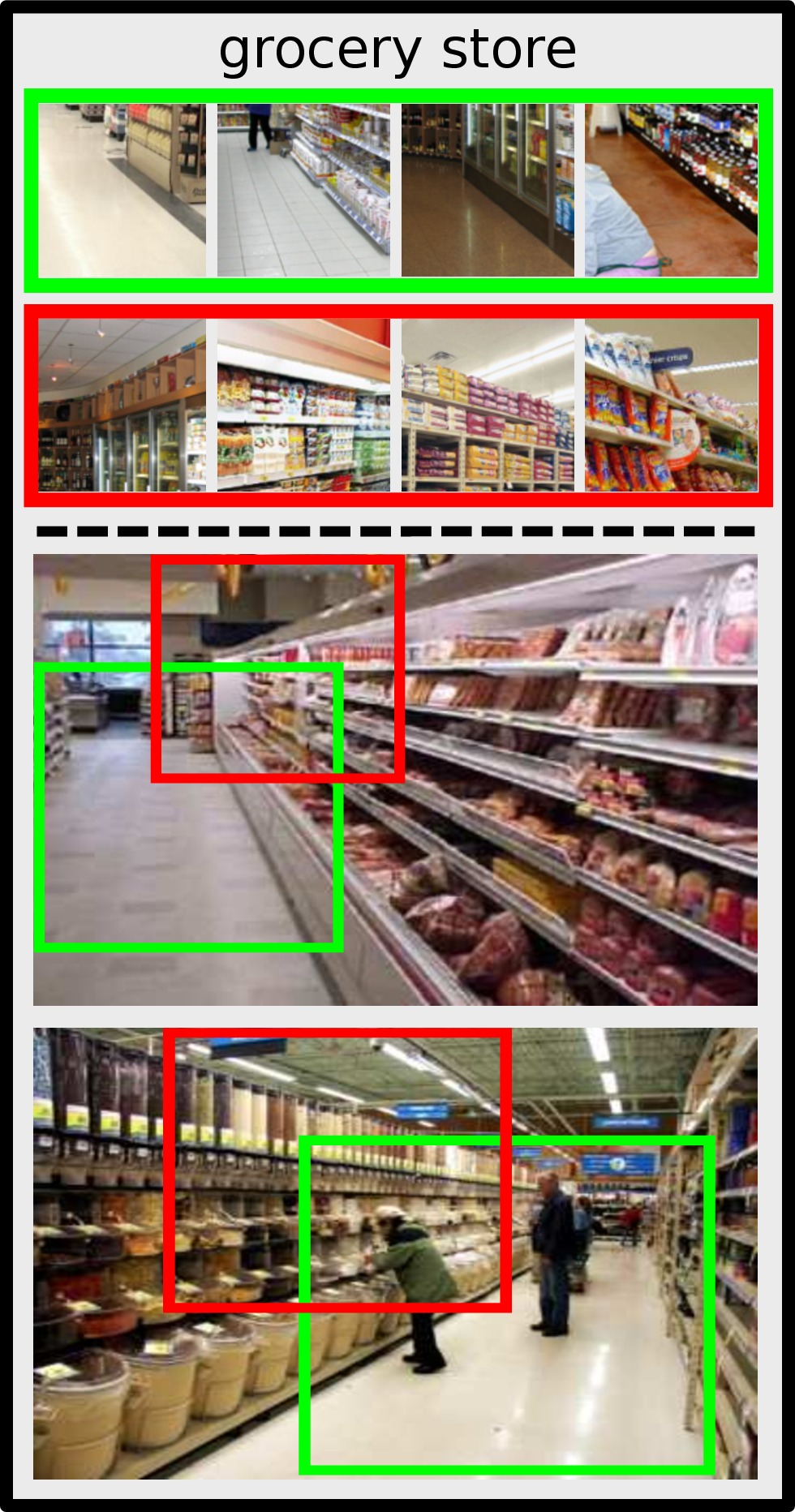} \ \\
\includegraphics[width=0.2\linewidth]{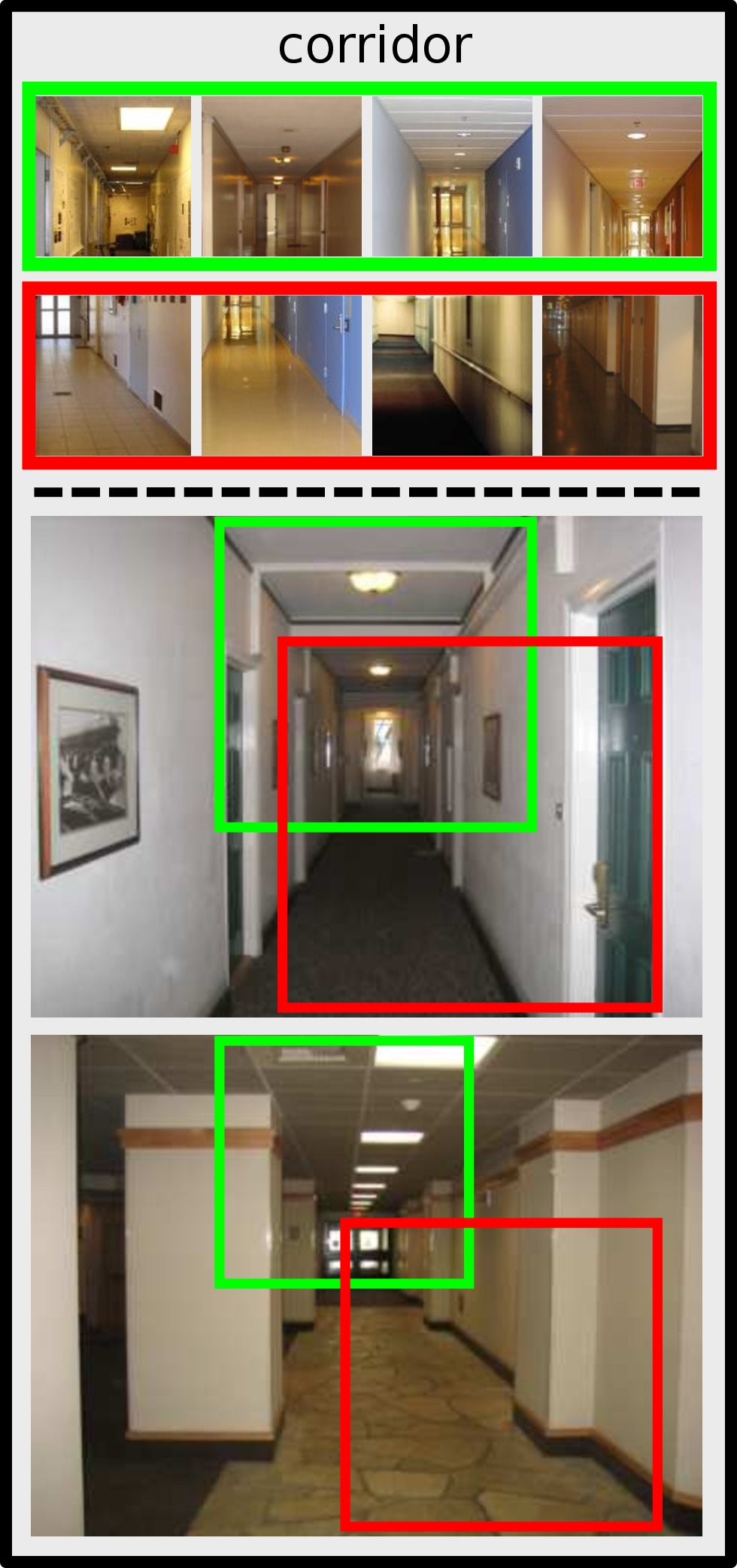} \ &
\includegraphics[width=0.2\linewidth]{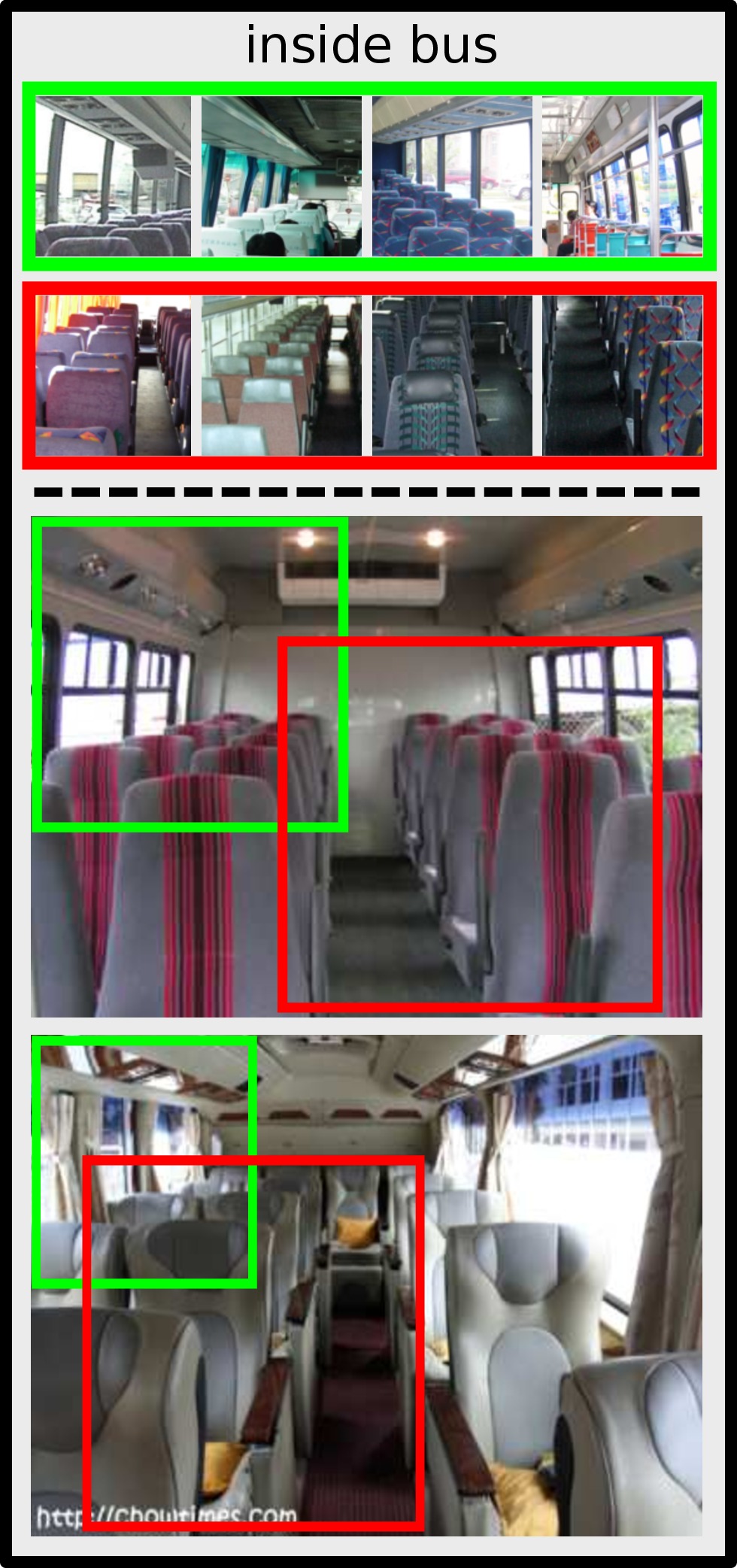} \ &
\includegraphics[width=0.2\linewidth]{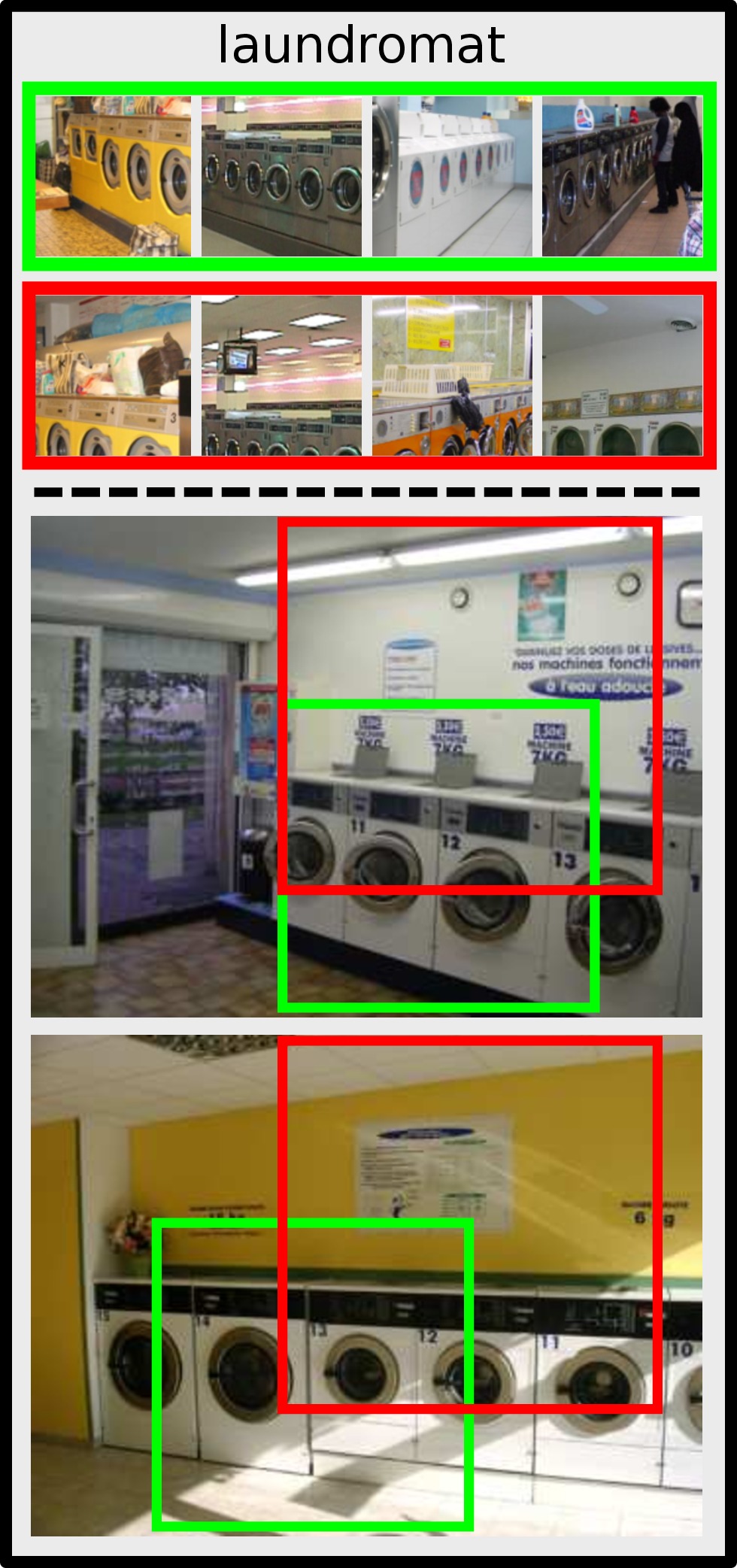} \ &
\includegraphics[width=0.2\linewidth]{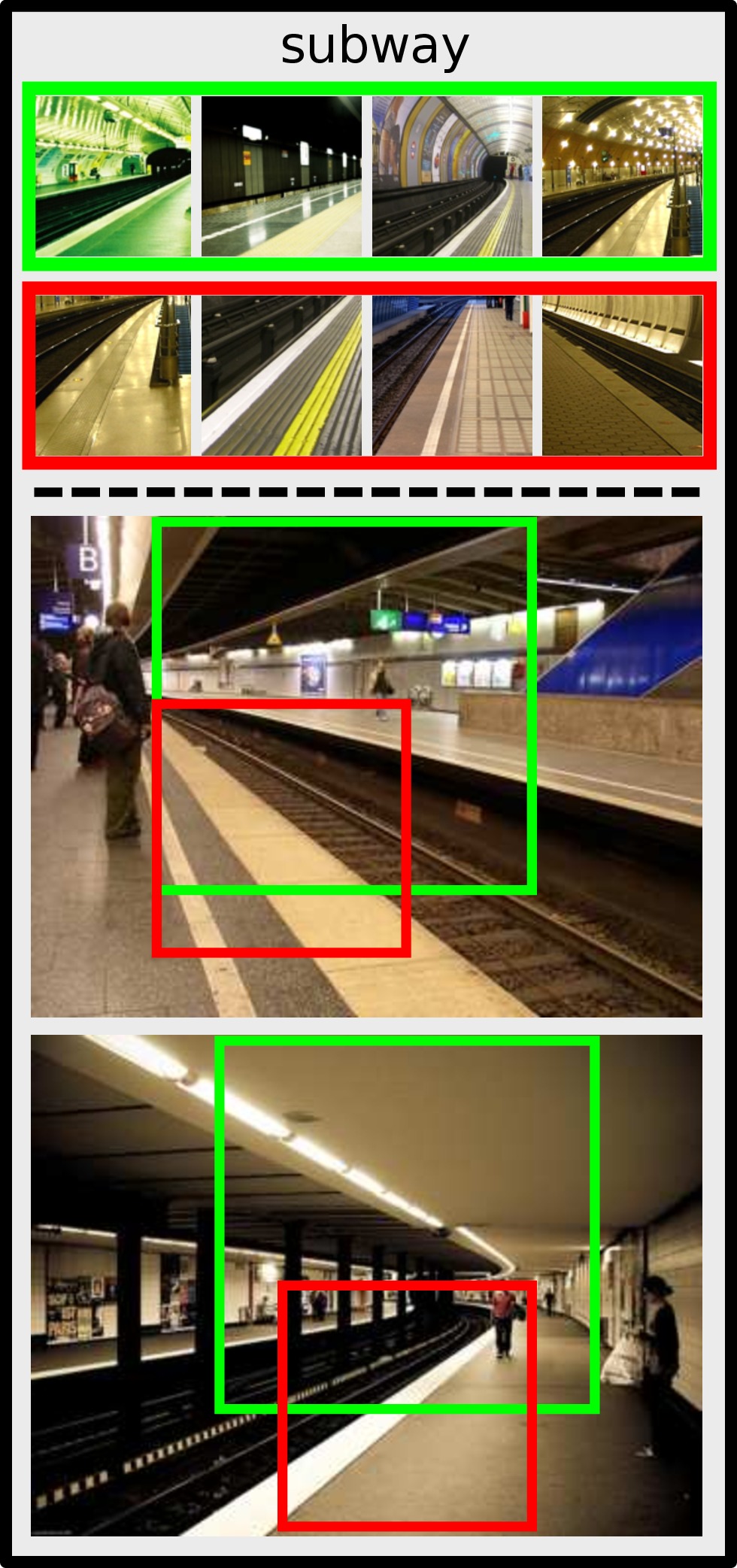} \ &
\includegraphics[width=0.2\linewidth]{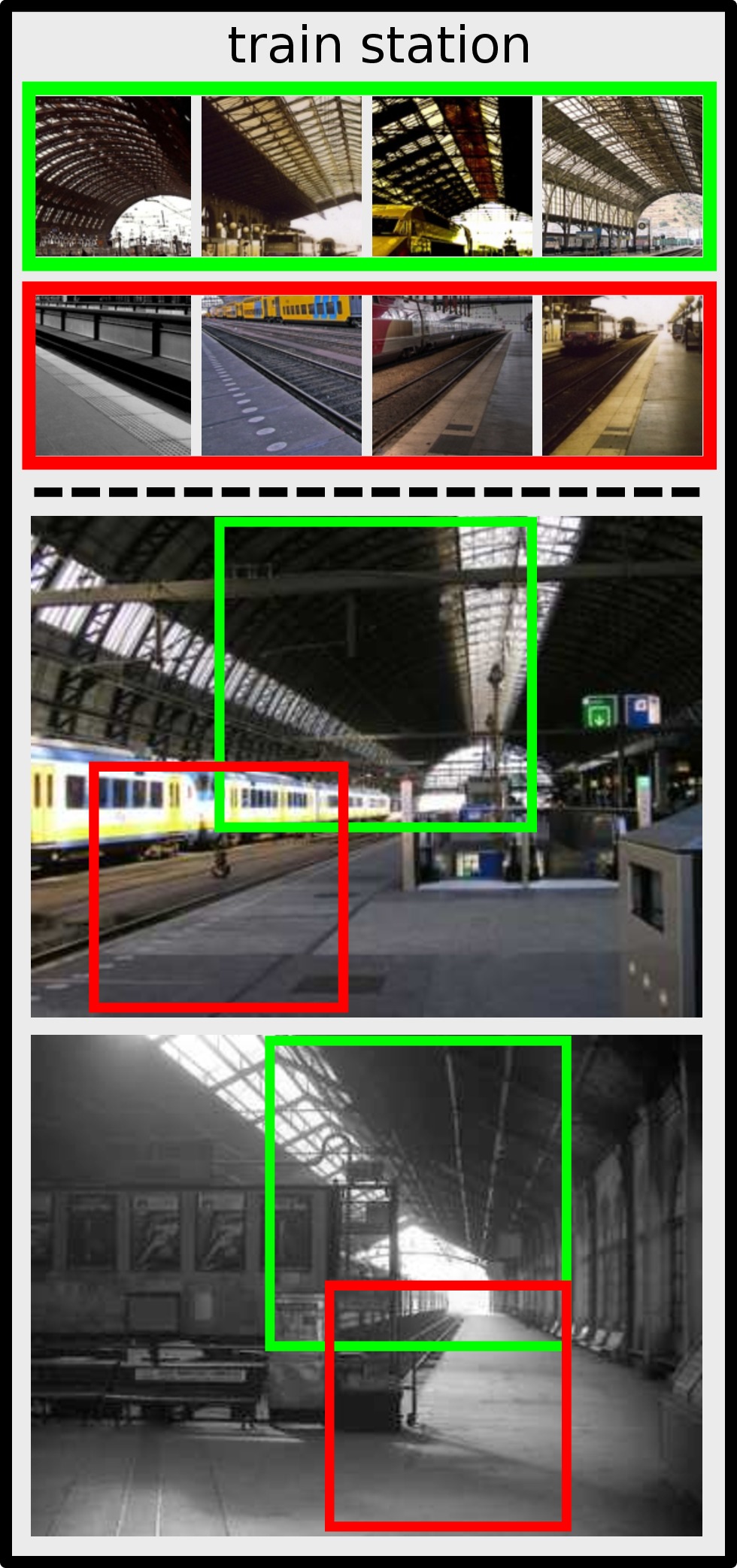} \ \\
\end{tabular}
\end{center}

\caption{Discovered mid-level visual elements and their corresponding detections on test images on the MIT Indoor dataset.}
\label{fig:det_MIT}
\end{figure*}

As hand-crafted features, especially HOG, are widely utilized as image patch representations in previous works,
we here analyze
the performance of previous approaches if CNN activations are used
in place of their original feature types.
We have thus designed two baseline methods so as to use
CNN activations as an image patch representation.

The first baseline ``LDA-Retrained'' initially trains Exemplar LDA using the CNN activation of a sampled patch
and then re-trains the detector $10$ times by adding top-$10$ positive detections as positive training samples at each iteration. This is
similar to the ``Expansion'' step of~\cite{DBLP:conf/cvpr/JunejaVJZ13}.
The second  baseline ``LDA-KNN'' retrieves 5-nearest neighbors of an image patch and trains an
LDA detector using the CNN activations of retrieved
patches (including itself) as positive training data.
For both baselines, discriminative detectors are selected based on the Entropy-Rank Curves proposed by~\cite{DBLP:conf/cvpr/JunejaVJZ13}.

As shown in Table~\ref{tab:discrminativePatch}, when using the
\emph{CaffeRef} model, MDPM achieves significantly better results than both baselines in the same setting.
This attests to the fact that the pattern mining approach at the core of MDPM is an important factor in its performance.

We also compare the proposed method against recent work in mid-level visual element discovery in Table~\ref{tab:discrminativePatch}.
Clearly, by combining the power of deep features and pattern mining techniques, the proposed method outperforms all previous mid-level element discovery methods by a sizeable margin.

\subsubsection{Comparison with methods using CNN}
\label{subsubsec:MIT_state-of-the-arts}
\begin{table}[t]
\begin{center}
\begin{tabular}{ l|c|l }
\hline
Method & Acc (\%)  & Comments\\
\hline\hline
FC (\emph{CaffeRef}) & $57.74$ & CNN for whole image\\
FC (\emph{VGG-VD}) & $68.87$ & CNN for whole image\\
\hline
\cite{6910029}  & $58.40$ & \texttt{OverFeat} toolbox\\
\cite{DBLP:conf/eccv/GongWGL14} (\emph{CaffeRef}) & $68.88$ & concatenation\\
\cite{DBLP:journals/corr/AzizpourRSMC14} & $65.90$ & jittered CNN \\
\cite{DBLP:journals/corr/AzizpourRSMC14} & $66.30$ & FT CNN \\
\cite{DBLP:conf/nips/ZhouLXTO14} & $68.24$ & Places dataset used\\
\cite{DBLP:conf/nips/LiuSWHW14} (\emph{CaffeRef}) & $68.20$ & new Fisher encoding\\
\cite{DBLP:conf/cvpr/LiuSH15} (\emph{CaffeRef}) & $68.80$ & cross-layer pooling\\
\cite{Parizi15ICLR} & $73.30$ & \textcolor{blue}{unified pipeline}\\
\cite{DBLP:conf/iccv/LinRM15} (\emph{VGG-VD}) & $77.55$ & \textcolor{blue}{Bilinear CNN}\\
\cite{DBLP:conf/cvpr/GaoBZD16} (\emph{VGG-VD}) & $76.17$ & \textcolor{blue}{compact Bilinear CNN}\\
\cite{DBLP:journals/cviu/MettesGS16} & $77.40$ & shared parts\\
\cite{DBLP:journals/ijcv/CimpoiMKV16} (\emph{VGG-VD}) & \boldsymbol{$81.00$} & Fisher Vector\\
\hline
BoE (\emph{CaffeRef}) & $69.69$ & $50$ elements\\
BoE (\emph{VGG-VD}) & $77.63$ & $50$ elements\\
\hline
\end{tabular}
\end{center}
\caption{Classification results of methods using CNN activations on MIT Indoor dataset.}%
\vspace{-3mm}
\label{tab:VLAD_comparison}
\end{table}

In Table~\ref{tab:VLAD_comparison}, we compare
the proposed
method to others in which CNN activations are used, at the task of scene classification.
The baseline method,  using fully-connected CNN activations extracted from the whole image using \emph{CaffeRef} (\emph{resp.} \emph{VGG-VD}), gives an accuracy of $57.74\%$ (\emph{resp.} $68.87$\%).
The proposed method achieves $69.69\%$ using \emph{CaffeRef} and $77.63$\% using
\emph{VGG-VD}, which are significant improvements over the corresponding baselines.

Our method is closely related to~\cite{DBLP:conf/eccv/GongWGL14} and ~\cite{DBLP:conf/cvpr/LiuSH15} in the sense that all rely on off-the-shelf CNN activations of image patches.
Our BoE representation, which is based on  mid-level elements discovered by
the MDPM algorithm, not only outperforms~\cite{DBLP:conf/eccv/GongWGL14}
on $128\times128$ and $64\times64$ patches by a considerable
margin ($69.69\%$ \emph{vs.} $65.52\%$ and $69.69\%$ \emph{vs.} $62.24\%$),
it also slightly outperforms that of \cite{DBLP:conf/cvpr/LiuSH15} ($69.69\%$ \emph{vs.} $68.20\%$).
\textcolor{blue}{Our performance ($77.63\%$) is also comparable to that of the recent works of bilinear CNN~\citep{DBLP:conf/iccv/LinRM15} ($77.55\%$) and its compact version~\citep{DBLP:conf/cvpr/GaoBZD16} ($76.17\%$) when the \emph{VGG-VD} model is adopted.}

\textcolor{blue}{Fine-tuning has been shown to be beneficial when transferring pre-trained CNN models to another dataset~\citep{DBLP:conf/cvpr/GirshickDDM14,DBLP:conf/cvpr/OquabBL14,
DBLP:conf/eccv/Agrawal14}.
We are interested in how the performance changes if a fine-tuned CNN model
is adopted in our framework.
For this purpose, we first fine-tuned the \emph{VGG-VD} model on the MIT Indoor dataset with a learning rate of $0.0005$.
The fine-tuned model reaches $69.85\%$ accuracy after $70k$ iterations.
After applying the fine-tuned model in our framework, the proposed approach reaches $71.82\%$ accuracy,
which is lower than the case of using a pre-trained model ($77.63\%$)  but still
improves the accuracy
of directly fine-tuning ($69.85\%$).
The underlying reason is probably due to
the small training data size of the MIT Indoor dataset and the large capacity of the
{\em  VGG-VD} model.
We plan to investigate this issue in our future work.
Similar observation was made in \cite{DBLP:conf/cvpr/GaoBZD16}.
}

\subsubsection{Visualizing mid-level visual elements}
\label{subsubsec:MIT_visualization} We visualize some visual elements
discovered and their firings on test images of the MIT Indoor dataset in
Fig.~\ref{fig:det_MIT}.
It is intuitive that the discovered mid-level visual elements capture the
visual patterns which are often repeated within a scene category.
Some of the mid-level visual elements refer to frequently occurring object
configurations, \eg, the configuration between table and chair in the
\emph{meeting room} category.
Some instead capture a particular type of object in the scene, such as washing
machines in the \emph{laundromat}.

\subsection{Do mid-level visual elements capture context?}
\label{sec:context}

\begin{table*}[t]
\scriptsize{
\setlength{\tabcolsep}{3pt}
\def\arraystretch{1.2}
\center
\begin{tabular}{l@{\hspace{0.6em}}c@{\hspace{0.6em}}c@{\hspace{0.6em}}c@{\hspace{0.6em}}c
@{\hspace{0.6em}}c@{\hspace{0.6em}}c@{\hspace{0.6em}}c@{\hspace{0.6em}}c
@{\hspace{0.6em}}c@{\hspace{0.6em}}c@{\hspace{0.6em}}c@{\hspace{0.6em}}c
@{\hspace{0.6em}}c@{\hspace{0.6em}}c@{\hspace{0.6em}}c@{\hspace{0.6em}}c
@{\hspace{0.6em}}c@{\hspace{0.6em}}c@{\hspace{0.6em}}c@{\hspace{0.6em}}c | c}
\hline
\textbf{VOC 2007} & aero  &  bike &  bird & boat &  bottle  &  bus  &  car  &  cat  &  chair & cow & table &  dog  & horse & mbike & person  & plant & sheep & sofa & train & tv & Average\\
\hline
gt-object & $95.0$ & $100.0$ & $90.0$ & $65.0$ & $43.7$ & $100.0$ & $100.0$ & $100.0$ & $55.0$ & $87.5$ & $100.0$ & $100.0$ & $100.0$ & $100.0$ & $40.0$ & $88.2$ & $90.0$ & $100.0$ & $100.0$ & $95.0$ & $87.5$\\
\hline
object-context & $0.0$ & $0.0$ & $0.0$ & $0.0$ & $56.3$ & $0.0$ & $0.0$ & $0.0$ & $45.0$ & $0.0$ & $0.0$ & $0.0$ & $0.0$ & $0.0$ & $60.0$ & $5.9$ & $0.0$ & $0.0$ & $0.0$ & $0.0$ & $8.3$\\
\hline
scene-context & $5.0$ & $0.0$ & $10.0$ & $35.0$ & $0.0$ & $0.0$ & $0.0$ & $0.0$ & $0.0$ & $12.5$ & $0.0$ & $0.0$ & $0.0$ & $0.0$ & $0.0$ & $5.9$ & $10.0$ & $0.0$ & $0.0$ & $5.0$ & $4.2$\\
\hline
\end{tabular}
\vspace{3pt}
\caption{Firing types of the top-$20$ mid-level elements on the Pascal VOC 2007 dataset (\emph{VGG-VD} model adopted).}
\label{tab:voc_2007_context}
}
\vspace{-0.1cm}
\end{table*}

It is well known that humans do not perceive every instance in the scene in isolation. Instead, context information plays an important role~\citep{DBLP:journals/ijcv/Torralba03,DBLP:journals/ijcv/HoiemEH08,
DBLP:conf/cvpr/DivvalaHHEH09,DBLP:conf/nips/MalisiewiczE09,
DBLP:journals/pami/ChoiTW12,DBLP:conf/cvpr/LiuW12}.
In the our scenario, we consider how likely that the discovered mid-level visual elements fire on context rather than the underlying object.
In this section, we give answer to this question based on the Pascal VOC07 dataset which has ground truth bounding boxes annotations.

\begin{figure}[t]
\vspace{-0.0cm}
\centering
\begin{tabular}{@{}c}
\includegraphics[width=.85\linewidth]{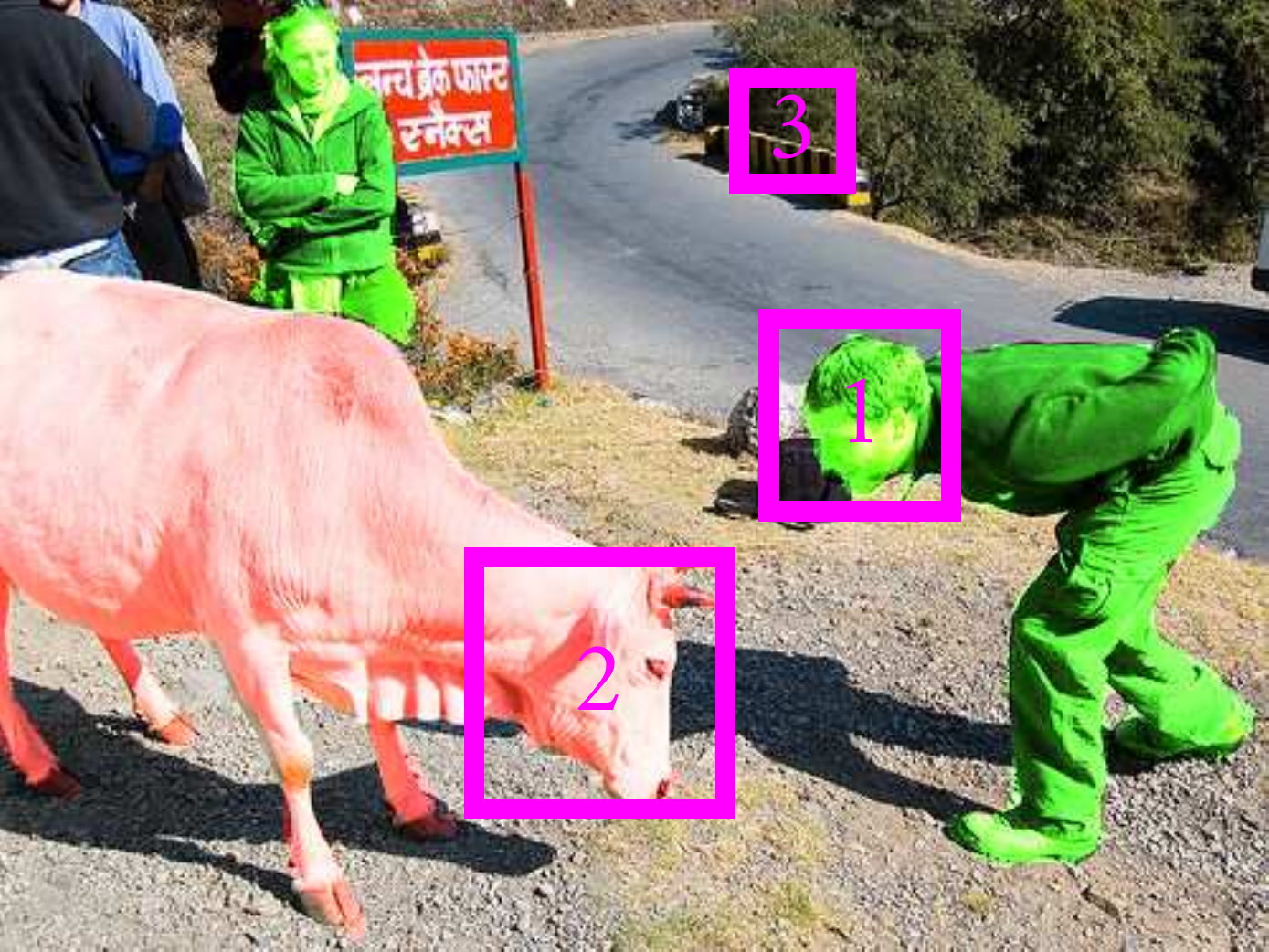} \\
\end{tabular}
\caption{\textcolor{blue}{An illustration of the three firing types of mid-level elements. In the image, ground-truth object instances of the underlying category (\eg, ``person'') are overlaid in green while instances of other categories (\eg, ``cow'') are overlaid in red. Obviously, the firing (1) fires on the ground-truth object while firings (2) and (3) belong to object and scene context respectively.} }
\label{fig:demoContext}
\end{figure}

\begin{figure*}[t]
\vspace{-0.0cm}
\begin{center}
\begin{tabular}{@{}c}
\includegraphics[width=1\linewidth]{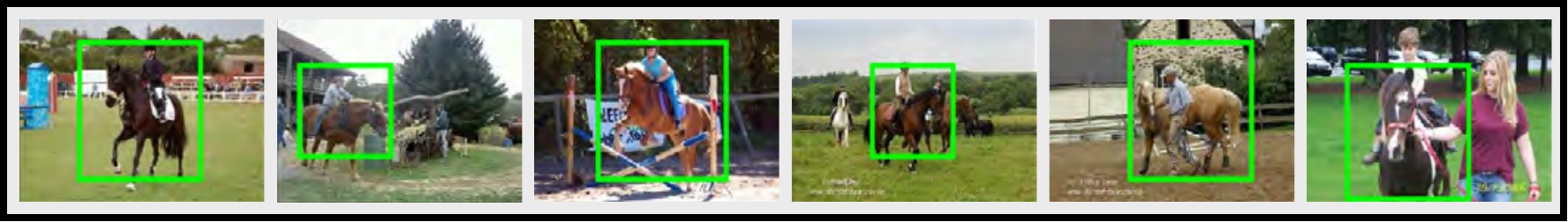} \ \\
\includegraphics[width=1\linewidth]{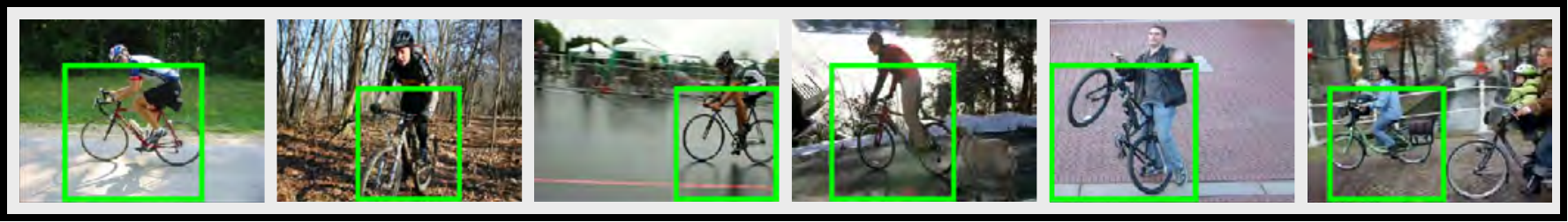}  \\
\end{tabular}
\end{center}
\caption{Detections of two object-context mid-level visual elements discovered from the ``person'' category on the Pascal VOC 2007 dataset.}
\label{fig:contextual_VOC}
\end{figure*}

\subsubsection{Object and scene context}
\label{subsec:context_definition}
We first need to define context qualitatively.
For this purpose, we leverage the test set of the segmentation challenge of
the Pascal VOC 2007 dataset in which per-pixel labeling is available.
Given a test image of a given object category, its ground-truth pixels annotations $S$ are categorized into the following
three categories,
\begin{itemize}
\item $S_{gt}$: pixels belong to the underlying object category.
\item $S_{ot}$: pixels belong to any of the rest $19$ object categories.
\item $S_{sc}$: pixels belong to none of the $20$ object categories, \ie,
belong to the background.
\end{itemize}
Accordingly, given a firing (\ie, predicted bounding box) $B$ of a mid-level visual element on an image, we compute an overlap ratio for each of the three types of pixels,
\begin{equation}
\label{eq:ratio}
O_{gt} = \frac{|B \cap S_{gt}|}{|B|}, O_{ot} = \frac{|B \cap S_{ot}|}{|B|},O_{sc} = \frac{|B \cap S_{sc}|}{|B|},
\end{equation}
where $|\cdot|$ measures cardinality.
Note that $O_{gt}+O_{ot}+O_{sc}=1$.
By comparing the three types of overlap ratios, we can easily define three firing types, which includes two types of context firing and one ground-truth object firing, %
\begin{itemize}
\item \textbf{Scene context}: if $O_{sc}>0.9$.
\item \textbf{Object context}: if $O_{sc}<=0.9$ and $O_{ot}>O_{gt}$.
\item \textbf{Ground-truth object}: if $O_{sc}<=0.9$ and $O_{ot}<O_{ht}$.
\end{itemize}

\textcolor{blue}{Fig.~\ref{fig:demoContext} depicts a visual example of the three firing types.} In practice, for each image in the test set, we collect the predicted bounding box with the maximum detection score if there exists any positive detections (larger than a threshold), followed by categorizing it into one of the three types based on Eq.~\ref{eq:ratio}.
Thus, a mid-level visual element is categorized into the three firing types based on its major votes of positive detections.

\subsubsection{Analysis}
Following the context definition in Sec.~\ref{subsec:context_definition}, we categorize the each of the top-$20$ discovered mid-level visual elements of each category of the Pascal VOC 2007 dataset into one of the three categories: gt-object, object or scene context.
The distribution of this categorization is illustrated in Table~\ref{tab:voc_2007_context}.

Interestingly, for many classes, the majority of the discovered mid-level visual elements fires on the underlying object, and context information seems to be less important.
More specifically, as shown in Table~\ref{tab:voc_2007_context}, mid-level visual elements in $10$ out of $20$ classes never capture context information, which reflects image patches capture context in these classes are neither representative nor discriminative.
On average, more than $87\%$ mid-level visual element capture the underlying object across all the categories.

We also observe that contextual information from other object categories plays a important role for discovering mid-level visual element
from \emph{person}($60.0\%$), \emph{bottle}($56.3\%$) and \emph{chair}($45.0\%$).
Fig.~\ref{fig:contextual_VOC} shows two examples of object-context mid-level visual elements discovered from class \emph{person}.

As depicted in Table~\ref{tab:voc_2007_context}, most categories have very low proportion of scene-context mid-level visual elements except for \emph{boat}, which has a relatively high value of $35\%$.

\begin{figure}[t]
\vspace{-0.0cm}
\begin{center}
\includegraphics[width=1\linewidth]{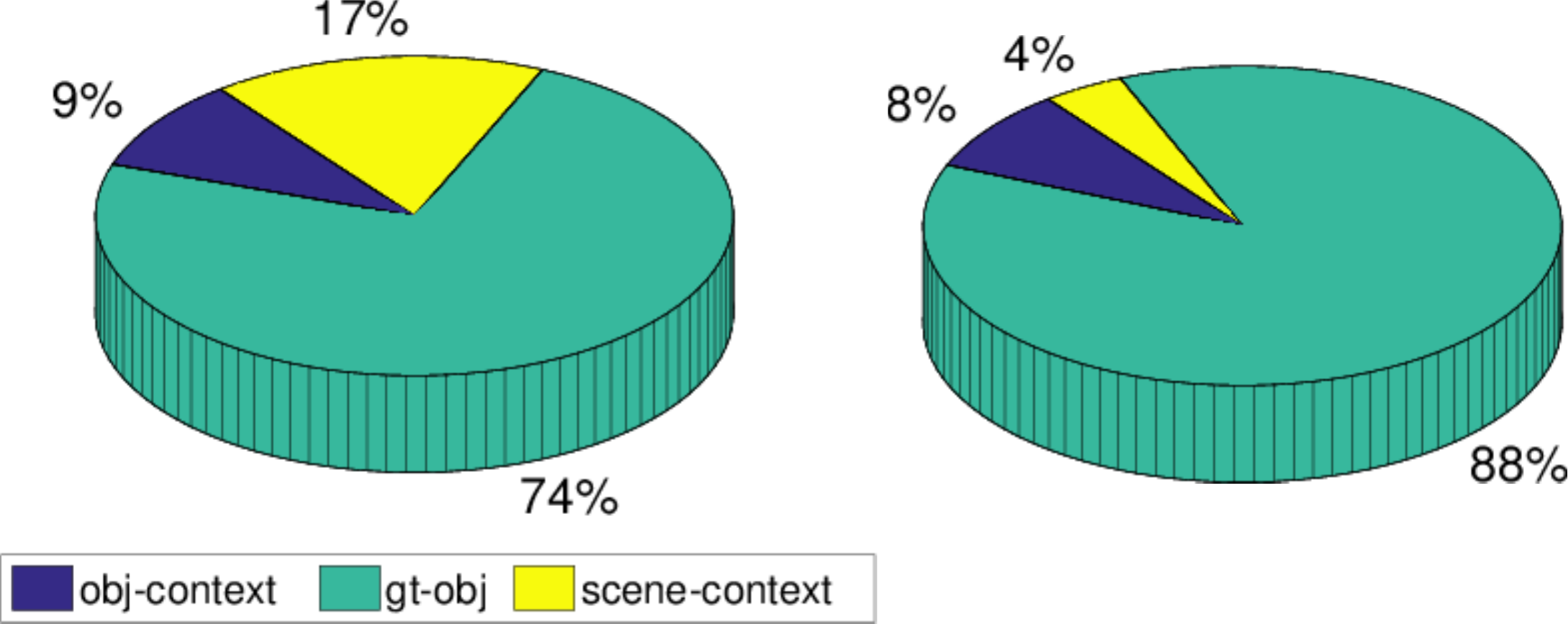} \\
\end{center}
\caption{Distributions of mid-level visual elements discovered using
different CNN models---\emph{CaffeRef} (left) and \emph{VGG-VD} (right).}
\label{fig:contextual_ratio__nets}
\vspace{-3mm}
\end{figure}

We also compare distributions of mid-level elements discovered using different CNN models (Fig.~\ref{fig:contextual_ratio__nets}).
As shown in Fig.~\ref{fig:contextual_ratio__nets}, for both CNN models,
the majority consists of those mid-level elements tend to capture parts of ground-truth objects and contextual ones only constitute a relatively small fraction.
Also, for mid-level visual elements capturing ground-truth objects, the fraction of those discovered from the \emph{VGG-VD} model bypasses that from the \emph{CaffeRef} model by $14\%$ ($88\%$ \vs $74\%$).
We thus conjecture that for image classification, deeper CNNs will more likely to learn to represent the underlying objects and contextual information may not be that valuable.

%% file: discussion.tex
\textcolor{blue}{Recently, some works on accelerating CNNs~\citep{courbariaux2016binarynet,rastegari2016xnor}
advocate using binary activation values in CNNs. 
It would be interesting to try binary CNN features for creating transactions.
In this case, for an image patch, all of its CNN dimensions with positive activation values will be kept to generate on transaction.
This means we do not need to select the K largest activation magnitudes as in the current approach (Sec.\ref{subsec:transaction}), and there will be no information loss for transaction creation at all.}

\textcolor{blue}{
As the feature dimension of the Bag-of-Elements representation (Sec.~\ref{subsubsec:Bag-of-Elements}) is proportion to the number of categories, most of the current works on mid-level visual elements, including ours, cannot 
be applied to image classification datasets which contain a huge number of categories (\eg, ImageNet~\citep{DBLP:conf/cvpr/DengDSLL009} and Places~\citep{DBLP:conf/nips/ZhouLXTO14}). 
A good indication of future work to address this scalability issue may be using shared mid-level visual elements~\citep{DBLP:journals/cviu/MettesGS16}. }

%% file: conclusion_future_work.tex
\textcolor{blue}{
We have addressed the task of mid-level visual element discovery from the perspective of pattern mining. More specifically, we have shown that CNN activation can be encoded into transactions, the data structure used by existing pattern mining techniques which can be readily applied to discover discriminative mid-level visual element candidates. We further develop different strategies to generate image representations from the mined visual element candidates. We experimentally demonstrate the effectiveness of the mined mid-level visual elements and achieve the state-of-the-art classification performance on various datasets by using the generated image representation. }

\textcolor{blue}{
Although this paper only addresses the image classification problem, our method can be extended to many other applications and serves as a bridge between visual recognition and pattern mining research fields. Since the publication of our conference paper~\citep{LiLSH15CVPR}, there have been several works~\citep{DBLP:conf/cvpr/DibaPPG,oramas2016modeling} which follow our approach to develop methods suited for various applications, including human action and attribute recognition~\citep{DBLP:conf/cvpr/DibaPPG} and modeling visual compatibility~\citep{oramas2016modeling}. }

\textcolor{blue}{In future work, we plan to investigate three directions to extend our approach. Firstly, we will develop efficient mining methods to mine the patterns that are shared across categories. This will address the limitation of the current method that it can only detect discriminative patterns for each category and thus is not very scalable to a dataset with a huge number of categories, \eg, ImageNet. Secondly, we will extend our method to the metric learning setting. In such a setting, the mined discriminative patterns are only used to make a binary decision, that is, whether the input two images are from the same category. Finally, we will apply our method to more applications, especially those that can leverage the state-of-the-art pattern mining techniques.}

%% file: template.bbl
\begin{thebibliography}{100}\itemsep=-1pt

\bibitem{DBLP:journals/ijcv/AgarwalT08}
A.~Agarwal and B.~Triggs.
\newblock Multilevel image coding with hyperfeatures.
\newblock {\em Int. J. Comp. Vis.}, 78(1):15--27, 2008.

\bibitem{DBLP:conf/eccv/Agrawal14}
P.~Agrawal, R.~Girshick, and J.~Malik.
\newblock Analyzing the performance of multilayer neural networks for object
  recognition.
\newblock In {\em Proc. Eur. Conf. Comp. Vis.}, pages 329--344, 2014.

\bibitem{DBLP:conf/vldb/AgrawalS94}
R.~Agrawal and R.~Srikant.
\newblock Fast algorithms for mining association rules in large databases.
\newblock In {\em Proc. Int. Conf. Very Large Databases}, pages 487--499, 1994.

\bibitem{DBLP:conf/cvpr/AubryMERSJ14}
M.~Aubry, D.~Maturana, A.~A. Efros, B.~C. Russell, and J.~Sivic.
\newblock Seeing 3d chairs: exemplar part-based 2d-3d alignment using a large
  dataset of cad models.
\newblock In {\em Proc. IEEE Conf. Comp. Vis. Patt. Recogn.}, pages 3762--3769,
  2014.

\bibitem{DBLP:journals/tog/AubryRS14}
M.~Aubry, B.~C. Russell, and J.~Sivic.
\newblock Painting-to-3d model alignment via discriminative visual elements.
\newblock {\em Proc. Ann. ACM SIGIR Conf.}, 33(2):14, 2014.

\bibitem{DBLP:journals/corr/AzizpourRSMC14}
H.~Azizpour, A.~S. Razavian, J.~Sullivan, A.~Maki, and S.~Carlsson.
\newblock Factors of transferability for a generic convnet representation.
\newblock {\em {IEEE} Trans. Pattern Anal. Mach. Intell.}, 2016.

\bibitem{bansal2015mid}
A.~Bansal, A.~Shrivastava, C.~Doersch, and A.~Gupta.
\newblock Mid-level elements for object detection.
\newblock {\em arXiv preprint arXiv:1504.07284}, 2015.

\bibitem{DBLP:journals/widm/Borgelt12}
C.~Borgelt.
\newblock Frequent item set mining.
\newblock {\em Wiley Interdisc. Rew.: Data Mining and Knowledge Discovery},
  2(6):437--456, 2012.

\bibitem{DBLP:conf/eccv/BossardGG14}
L.~Bossard, M.~Guillaumin, and L.~V. Gool.
\newblock Food-101 – mining discriminative components with random forests.
\newblock In {\em Proc. Eur. Conf. Comp. Vis.}, pages 446--461, 2014.

\bibitem{DBLP:conf/eccv/BourdevMBM10}
L.~D. Bourdev, S.~Maji, T.~Brox, and J.~Malik.
\newblock Detecting people using mutually consistent poselet activations.
\newblock In {\em Proc. Eur. Conf. Comp. Vis.}, pages 168--181, 2010.

\bibitem{DBLP:conf/iccv/BourdevMM11}
L.~D. Bourdev, S.~Maji, and J.~Malik.
\newblock Describing people: {A} poselet-based approach to attribute
  classification.
\newblock In {\em Proc. IEEE Int. Conf. Comp. Vis.}, pages 1543--1550, 2011.

\bibitem{DBLP:conf/iccv/BourdevM09}
L.~D. Bourdev and J.~Malik.
\newblock Poselets: Body part detectors trained using 3d human pose
  annotations.
\newblock In {\em Proc. IEEE Int. Conf. Comp. Vis.}, pages 1365--1372, 2009.

\bibitem{DBLP:conf/cvpr/BoureauBLP10}
Y.~Boureau, F.~R. Bach, Y.~LeCun, and J.~Ponce.
\newblock Learning mid-level features for recognition.
\newblock In {\em Proc. IEEE Conf. Comp. Vis. Patt. Recogn.}, 2010.

\bibitem{DBLP:conf/bmvc/ChatfieldSVZ14}
K.~Chatfield, K.~Simonyan, A.~Vedaldi, and A.~Zisserman.
\newblock Return of the devil in the details: Delving deep into convolutional
  nets.
\newblock In {\em Proc. Brit. Mach. Vis. Conf.}, 2014.

\bibitem{DBLP:conf/icde/ChengYHY08}
H.~Cheng, X.~Yan, J.~Han, and P.~S. Yu.
\newblock Direct discriminative pattern mining for effective classification.
\newblock In {\em Proc. IEEE Int. Conf. Data Engr.}, pages 169--178, 2008.

\bibitem{DBLP:journals/pami/ChoiTW12}
M.~J. Choi, A.~Torralba, and A.~S. Willsky.
\newblock A tree-based context model for object recognition.
\newblock {\em {IEEE} Trans. Pattern Anal. Mach. Intell.}, 34(2):240--252,
  2012.

\bibitem{DBLP:journals/ijcv/CimpoiMKV16}
M.~Cimpoi, S.~Maji, I.~Kokkinos, and A.~Vedaldi.
\newblock Deep filter banks for texture recognition, description, and
  segmentation.
\newblock {\em Int. J. Comp. Vis.}, 118(1):65--94, 2016.

\bibitem{CimpoiMV15CVPR}
M.~Cimpoi, S.~Maji, and A.~Vedaldi.
\newblock Deep filter banks for texture recognition and segmentation.
\newblock In {\em Proc. IEEE Conf. Comp. Vis. Patt. Recogn.}, pages 3828--3836,
  2015.

\bibitem{courbariaux2016binarynet}
M.~Courbariaux and Y.~Bengio.
\newblock Binarynet: Training deep neural networks with weights and activations
  constrained to+ 1 or-1.
\newblock {\em arXiv preprint arXiv:1602.02830}, 2016.

\bibitem{DBLP:conf/bmvc/CrowleyZ14}
E.~Crowley and A.~Zisserman.
\newblock The state of the art: Object retrieval in paintings using
  discriminative regions.
\newblock In {\em Proc. Brit. Mach. Vis. Conf.}, 2014.

\bibitem{DBLP:conf/cvpr/DengDSLL009}
J.~Deng, W.~Dong, R.~Socher, L.-J. Li, K.~Li, and F.-F. Li.
\newblock Imagenet: A large-scale hierarchical image database.
\newblock In {\em Proc. IEEE Conf. Comp. Vis. Patt. Recogn.}, pages 248--255,
  2009.

\bibitem{DBLP:conf/cvpr/DibaPPG}
A.~Diba, A.~M. Pazandeh, H.~Pirsiavash, and L.~V. Gool.
\newblock Deepcamp: Deep convolutional action \& attribute mid-level patterns.
\newblock In {\em Proc. IEEE Conf. Comp. Vis. Patt. Recogn.}, 2016.

\bibitem{DBLP:conf/cvpr/DivvalaHHEH09}
S.~K. Divvala, D.~Hoiem, J.~Hays, A.~A. Efros, and M.~Hebert.
\newblock An empirical study of context in object detection.
\newblock In {\em Proc. IEEE Conf. Comp. Vis. Patt. Recogn.}, pages 1271--1278,
  2009.

\bibitem{DBLP:conf/nips/DoerschGE13}
C.~Doersch, A.~Gupta, and A.~A. Efros.
\newblock Mid-level visual element discovery as discriminative mode seeking.
\newblock In {\em Proc. Adv. Neural Inf. Process. Syst.}, pages 494--502, 2013.

\bibitem{DBLP:journals/tog/DoerschSGSE12}
C.~Doersch, S.~Singh, A.~Gupta, J.~Sivic, and A.~A. Efros.
\newblock What makes paris look like paris?
\newblock {\em Proc. Ann. ACM SIGIR Conf.}, 31(4):101, 2012.

\bibitem{dosovitskiy2015inverting}
A.~Dosovitskiy and T.~Brox.
\newblock Inverting visual representations with convolutional networks.
\newblock In {\em Proc. IEEE Conf. Comp. Vis. Patt. Recogn.}, 2016.

\bibitem{DBLP:conf/cvpr/EndresSJH13}
I.~Endres, K.~J. Shih, J.~Jiaa, and D.~Hoiem.
\newblock Learning collections of part models for object recognition.
\newblock In {\em Proc. IEEE Conf. Comp. Vis. Patt. Recogn.}, pages 939--946,
  2013.

\bibitem{DBLP:journals/ijcv/EveringhamEGWWZ15}
M.~Everingham, S.~M.~A. Eslami, L.~V. Gool, C.~K.~I. Williams, J.~M. Winn, and
  A.~Zisserman.
\newblock The pascal visual object classes challenge: {A} retrospective.
\newblock {\em Int. J. Comp. Vis.}, 111(1):98--136, 2015.

\bibitem{DBLP:journals/ijcv/EveringhamGWWZ10}
M.~Everingham, L.~J.~V. Gool, C.~K.~I. Williams, J.~M. Winn, and A.~Zisserman.
\newblock The pascal visual object classes {(VOC)} challenge.
\newblock {\em Int. J. Comp. Vis.}, 88(2):303--338, 2010.

\bibitem{DBLP:journals/jmlr/FanCHWL08}
R.-E. Fan, K.-W. Chang, C.-J. Hsieh, X.-R. Wang, and C.-J. Lin.
\newblock Liblinear: A library for large linear classification.
\newblock {\em J. Mach. Learn. Res.}, 9:1871--1874, 2008.

\bibitem{DBLP:journals/pami/FelzenszwalbGMR10}
P.~F. Felzenszwalb, R.~B. Girshick, D.~A. McAllester, and D.~Ramanan.
\newblock Object detection with discriminatively trained part-based models.
\newblock {\em {IEEE} Trans. Pattern Anal. Mach. Intell.}, 32(9):1627--1645,
  2010.

\bibitem{DBLP:conf/eccv/FernandoFT12}
B.~Fernando, {\'E}.~Fromont, and T.~Tuytelaars.
\newblock Effective use of frequent itemset mining for image classification.
\newblock In {\em Proc. Eur. Conf. Comp. Vis.}, pages 214--227, 2012.

\bibitem{DBLP:journals/ijcv/FernandoFT14}
B.~Fernando, {\'E}.~Fromont, and T.~Tuytelaars.
\newblock Mining mid-level features for image classification.
\newblock {\em Int. J. Comp. Vis.}, 108(3):186--203, 2014.

\bibitem{DBLP:conf/iccv/FernandoT13}
B.~Fernando and T.~Tuytelaars.
\newblock Mining multiple queries for image retrieval: On-the-fly learning of
  an object-specific mid-level representation.
\newblock In {\em Proc. IEEE Int. Conf. Comp. Vis.}, pages 2544--2551, 2013.

\bibitem{DBLP:conf/iccv/FouheyGH13}
D.~F. Fouhey, A.~Gupta, and M.~Hebert.
\newblock Data-driven 3d primitives for single image understanding.
\newblock In {\em Proc. IEEE Int. Conf. Comp. Vis.}, pages 3392--3399, 2013.

\bibitem{DBLP:conf/iccv/FouheyHGH15}
D.~F. Fouhey, W.~Hussain, A.~Gupta, and M.~Hebert.
\newblock Single image 3d without a single 3d image.
\newblock In {\em Proc. IEEE Int. Conf. Comp. Vis.}, pages 1053--1061, 2015.

\bibitem{DBLP:conf/cvpr/GaoBZD16}
Y.~Gao, O.~Beijbom, N.~Zhang, and T.~Darrell.
\newblock Compact bilinear pooling.
\newblock In {\em Proc. IEEE Conf. Comp. Vis. Patt. Recogn.}, 2016.

\bibitem{DBLP:conf/accv/GilbertB14}
A.~Gilbert and R.~Bowden.
\newblock Data mining for action recognition.
\newblock In {\em Proc. Asian Conf. Comp. Vis.}, pages 290--303, 2014.

\bibitem{DBLP:journals/pami/GilbertIB11}
A.~Gilbert, J.~Illingworth, and R.~Bowden.
\newblock Action recognition using mined hierarchical compound features.
\newblock {\em {IEEE} Trans. Pattern Anal. Mach. Intell.}, 33(5):883--897,
  2011.

\bibitem{DBLP:conf/cvpr/GirshickDDM14}
R.~Girshick, J.~Donahue, T.~Darrell, and J.~Malik.
\newblock Rich feature hierarchies for accurate object detection and semantic
  segmentation.
\newblock In {\em Proc. IEEE Conf. Comp. Vis. Patt. Recogn.}, pages 580--587,
  2014.

\bibitem{DBLP:journals/pami/GirshickDDM16}
R.~B. Girshick, J.~Donahue, T.~Darrell, and J.~Malik.
\newblock Region-based convolutional networks for accurate object detection and
  segmentation.
\newblock {\em {IEEE} Trans. Pattern Anal. Mach. Intell.}, 38(1):142--158,
  2016.

\bibitem{DBLP:conf/eccv/GongWGL14}
Y.~Gong, L.~Wang, R.~Guo, and S.~Lazebnik.
\newblock Multi-scale orderless pooling of deep convolutional activation
  features.
\newblock In {\em Proc. Eur. Conf. Comp. Vis.}, pages 392--407, 2014.

\bibitem{DBLP:journals/tkde/GrahneZ05}
G.~Grahne and J.~Zhu.
\newblock Fast algorithms for frequent itemset mining using fp-trees.
\newblock {\em {IEEE} Trans. Knowl. Data Eng.}, 17(10):1347--1362, 2005.

\bibitem{DBLP:conf/eccv/HariharanMR12}
B.~Hariharan, J.~Malik, and D.~Ramanan.
\newblock Discriminative decorrelation for clustering and classification.
\newblock In {\em Proc. Eur. Conf. Comp. Vis.}, pages 459--472, 2012.

\bibitem{DBLP:journals/pami/HeZRS15}
K.~He, X.~Zhang, S.~Ren, and J.~Sun.
\newblock Spatial pyramid pooling in deep convolutional networks for visual
  recognition.
\newblock {\em {IEEE} Trans. Pattern Anal. Mach. Intell.}, 37(9):1904--1916,
  2015.

\bibitem{DBLP:journals/ijcv/HoiemEH08}
D.~Hoiem, A.~A. Efros, and M.~Hebert.
\newblock Putting objects in perspective.
\newblock {\em Int. J. Comp. Vis.}, 80(1):3--15, 2008.

\bibitem{DBLP:conf/cvpr/JainGRD13}
A.~Jain, A.~Gupta, M.~Rodriguez, and L.~S. Davis.
\newblock Representing videos using mid-level discriminative patches.
\newblock In {\em Proc. IEEE Conf. Comp. Vis. Patt. Recogn.}, pages 2571--2578,
  2013.

\bibitem{DBLP:conf/cvpr/JegouDSP10}
H.~Jegou, M.~Douze, C.~Schmid, and P.~P{\'e}rez.
\newblock Aggregating local descriptors into a compact image representation.
\newblock In {\em Proc. IEEE Conf. Comp. Vis. Patt. Recogn.}, pages 3304--3311,
  2010.

\bibitem{Jia13caffe}
Y.~Jia, E.~Shelhamer, J.~Donahue, S.~Karayev, J.~Long, R.~Girshick,
  S.~Guadarrama, and T.~Darrell.
\newblock Caffe: Convolutional architecture for fast feature embedding.
\newblock {\em arXiv preprint arXiv:1408.5093}, 2014.

\bibitem{DBLP:conf/cvpr/JunejaVJZ13}
M.~Juneja, A.~Vedaldi, C.~V. Jawahar, and A.~Zisserman.
\newblock Blocks that shout: Distinctive parts for scene classification.
\newblock In {\em Proc. IEEE Conf. Comp. Vis. Patt. Recogn.}, pages 923--930,
  2013.

\bibitem{DBLP:conf/nips/KrizhevskySH12}
A.~Krizhevsky, I.~Sutskever, and G.~E. Hinton.
\newblock Imagenet classification with deep convolutional neural networks.
\newblock In {\em Proc. Adv. Neural Inf. Process. Syst.}, pages 1106--1114,
  2012.

\bibitem{DBLP:conf/cvpr/LazebnikSP06}
S.~Lazebnik, C.~Schmid, and J.~Ponce.
\newblock Beyond bags of features: Spatial pyramid matching for recognizing
  natural scene categories.
\newblock In {\em Proc. IEEE Conf. Comp. Vis. Patt. Recogn.}, pages 2169--2178,
  2006.

\bibitem{DBLP:conf/iccv/LeeEM13}
Y.~J. Lee, A.~A. Efros, and M.~Hebert.
\newblock Style-aware mid-level representation for discovering visual
  connections in space and time.
\newblock In {\em Proc. IEEE Int. Conf. Comp. Vis.}, pages 1857--1864, 2013.

\bibitem{DBLP:conf/cvpr/LiWT13}
Q.~Li, J.~Wu, and Z.~Tu.
\newblock Harvesting mid-level visual concepts from large-scale internet
  images.
\newblock In {\em Proc. IEEE Conf. Comp. Vis. Patt. Recogn.}, pages 851--858,
  2013.

\bibitem{LiLSH15CVPR}
Y.~Li, L.~Liu, C.~Shen, and A.~van~den Hengel.
\newblock Mid-level deep pattern mining.
\newblock In {\em Proc. IEEE Conf. Comp. Vis. Patt. Recogn.}, pages 971--980,
  2015.

\bibitem{DBLP:conf/iccv/LinRM15}
T.~Lin, A.~RoyChowdhury, and S.~Maji.
\newblock Bilinear {CNN} models for fine-grained visual recognition.
\newblock In {\em Proc. IEEE Int. Conf. Comp. Vis.}, pages 1449--1457, 2015.

\bibitem{DBLP:conf/cvpr/LiuSH15}
L.~Liu, C.~Shen, and A.~van~den Hengel.
\newblock The treasure beneath convolutional layers: Cross convolutional layer
  pooling for image classification.
\newblock In {\em Proc. IEEE Conf. Comp. Vis. Patt. Recogn.}, pages 4749--4757,
  2015.

\bibitem{DBLP:conf/nips/LiuSWHW14}
L.~Liu, C.~Shen, L.~Wang, A.~van~den Hengel, and C.~Wang.
\newblock Encoding high dimensional local features by sparse coding based
  fisher vectors.
\newblock In {\em Proc. Adv. Neural Inf. Process. Syst.}, pages 1143--1151,
  2014.

\bibitem{DBLP:conf/cvpr/LiuW12}
L.~Liu and L.~Wang.
\newblock What has my classifier learned? visualizing the classification rules
  of bag-of-feature model by support region detection.
\newblock In {\em Proc. IEEE Conf. Comp. Vis. Patt. Recogn.}, pages 3586--3593,
  2012.

\bibitem{DBLP:conf/nips/MalisiewiczE09}
T.~Malisiewicz and A.~A. Efros.
\newblock Beyond categories: The visual memex model for reasoning about object
  relationships.
\newblock In {\em Proc. Adv. Neural Inf. Process. Syst.}, pages 1222--1230,
  2009.

\bibitem{DBLP:conf/iccv/MalisiewiczGE11}
T.~Malisiewicz, A.~Gupta, and A.~A. Efros.
\newblock Ensemble of exemplar-svms for object detection and beyond.
\newblock In {\em Proc. IEEE Int. Conf. Comp. Vis.}, pages 89--96, 2011.

\bibitem{DBLP:conf/iccv/MatzenS15}
K.~Matzen and N.~Snavely.
\newblock Bubblenet: Foveated imaging for visual discovery.
\newblock In {\em Proc. IEEE Int. Conf. Comp. Vis.}, pages 1931--1939, 2015.

\bibitem{DBLP:journals/cviu/MettesGS16}
P.~Mettes, J.~C. van Gemert, and C.~G.~M. Snoek.
\newblock No spare parts: Sharing part detectors for image categorization.
\newblock {\em Comp. Vis. Image Understanding}, 2016.

\bibitem{DBLP:conf/cvpr/OquabBL14}
M.~Oquab, L.~Bottou, I.~Laptev, and J.~Sivic.
\newblock Learning and transferring mid-level image representations using
  convolutional neural networks.
\newblock In {\em Proc. IEEE Conf. Comp. Vis. Patt. Recogn.}, pages 1717--1724,
  2014.

\bibitem{oramas2016modeling}
J.~Oramas and T.~Tuytelaars.
\newblock Modeling visual compatibility through hierarchical mid-level
  elements.
\newblock {\em arXiv preprint arXiv:1604.00036}, 2016.

\bibitem{DBLP:conf/iccv/OwensXTF13}
A.~Owens, J.~Xiao, A.~Torralba, and W.~T. Freeman.
\newblock Shape anchors for data-driven multi-view reconstruction.
\newblock In {\em Proc. IEEE Int. Conf. Comp. Vis.}, pages 33--40, 2013.

\bibitem{Parizi15ICLR}
S.~N. Parizi, A.~Vedaldi, A.~Zisserman, and P.~Felzenszwalb.
\newblock Automatic discovery and optimization of parts for image
  classification.
\newblock In {\em Proc. Int. Conf. Learn. Repr.}, 2015.

\bibitem{DBLP:conf/cvpr/PerronninLSP10}
F.~Perronnin, Y.~Liu, J.~S{\'a}nchez, and H.~Poirier.
\newblock Large-scale image retrieval with compressed fisher vectors.
\newblock In {\em Proc. IEEE Conf. Comp. Vis. Patt. Recogn.}, pages 3384--3391,
  2010.

\bibitem{DBLP:conf/eccv/PerronninSM10}
F.~Perronnin, J.~S{\'{a}}nchez, and T.~Mensink.
\newblock Improving the fisher kernel for large-scale image classification.
\newblock In {\em Proc. Eur. Conf. Comp. Vis.}, pages 143--156, 2010.

\bibitem{DBLP:conf/iccv/QuackFLG07}
T.~Quack, V.~Ferrari, B.~Leibe, and L.~J.~V. Gool.
\newblock Efficient mining of frequent and distinctive feature configurations.
\newblock In {\em Proc. IEEE Int. Conf. Comp. Vis.}, pages 1--8, 2007.

\bibitem{DBLP:conf/cvpr/QuattoniT09}
A.~Quattoni and A.~Torralba.
\newblock Recognizing indoor scenes.
\newblock In {\em Proc. IEEE Conf. Comp. Vis. Patt. Recogn.}, pages 413--420,
  2009.

\bibitem{rastegari2016xnor}
M.~Rastegari, V.~Ordonez, J.~Redmon, and A.~Farhadi.
\newblock Xnor-net: Imagenet classification using binary convolutional neural
  networks.
\newblock {\em arXiv preprint arXiv:1603.05279}, 2016.

\bibitem{6910029}
A.~S. Razavian, H.~Azizpour, J.~Sullivan, and S.~Carlsson.
\newblock Cnn features off-the-shelf: An astounding baseline for recognition.
\newblock In {\em Proc. IEEE Conf. Comp. Vis. Patt. Recogn. Workshops}, pages
  512--519, 2014.

\bibitem{Rematas15CVPR}
K.~Rematas, B.~Fernando, F.~Dellaert, and T.~Tuytelaars.
\newblock Dataset fingerprints: Exploring image collections through data
  mining.
\newblock In {\em Proc. IEEE Conf. Comp. Vis. Patt. Recogn.}, pages 4867--4875,
  2015.

\bibitem{DBLP:journals/corr/RussakovskyDSKSMHKKBBF14}
O.~Russakovsky, J.~Deng, H.~Su, J.~Krause, S.~Satheesh, S.~Ma, Z.~Huang,
  A.~Karpathy, A.~Khosla, M.~S. Bernstein, A.~C. Berg, and L.~Fei{-}Fei.
\newblock Imagenet large scale visual recognition challenge.
\newblock {\em Int. J. Comp. Vis.}, 115(3):211--252, 2015.

\bibitem{DBLP:journals/pami/ShihEH15}
K.~J. Shih, I.~Endres, and D.~Hoiem.
\newblock Learning discriminative collections of part detectors for object
  recognition.
\newblock {\em {IEEE} Trans. Pattern Anal. Mach. Intell.}, 37(8):1571--1584,
  2015.

\bibitem{DBLP:journals/tog/ShrivastavaMGE11}
A.~Shrivastava, T.~Malisiewicz, A.~Gupta, and A.~A. Efros.
\newblock Data-driven visual similarity for cross-domain image matching.
\newblock {\em Proc. Ann. ACM SIGIR Conf.}, 30(6):154, 2011.

\bibitem{DBLP:conf/nips/SimonyanVZ13}
K.~Simonyan, A.~Vedaldi, and A.~Zisserman.
\newblock Deep fisher networks for large-scale image classification.
\newblock In {\em Proc. Adv. Neural Inf. Process. Syst.}, pages 163--171, 2013.

\bibitem{Simonyan15ICLR}
K.~Simonyan and A.~Zisserman.
\newblock Very deep convolutional networks for large-scale image recognition.
\newblock In {\em Proc. Int. Conf. Learn. Repr.}, 2015.

\bibitem{DBLP:conf/eccv/SinghGE12}
S.~Singh, A.~Gupta, and A.~A. Efros.
\newblock Unsupervised discovery of mid-level discriminative patches.
\newblock In {\em Proc. Eur. Conf. Comp. Vis.}, pages 73--86, 2012.

\bibitem{DBLP:conf/iccv/SivicZ03}
J.~Sivic and A.~Zisserman.
\newblock Video google: A text retrieval approach to object matching in videos.
\newblock In {\em Proc. IEEE Int. Conf. Comp. Vis.}, pages 1470--1477, 2003.

\bibitem{DBLP:conf/nips/SongLJD14}
H.~O. Song, Y.~J. Lee, S.~Jegelka, and T.~Darrell.
\newblock Weakly-supervised discovery of visual pattern configurations.
\newblock In {\em Proc. Adv. Neural Inf. Process. Syst.}, pages 1637--1645,
  2014.

\bibitem{DBLP:conf/iccv/SunP13}
J.~Sun and J.~Ponce.
\newblock Learning discriminative part detectors for image classification and
  cosegmentation.
\newblock In {\em Proc. IEEE Int. Conf. Comp. Vis.}, pages 3400--3407, 2013.

\bibitem{DBLP:journals/ijcv/SunP16}
J.~Sun and J.~Ponce.
\newblock Learning dictionary of discriminative part detectors for image
  categorization and cosegmentation.
\newblock {\em Int. J. Comp. Vis.}, pages 1--23, 2016.

\bibitem{DBLP:journals/ijcv/Torralba03}
A.~Torralba.
\newblock Contextual priming for object detection.
\newblock {\em Int. J. Comp. Vis.}, 53(2):169--191, 2003.

\bibitem{DBLP:conf/fimi/UnoAUA03}
T.~Uno, T.~Asai, Y.~Uchida, and H.~Arimura.
\newblock {LCM:} an efficient algorithm for enumerating frequent closed item
  sets.
\newblock In {\em FIMI}, 2003.

\bibitem{DBLP:conf/cvpr/Voravuthikunchai14}
W.~Voravuthikunchai, B.~Cr{\'e}milleux, and F.~Jurie.
\newblock Histograms of pattern sets for image classification and object
  recognition.
\newblock In {\em Proc. IEEE Conf. Comp. Vis. Patt. Recogn.}, pages 224--231,
  2014.

\bibitem{DBLP:journals/datamine/VreekenLS11}
J.~Vreeken, M.~van Leeuwen, and A.~Siebes.
\newblock Krimp: mining itemsets that compress.
\newblock {\em Data Min. Knowl. Discov.}, 23(1):169--214, 2011.

\bibitem{DBLP:journals/pami/WangLWY14}
J.~Wang, Z.~Liu, Y.~Wu, and J.~Yuan.
\newblock Learning actionlet ensemble for 3d human action recognition.
\newblock {\em {IEEE} Trans. Pattern Anal. Mach. Intell.}, 36(5):914--927,
  2014.

\bibitem{DBLP:conf/cvpr/WangYMHHX16}
J.~Wang, Y.~Yang, J.~Mao, Z.~Huang, and C.~H.~W. Xu.
\newblock Cnn--rnn: A unified framework for multi-label image classification.
\newblock In {\em Proc. IEEE Conf. Comp. Vis. Patt. Recogn.}, 2016.

\bibitem{DBLP:conf/cvpr/WangQT13}
L.~Wang, Y.~Qiao, and X.~Tang.
\newblock Motionlets: Mid-level 3d parts for human motion recognition.
\newblock In {\em Proc. IEEE Conf. Comp. Vis. Patt. Recogn.}, pages 2674--2681,
  2013.

\bibitem{DBLP:conf/icml/WangWBLT13}
X.~Wang, B.~Wang, X.~Bai, W.~Liu, and Z.~Tu.
\newblock Max-margin multiple-instance dictionary learning.
\newblock In {\em Proc. Int. Conf. Mach. Learn.}, pages 846--854, 2013.

\bibitem{DBLP:conf/cvpr/WangCMD16}
Y.~Wang, J.~Choi, V.~I. Morariu, and L.~S. Davis.
\newblock Mining discriminative triplets of patches for fine-grained
  classification.
\newblock In {\em Proc. IEEE Conf. Comp. Vis. Patt. Recogn.}, 2016.

\bibitem{DBLP:journals/corr/WeiXHNDZY14}
Y.~Wei, W.~Xia, J.~Huang, B.~Ni, J.~Dong, Y.~Zhao, and S.~Yan.
\newblock {CNN:} single-label to multi-label.
\newblock {\em CoRR}, abs/1406.5726, 2014.

\bibitem{DBLP:conf/cvpr/YaoF10}
B.~Yao and L.~Fei-Fei.
\newblock Grouplet: A structured image representation for recognizing human and
  object interactions.
\newblock In {\em Proc. IEEE Conf. Comp. Vis. Patt. Recogn.}, pages 9--16,
  2010.

\bibitem{yoo2015multiscale}
D.~Yoo, S.~Park, J.-Y. Lee, and I.~S. Kweon.
\newblock Multi-scale pyramid pooling for deep convolutional representation.
\newblock In {\em Proc. IEEE Conf. Comp. Vis. Patt. Recogn. Workshops}, pages
  71--80, 2015.

\bibitem{DBLP:conf/cvpr/YuanWY07}
J.~Yuan, Y.~Wu, and M.~Yang.
\newblock Discovery of collocation patterns: from visual words to visual
  phrases.
\newblock In {\em Proc. IEEE Conf. Comp. Vis. Patt. Recogn.}, 2007.

\bibitem{DBLP:conf/eccv/ZeilerF14}
M.~D. Zeiler and R.~Fergus.
\newblock Visualizing and understanding convolutional networks.
\newblock In {\em Proc. Eur. Conf. Comp. Vis.}, pages 818--833, 2014.

\bibitem{DBLP:conf/cvpr/ZhaoOW14}
R.~Zhao, W.~Ouyang, and X.~Wang.
\newblock Learning mid-level filters for person re-identification.
\newblock In {\em Proc. IEEE Conf. Comp. Vis. Patt. Recogn.}, pages 144--151,
  2014.

\bibitem{DBLP:conf/nips/ZhouLXTO14}
B.~Zhou, {\`{A}}.~Lapedriza, J.~Xiao, A.~Torralba, and A.~Oliva.
\newblock Learning deep features for scene recognition using places database.
\newblock In {\em Proc. Adv. Neural Inf. Process. Syst.}, pages 487--495, 2014.

\end{thebibliography}
